%% file: main_dissertation.tex
\documentclass[12pt]{report}	

\usepackage{utdiss2}  		

\usepackage{epsfig}

\usepackage{placeins}
\usepackage{adjustbox}
\usepackage{stackengine}
\usepackage{booktabs}

\usepackage{widetable}

\usepackage{color}
\usepackage[dvipsnames]{xcolor}

\usepackage{xspace}
\usepackage{float}
\usepackage{contour}
\usepackage{xkeyval}

\usepackage{epsfig}
\usepackage{graphicx}
\usepackage{amsmath}
\usepackage{amssymb}
\usepackage{graphicx,multirow}
\usepackage{subfig}
\usepackage{url}
\usepackage{natbib}
\usepackage{caption}
\usepackage{overpic}
\usepackage{wrapfig}

\usepackage{siunitx}
\usepackage{algpseudocode}
\usepackage{svg}
\usepackage{rotating}
\usepackage{lscape}
\usepackage{enumitem}

\usepackage{pifont}         
\usepackage{amsmath,amsthm,amsfonts,amscd} 
\usepackage{eucal} 	 	
\usepackage{verbatim}      	
\usepackage{makeidx}       	
\usepackage{url}		

\author{Siming Yan}  	

\address{2317 Speedway\\ Austin, Texas 78712}  

\title{Representation Learning for Point Cloud Understanding}

\supervisor
	{Qixing Huang}

\committeemembers
	[Qiang Liu]
	[Georgios Pavlakos]
	{Gang Hua}

\previousdegrees{B.S.}
\oneandonehalfspacequote

\newtheorem{thm}{Theorem}[section]
\newtheorem{cor}[thm]{Corollary}

\newtheorem{prop}[thm]{Proposition}

\theoremstyle{definition}
\newtheorem{defn}{Definition}[section]

\theoremstyle{remark}

\input{math_commands}

\newcommand{\latexe}{{\LaTeX\kern.125em2%
                      \lower.5ex\hbox{$\varepsilon$}}}

\chardef\bslash=`\\	

\makeatletter		
\def\square{\RIfM@\bgroup\else$\bgroup\aftergroup$\fi
  \vcenter{\hrule\hbox{\vrule\@height.6em\kern.6em\vrule}%
                                              \hrule}\egroup}
\makeatother		

\makeindex    

\makeatletter
\DeclareRobustCommand\onedot{\futurelet\@let@token\@onedot}
\def\@onedot{\ifx\@let@token.\else.\null\fi\xspace}

\def\eg{\emph{e.g}\onedot} 
\def\ie{\emph{i.e}\onedot} 
\def\cf{\emph{c.f}\onedot}

\def\etal{\emph{et al}\onedot}
\makeatother

\usepackage{hyperref}
\hypersetup{
    colorlinks,
    citecolor=black,
    filecolor=black,
    linkcolor=black,
    urlcolor=black
}

\usepackage[ruled]{algorithm2e} 
\input{macros}

\begin{document}


%
%
%


%


%
\input{chapters/abstract}

\tableofcontents   

\listoftables      

\makeatletter
\renewcommand{\l@figure}{\@dottedtocline{1}{2.5em}{2.3em}}
\makeatother

\sloppy \listoffigures     

%
%

%
%

\input{chapters/introduction}

\pagebreak
\input{chapters/background}

\pagebreak

\renewcommand{\thefootnote}{\arabic{footnote}}

\input{chapters/pc_analysis}

\input{chapters/pc_ssl}

\input{chapters/pc_2dto3d}

\pagebreak

\pagebreak
\input{chapters/conclusion}

\pagebreak

\bibliographystyle{plain}  
\bibliography{refs}        

\index{Bibliography@\emph{Bibliography}}






\end{document}

%% file: math_commands.tex
\usepackage{amsmath,amsfonts,bm}

\def\eqref#1{equation~\ref{#1}}

\def\1{\bm{1}}

\DeclareMathAlphabet{\mathsfit}{\encodingdefault}{\sfdefault}{m}{sl}
\SetMathAlphabet{\mathsfit}{bold}{\encodingdefault}{\sfdefault}{bx}{n}

\newcommand{\R}{\mathbb{R}}

\DeclareMathOperator*{\argmin}{arg\,min}

%% file: macros.tex
\SetAlFnt{\small}
\SetAlCapFnt{\small}
\SetAlCapNameFnt{\small}
\SetAlCapHSkip{0pt}
\IncMargin{-\parindent}

\let \bs=\mathbf
\let \set=\mathcal

\def \good {\textup{good}}

\def \gt {\textup{gt}}

\def \saliency {\textup{\saliency}}

\def \gt {\mathit{gt}}

\def \path {\mathit{path}}

\def \line {\textup{edge}}

\def \edge {\textup{edge}}

\def \saliency {\textup{\saliency}}

\def \gt {\mathit{gt}}

\def \path {\textup{path}}

%% file: chapters/abstract.tex
\utabstract
\index{Abstract}%
\indent

With the rapid advancement of technology, 3D data acquisition and utilization have become increasingly prevalent across various fields, including computer vision, robotics, and geospatial analysis. 3D data, captured through methods such as 3D scanners, LiDARs, and RGB-D cameras, provides rich geometric, shape, and scale information. When combined with 2D images, 3D data offers machines a comprehensive understanding of their environment, benefiting applications like autonomous driving, robotics, remote sensing, and medical treatment.
3D data can be represented in various forms—voxel grids, meshes, and point clouds—each suited for different applications and processing techniques. Among these, point clouds are essential for tasks such as classification, segmentation, and object detection. Researchers have explored both supervised and self-supervised representation learning techniques to enhance these tasks. Supervised learning, relying on large annotated datasets, has achieved significant success but is limited by the availability of labeled data. Self-supervised learning, on the other hand, leverages vast amounts of unlabeled data through pretext tasks to learn effective representations, showing promise in reducing the need for labeled data.
This dissertation focuses on three main areas: supervised representation learning for point cloud primitive segmentation, self-supervised learning methods, and transfer learning from 2D to 3D. First, we introduce HPNet, a novel model for primitive segmentation that combines traditional geometric heuristics with deep learning approaches. HPNet achieves significant performance gains by using a hybrid point descriptor that merges learned semantic and spectral descriptors.
In the second part, we propose an asymmetric point-cloud autoencoder integrating with implicit function, which is more effective at capturing generalizable features from true 3D geometry than standard point-cloud autoencoders. We also design a novel attention-based decoder for masked autoencoders (MAEs), focusing on recovering high-order features rather than point positions. Our methods demonstrate superior performance in leveraging intrinsic point features for point cloud self-supervised learning.
Finally, we explore transfer learning from 2D to 3D by utilizing multi-view representation as a connecting bridge. Our approach, which integrates pre-trained 2D models to support 3D network training, significantly improves 3D understanding without merely transforming 2D data. Extensive experiments validate the effectiveness of our methods, showcasing their potential to advance point cloud representation learning by effectively integrating 2D knowledge.

%% file: chapters/introduction.tex
\chapter{Introduction}
\label{sec:introduction}

With the rapid advancement of technology, the acquisition and utilization of 3D data~\cite{daneshmand20183d} have become increasingly prevalent across various fields, including computer vision, robotics, and geospatial analysis. The collection of 3D data can be achieved through a variety of methods, such as different types of 3D scanners, LiDARs, and RGB-D cameras such as Kinect. These sensors capture rich geometric, shape, and scale information. When complemented with 2D images, 3D data enables machines to gain a more comprehensive understanding of their surrounding environment. This enhanced understanding has numerous applications across different areas, including autonomous driving, robotics, remote sensing, and medical treatment.

3D data can be represented in several forms, each suited to different applications and processing techniques. The primary representations include voxel grids, meshes, and point clouds. Voxel grids divide the 3D space into a regular grid of voxels, where each voxel holds information about the presence or absence of an object within that space. Meshes represent surfaces using vertices, edges, and faces, providing a continuous and detailed depiction of 3D shapes. Point clouds represent 3D data as a collection of discrete points, each defined by its spatial coordinates (x, y, z) and, occasionally, additional attributes such as color or intensity. As a commonly used format, point cloud representation preserves the original geometric information in 3D space. These collections of points capture spatial information about the surrounding environment, making them essential for various applications such as autonomous driving, robotics, and augmented reality. Understanding point clouds involves a range of tasks including classification, segmentation, and object detection, each requiring robust and efficient methods to process and analyze the 3D data.

\textbf{Point cloud understanding tasks} can be broadly categorized into three main areas. \textit{Point cloud classification} ~\cite{kanezaki2018rotationnet, yu2018multi, su2015multi, maturana2015voxnet, wu20153d, qi2017pointnet} involves assigning a label to an entire point cloud to identify the object it represents. This task requires models to understand the global structure of the 3D data. \textit{Point cloud segmentation}~\cite{yi2016scalable, sharma2020parsenet, armeni20163d} is the process of partitioning a point cloud into meaningful sub-regions, either by segmenting individual objects from a scene (instance segmentation) or by assigning a label to each point based on its local neighborhood (semantic segmentation). \textit{Point cloud object detection}~\cite{shi2020pv, shi2020points, yan2018second, yang2018pixor, liu2021group, qi2019deep} extends these tasks by identifying objects within a scene and providing their spatial locations and orientations. This involves detecting bounding boxes or other geometric shapes that enclose the objects of interest.

To enhance the performance of these tasks, researchers have explored various representation learning techniques. \textbf{Supervised representation learning for point clouds} relies on large annotated datasets to learn features that are discriminative for specific tasks. In supervised learning, models are trained with pairs of input data and corresponding labels, allowing them to learn directly from the annotations. This approach has achieved significant success, as supervised methods can leverage rich annotations to guide the learning process, resulting in highly accurate models. However, the dependency on extensive labeled data is a major limitation, as annotating 3D data is both time-consuming and expensive. This limitation makes it challenging to obtain the large-scale labeled datasets required for training robust models.

In contrast, \textbf{self-supervised representation learning for point clouds}~\cite{wang2020unsupervised, sauder2019self, yang2018foldingnet, poursaeed2020self, achlioptas2018learning, hassani2019unsupervised, li2018so, chen2021shape, yu2021point} aims to learn useful features without relying on labeled data. By designing pretext tasks that generate pseudo-labels from the data itself, self-supervised methods can leverage vast amounts of unlabeled point clouds to learn effective representations. These pretext tasks include point cloud reconstruction, context prediction, and transformation recognition, among others. Self-supervised learning has shown promise in reducing the need for labeled data and improving the generalization capabilities of learned representations. These methods enable models to leverage the abundant unlabeled 3D data available from various sources, such as LiDAR scans and depth sensors.

Despite the advances in both supervised and self-supervised learning, the availability of 3D data remains limited compared to 2D data. The collection and annotation of 3D data are challenging and resource-intensive, leading to smaller and less diverse datasets. To address this challenge, transferring knowledge from 2D to 3D has emerged as a promising direction. \textbf{Transfer-learning-based representation learning techniques} exploit the rich information present in 2D images to enhance 3D point cloud understanding. By leveraging pre-trained 2D models and transferring their knowledge to 3D tasks, it is possible to reduce the dependency on extensive 3D datasets and improve performance.

In this dissertation, we first focus on the point cloud supervised representation learning, specifically on the point cloud primitive segmentation task. Decomposing a 3D model into primitive surfaces is of fundamental importance with applications in reverse engineering, shape compression, shape understanding, shape editing, and robot learning. Primitive segmentation is the task of grouping and labeling points on an object based on primitive shape, and is fundamentally challenging due to the large search space and the fact that primitive patches may only approximately fit the object. We study how to develop deep learning models for primitive segmentation. In chapter~\ref{sec:hpnet}, we introduce a novel model called HPNet. It takes as input a point cloud (optionally including normals) and outputs a segmentation of the point cloud into primitives, with a type label for each primitive segment. The main idea of HPNet is to combine traditional tried-and-true geometric heuristics for primitive detection (for instance, algebraic relations between points and shape primitives, and segmentation from sharp edges) with a deep primitive detection approach based on powerful feature learning. We achieve this union through the use of a hybrid point descriptor that combines a learned semantic descriptor and two spectral descriptors. Experimental results on benchmark show that HPNet leads to significant performance gains from baseline approaches.

In the second part of this dissertation, we further study the self-supervised learning methods for point cloud representation learning. We mainly focus on the reconstruction-based methods. We first study the existing problem of point cloud autoencoder. Point clouds are noisy, discretized, and unstructured representations of 3D geometry. A 3D shape can be represented by many different point clouds, all of which are valid representations of the same object. Different point cloud samples are subject to different noises, which are induced from various sources such as the intrinsic noises from the sensors and the interference from the environment. The unordered and noisy nature distinguishes point clouds from conventional structured data, such as pixel images defined on rectangular grids.  When training a point-cloud autoencoder, the encoder is forced to capture sampling variations, limiting the model's ability to extract valuable information about the true 3D geometry.
To solve these problems, we introduce an asymmetric point-cloud autoencoder scheme, where the encoder takes a point cloud as input, and the decoder uses the implicit function as the output 3D representation. Reconstruction under the implicit representation discourages latent space learning from being distracted by the imperfections brought by sampling variations and encourages the encoder to capture generalizable features from the true 3D geometry.

Then, we study the recent and increasingly popular self-supervised learning scheme known as the masked autoencoder (MAE). Initially introduced for natural language processing (NLP) and computer vision, MAEs have demonstrated significant success and have recently been applied to 3D self-supervised pretraining for point clouds. In the image domain, MAEs typically involve a pretext task that restores features at masked pixels, such as colors. However, existing 3D MAEs focus solely on reconstructing the missing geometry, specifically the location of the masked points. 
In contrast to these previous studies, we assert that recovering point locations is not essential and that restoring intrinsic point features is far superior. Therefore, we propose to bypass point position reconstruction and instead recover high-order features at masked points, such as surface normals and surface variations. This is achieved through a novel attention-based decoder that operates independently of the encoder design. 
To validate the effectiveness of our pretext task and decoder design, we evaluated various encoder structures for point cloud understanding. Our experiments demonstrate the advantages of our pretrained networks across a range of point cloud analysis tasks, highlighting the superiority of our approach in leveraging intrinsic point features for point cloud self-supervised learning.

The last part of this dissertation addresses the challenge of transferring knowledge from 2D to 3D. Our key idea is to leverage pre-trained 2D networks in the image domain to enhance the performance of 3D representation learning, effectively bridging the gap between 2D and 3D domains. To achieve this, we utilize multi-view representation as the connecting bridge. Unlike existing methods that rely on multi-view representations by either aggregating image-based analysis results or converting 2D feature maps into 3D for interpretation, our approach is fundamentally different. We employ a 3D network for feature extraction, which is trained with the additional support of pre-trained 2D models. This innovative strategy allows us to harness the strengths of 2D models to improve 3D understanding without merely transforming 2D data. We conducted extensive experiments to validate the effectiveness of our approach across a wide range of downstream tasks. The results demonstrate significant improvements in performance, showcasing the potential of our method to advance 3D representation learning by effectively integrating 2D knowledge.

In summary, the contributions of this dissertation are the following:

\begin{itemize}

\item A novel supervised representation learning method for point cloud primitive segmentation tasks.

\item A new asymmetric point-cloud autoencoder called Implicit AutoEncoder, which is more effective at capturing generalizable features from true 3D geometry than standard point-cloud autoencoders.

\item A novel masked autoencoding method for point cloud self-supervised representation learning that predicts intrinsic point features at masked points instead of their positions.

\item A pipeline to leverage pre-trained 2D image-based models to supervise 3D pre-training through perspective projection and hierarchical feature-based 2D knowledge transfer.

\end{itemize}

%% file: chapters/background.tex
\chapter{Related Work}
\label{sec:background}

\section{Point Cloud Understanding Task}

Point cloud understanding is vital for various applications, including autonomous driving, robotics, augmented reality, and geographic information systems (GIS). Understanding point clouds involves several tasks such as classification, segmentation, and object detection. The challenges in this field include handling the large volume of data, dealing with noise and occlusions, and developing efficient algorithms that can operate in real-time. Advances in machine learning and deep learning have significantly contributed to progress in point cloud understanding, enabling more accurate and robust analysis of 3D data.

\subsection{3D Object Classification}

3D object classification with point cloud data has advanced significantly in recent years. Early efforts adapted deep learning techniques from image processing, such as using multiple view images~\cite{kanezaki2018rotationnet, yu2018multi, su2015multi} or applying convolutions on 3D voxel grids~\cite{maturana2015voxnet, wu20153d}. Although extending convolution operations from 2D to 3D seems intuitive, performing convolutions on a point cloud is challenging due to the lack of a well-defined order of points. Qi et al.\cite{qi2017pointnet} addressed this by learning global features of point clouds using a symmetric function invariant to point order. Alternatively, other methods have proposed learning local features through convolutions or autoencoders\cite{yang2018foldingnet}. Some approaches jointly learn features from point clouds and multi-view projections, treat point clouds and views as sequences, or use unsupervised learning.

Recent works have shown competitive and compelling performances on standard datasets. For instance, the performance gap between state-of-the-art methods such as DGCNN~\cite{dgcnn:2019:tog} and PointCNN~\cite{pointcnn} is less than 1\% on the ModelNet40 dataset. The online leaderboard maintained by the authors of ModelNet40 shows that the accuracy of object classification is approaching perfection, with point cloud methods achieving 92\%. ModelNet40 is a widely used synthetic dataset comprising 40 classes, with \num{9832} training objects and \num{2468} test objects. In contrast, ScanObjectNN is a real-world dataset that includes approximately \num{15000} scanned objects from 15 classes. Recent works typically evaluate new methods on these two datasets.

\subsection{3D Object Detection}
Detecting 3D objects from point clouds is challenging due to their orderless, sparse, and irregular characteristics. Previous approaches can be broadly classified into two categories based on point representations: voxel-based methods~\cite{shi2020pv, shi2020points, yan2018second, yang2018pixor} and point-based methods~\cite{liu2021group, qi2019deep}. 

Voxel-based methods are primarily used in outdoor autonomous driving scenarios where objects are distributed on a large-scale 2D ground plane. These methods process sparse point clouds using efficient 3D sparse convolution, then project these 3D volumes onto 2D grids to detect bird’s eye view (BEV) bounding boxes using 2D ConvNets. 

Powered by the PointNet series~\cite{qi2017pointnet, qi2017pointnet++}, point-based methods are also widely used to predict 3D bounding boxes. Most existing methods follow a bottom-up approach, extracting point-wise features and grouping them to obtain object features. This pipeline has been highly successful for estimating 3D bounding boxes directly from cluttered and dense 3D scenes. However, due to the hand-crafted point sampling and computationally intensive grouping scheme used in PointNet++~\cite{qi2017pointnet++}, these methods are difficult to extend to large-scale point clouds. 

To address these limitations, we propose an efficient, fully convolutional bottom-up framework to detect 3D bounding boxes directly from dense 3D point clouds.

\subsection{3D Primitive Segmentation}
Primitive segmentation from 3D shapes has been studied considerably in the past. It is beyond the scope of this section to provide a comprehensive overview. We refer to~\cite{KaiserZB19} for a thorough survey.  

\paragraph{Non-deep learning based approaches.}

Traditional approaches leverage stochastic paradigms~\cite{fischler1981random,schnabel-2007-efficient,MOD3GP}, parameter spaces~\cite{RABBANI2007355} or clustering and segmentation techniques~\cite{yan2012variational,lafarge:hal-00759265,rtpsrd}. Among these early works, RANSAC~\cite{fischler1981random} and its variants~\cite{schnabel-2007-efficient,Li:2011:GF,MATAS2004837,10.1371/journal.pone.0117341} are the most widely used methods for primitive segmentation and fitting. RANSAC-based methods can estimate model parameters iteratively and showed state-of-the-art results. However, they also suffer from laborious parameter tuning for each primitive. They also do not fully utilize each shape's information (e.g., sharp edges) and prior knowledge about primitive segments, which can be incorporated via machine learning approaches.  

\paragraph{Deep learning based approaches.}
Several recent works~\cite{Sharma_2018_CVPR,abstractionTulsiani17,zou20173d, li2019supervised, SharmaLMKCM20} have studied how to develop deep learning models for primitive segmentation. \cite{zou20173d} and \cite{abstractionTulsiani17} proposed to detect cuboids as rough abstractions of the input shapes. However, their performance is limited on other types of primitives. CSGNet~\cite{sharma2018csgnet} can handle more variety of primitives, but it requires a labeled hierarchical structure for the underlying primitives. This hierarchical structure is not always well-defined.  

SPFN~\cite{li2019supervised} and ParseNet~\cite{Sharma_2018_CVPR} are the most recent works. SPFN proposed a supervised primitive fitting method to predict per-point properties, including segmentation labels, type labels, and normals. Then they introduced a differential model estimation module to fit the primitive parameters. ParseNet proposed a complete model that can handle more primitives, including B-spline patches. However, in the segmentation part of their work, they mainly utilized semantic supervision, which ignored the importance of combining geometric features such as sharp edges.

\section{Point Cloud Network Architecture}

\paragraph{Point-based Network}
Point-based network models process each point independently using shared Multi-Layer Perceptrons and then aggregate a global feature through a symmetric function. Unlike typical deep learning methods for 2D images, these models cannot be directly applied to 3D point clouds due to their inherent irregularities. PointNet~\cite{qi2017pointnet} was a pioneering work in this field, taking point clouds as input and achieving permutation invariance using a symmetric function. It learns pointwise features independently with several MLP layers and extracts global features with a max-pooling layer.

Deep Sets~\cite{zaheer2017deep} also achieves permutation invariance by summing all representations and applying nonlinear transformations. However, since PointNet learns features independently for each point, it cannot capture the local structural information between points. To address this limitation, Qi et al.~\cite{qi2017pointnet++} proposed PointNet++, a hierarchical network designed to capture fine geometric structures from the neighborhood of each point.

\paragraph{Convolution-based Network}
Compared to kernels defined on 2D grid structures, such as images, designing convolutional kernels for 3D point clouds is challenging due to their irregularity. 3D discrete convolution methods define convolutional kernels on regular grids, where the weights for neighboring points are based on their offsets relative to the center point. For instance, PointCNN~\cite{li2018pointcnn} transforms the input points into a latent and potentially canonical order using an X-conv transformation (implemented through MLP), and then applies a typical convolutional operator on the transformed features.

On the other hand, 3D continuous convolution methods define convolutional kernels in a continuous space, where the weights for neighboring points are based on their spatial distribution relative to the center point. In PointConv~\cite{wu2019pointconv}, for example, convolution is defined as a Monte Carlo estimation of continuous 3D convolution using importance sampling. The convolutional kernels in PointConv consist of a weighting function (learned through MLP layers) and a density function (learned through kernelized density estimation and an MLP layer).

\paragraph{Graph-based Network}

Graph-based networks treat each point in a point cloud as a vertex of a graph and create directed edges based on the neighbors of each point. Feature learning is then conducted in either the spatial or spectral domains. Simonovsky et al.~\cite{simonovsky2017dynamic} were pioneers in this area, considering each point as a vertex and connecting each vertex to its neighbors with directed edges. They introduced Edge-Conditioned Convolution (ECC), which uses a filter-generating network (such as an MLP). Max pooling is employed to aggregate neighborhood information, and graph coarsening is implemented based on VoxelGrid~\cite{rusu20113d}. In DGCNN~\cite{wang2019dynamic}, the graph is constructed in the feature space and is dynamically updated after each layer of the network.

\paragraph{Transformer-based Network}

Transformers have recently become one of the most prevalent architectures across various fields. They leverage the multihead self-attention mechanism, which enables them to capture long-range dependencies between point patches and uncover implicit regional correlations. Transformer-based models, such as the one proposed by~\cite{yan20233d}, have achieved state-of-the-art performance in self-supervised learning (SSL) for point cloud classification and part segmentation. Additionally, the Point Cloud Transformer (PCT) introduced by~\cite{guo2021pct}, a variant specifically adapted for point clouds, enhances local feature extraction through techniques like farthest point sampling and nearest neighbor search, further improving performance in various downstream tasks.

\section{Point Cloud Self-Supervised Representation Learning}
\label{sec:relate-ssl}
To improve point cloud understanding tasks, it is crucial to design effective point cloud representation learning methods. 
Representation learning for point cloud data involves developing techniques to represent 3D data points in ways that capture the geometric structure and relevant features, facilitating various point cloud understanding tasks.
With the advent of deep neural networks, deep point cloud networks have become the optimal models for representation learning. These models are typically trained using large-scale, densely-labeled point cloud data. However, unsupervised point cloud representation learning, which aims to derive general and useful representations from unlabeled point cloud data, has gained increasing attention due to the challenges associated with large-scale point cloud labeling.
Self-Supervised Learning (SSL)~\cite{Chen2020,Grill2020,He2020,bao2022beit,HeCXLDG22,zhou2021ibot, zhuang2021unsupervised, zhuang2019self, Devlin2018,Brown2020}, an unsupervised training paradigm that extracts useful information from the data itself, is seen as a vital solution to the labor-intensive and time-consuming process of data labeling. SSL designs smart pretraining tasks to extract both basic geometric and advanced semantic information, which can then be transferred to downstream tasks through transfer learning. This approach mimics human learning by discovering objective principles through observation and summarization into a system of experience and knowledge.
Current point cloud SSL models can be broadly classified into two categories based on the nature of the pretext tasks: reconstruction-based methods and feature-space-based methods.

\paragraph{Reconstruction-based methods} Reconstruction-based methods~\cite{yu2021point, zhang2022point, Pang2022MaskedAF, sauder2019self} learn point cloud representations by reconstructing corrupted point clouds and recovering the original ones as accurately as possible. During the reconstruction process, both global features and the mappings between local and global areas are learned.
Recently, masked signal modeling has become a popular trend in point cloud SSL. Transformer-based architectures used for self-supervised learning (SSL) have demonstrated great simplicity and superior performance. PointBERT~\citep{Yu_2022_CVPR} and PointMAE~\citep{Pang2022MaskedAF} are two notable works following this idea. PointBERT partitions a point cloud into patches and trains a transformer-based autoencoder to recover the tokens of masked patches. In contrast, PointMAE directly reconstructs point patches without the need for costly tokenizer training, using Chamfer distance as the reconstruction loss.

\paragraph{Feature-Space-based methods} Contrastive learning is a popular approach in self-supervised learning (SSL) that encourages augmentations of the same input to have more similar representations in the feature space. The general strategy involves expanding the views of input point clouds (anchors) using various data augmentation techniques. Specifically, it aims to make positive samples (augmented from the same anchor) more similar to each other in the feature space than negative samples (augmented from different anchors)~\cite{xie2020pointcontrast, zhang2021self, Hou2021}.

\section{Transfer-Learning-Based Point Cloud Representation Learning}

Driven by the vast amounts of image data [10, 37, 53, 58], pre-training for improved visual representations has garnered significant attention in computer vision, benefiting a wide range of downstream tasks [7, 25, 36, 52, 59]. Beyond supervised pre-training with labels, many researchers have developed advanced self-supervised approaches to fully leverage raw image data through pre-text tasks, such as image-image contrast [5, 8, 9, 18, 24], language-image contrast [11, 50], and masked image modeling [3, 4, 16, 23, 28, 69]. Despite the popularity of 2D pretrained models, the community still lacks large-scale 3D datasets, mainly due to the high costs of data acquisition and labor-intensive annotation. 

One potential solution is to apply transfer-learning-based methods to bridge 3D pre-training with 2D models~\cite{afham2022crosspoint, xu2021image2point, dong2022autoencoders, zhang2022learning}. For instance, Dong et al.\cite{dong2022autoencoders} utilize pretrained 2D transformers as cross-modal teachers, while Zhang et al.\cite{zhang2022learning} leverage self-supervised pre-training and masked autoencoding to extract high-quality 3D features from 2D pretrained models. The aim is to harness the rich knowledge embedded in 2D pretrained models to enhance 3D point cloud representation learning.

%% file: chapters/pc_analysis.tex
\chapter{Point Cloud Supervised Representation Learning}\label{ch:pc-sl}

\renewcommand{\thefootnote}{\color{red}\fnsymbol{footnote}}
\footnote[6]{The content of this chapter includes my publication ``HPNet: Deep Primitive Segmentation Using Hybrid Representations''~\cite{yan2021hpnet} in International Conference on Computer Vision (ICCV) 2021. I am the first author of the publication.}

Point cloud supervised representation learning involves training models to understand and interpret 3D data represented by point clouds, which are sets of data points in space. 
In supervised learning, each point cloud or subset of points is labeled with a specific category, segmentation class, or other annotations, providing the ground truth needed for training.
The goal is to learn a mapping from the input point clouds to their corresponding labels through the use of neural networks, often employing architectures such as  PointNet. 
During training, the model learns to map the input point clouds to their corresponding labels by minimizing a loss function, often the cross-entropy loss for classification tasks or the mean squared error for regression tasks. This optimization process adjusts the model’s parameters to reduce prediction errors. Techniques like data augmentation (e.g., random rotations, scaling) and regularization (e.g., dropout) are commonly used to improve generalization and robustness. 
The learned representations from these models are crucial for various applications, such as 3D object recognition, where the goal is to identify and classify objects within a scene; 3D semantic segmentation, where each point in the point cloud is labeled with a semantic class; and 3D shape completion and reconstruction, where the model predicts missing parts of an object or scene.
Overall, point cloud supervised representation learning leverages labeled 3D data to train models that can accurately interpret and analyze complex 3D environments, enabling advancements in fields like robotics, autonomous driving, virtual reality, and more.
In this chapter, we study the 3D point cloud primitive segmentation task. Primitive segmentation is the task of grouping and labeling points on an object based on primitive shape, and is fundamentally challenging due to the large search space and the fact that primitive patches may only approximately fit the object.
We proposed a deep primitive segmentation network for point cloud data using hybrid representations. I am the main author of the presented works.

\input{chapters/HPNet/HPNet}

%% file: chapters/HPNet/HPNet.tex
\input{chapters/HPNet/macros}
\input{chapters/HPNet/symbols}

\section{HPNet: Deep Primitive Segmentation Using Hybrid Representations}\label{sec:hpnet}

\renewcommand{\thefootnote}{\color{red}\fnsymbol{footnote}}
\footnote[6]{The content of this section is based on my publication ``HPNet: Deep Primitive Segmentation Using Hybrid Representations''~\cite{yan2021hpnet} in International Conference on Computer Vision (ICCV) 2021. I am the first author of the publication.}

This section introduces HPNet, a novel deep-learning approach for segmenting a 3D shape represented as a point cloud into primitive patches. The key to deep primitive segmentation is learning a feature representation that can separate points of different primitives. Unlike utilizing a single feature representation, HPNet leverages hybrid representations that combine one learned semantic descriptor, two spectral descriptors derived from predicted geometric parameters, as well as an adjacency matrix that encodes sharp edges. Moreover, instead of merely concatenating the descriptors, HPNet optimally combines hybrid representations by learning combination weights. This weighting module builds on the entropy of input features. The output primitive segmentation is obtained from a mean-shift clustering module. Experimental results on benchmark datasets ANSI and ABCParts show that HPNet leads to significant performance gains from baseline approaches.

\input{chapters/HPNet/sections/01_intro}
\input{chapters/HPNet/sections/02_related}
\input{chapters/HPNet/sections/03_overview}
\input{chapters/HPNet/sections/04_approach}
\input{chapters/HPNet/sections/05_results}

\input{chapters/HPNet/sections/06_conclusions}

%% file: chapters/HPNet/macros.tex
\newcommand{\final}{0}

\def\etal{\emph{et al}.}

\definecolor{SithColor}{rgb}{0.7,0,0} 
\definecolor{SimingColor}{rgb}{0,0.7,0}
\definecolor{HaitaoColor}{rgb}{0,0,0.7}
\newcommand{\chongyang}[1]{{\color{SithColor} Chongyang: #1}}
\newcommand{\siming}[1]{{\color{SimingColor} Siming: #1}}
\newcommand{\haitao}[1]{{\color{HaitaoColor} Haitao: #1}}
\newcommand{\qixing}[1]{{\color{red} Q: #1}}
\newcommand{\warning}[1]{{\it\color{red} #1}}
\newcommand{\note}[1]{{\it\color{blue} #1}}
\newcommand{\nothing}[1]{}

\newcommand{\psdraftboxDefault}{\psnodraftbox}

\ifthenelse{\equal{\final}{1}}
{
\renewcommand{\chongyang}[1]{}
\renewcommand{\siming}[1]{}
\renewcommand{\haitao}[1]{}
\renewcommand{\warning}[1]{}
\renewcommand{\note}[1]{}
}
{}

\newcommand{\filename}[1]{\url{#1}}
\newcommand{\foldername}[1]{\url{#1}}

\newcommand{\inputsection}[1]{\input{sections/#1}}
\newcommand{\inputfigure}[1]{\input{figs/#1}}

\hyphenpenalty=1000 

%% file: chapters/HPNet/symbols.tex
\let \bs = \boldsymbol
\let \set = \mathcal

\newcommand{\pos}{\mathbf{p}}
\newcommand{\vertex}{\bs{v}}
\newcommand{\cons}{\textup{c}}
\newcommand{\diff}{\textup{diff}}
\newcommand{\plane}{\textup{Plane}}
\newcommand{\sphere}{\textup{Sphere}}
\newcommand{\cone}{\textup{Cone}}
\newcommand{\type}{\textup{type}}
\newcommand{\emb}{\textup{emb}}
\newcommand{\param}{\textup{param}}
\newcommand{\pull}{\textup{pull}}
\newcommand{\push}{\textup{push}}
\newcommand{\cylinder}{\textup{Cylinder}}
\newcommand{\bsplineo}{\textup{B-spline-Open}}
\newcommand{\bsplinec}{\textup{B-spline-Closed}}
\newcommand{\graph}{\mathcal{G}}
\newcommand{\vertexset}{\mathcal{V}}
\newcommand{\edgeset}{\mathcal{E}}
\newcommand{\edgeweightset}{\mathcal{W}}
\newcommand{\edgeweight}{w}
\newcommand{\normalizededgeweight}{\hat{w}}

\newcommand{\loss}{\mathcal{L}}
\newcommand{\klloss}{\loss_{KL}}

\newcommand{\smooth}{\textup{s}}

\newcommand{\p}{\bs{p}}
\newcommand{\q}{\bs{q}}
\newcommand{\tn}{\bs{t}} 
\newcommand{\n}{\bs{n}} 
\newcommand{\om}{\bs{\omega}} 

\newcommand{\ef}{\bs{e}} 

\newcommand{\realnum}{\mathbb R}
\newcommand{\Q}{\bs{Q}}

\newcommand{\norm}[1]{\left\lVert#1\right\rVert}
\newcommand{\skewm}[1]{\left[#1\right]_{\times}}

\newcommand{\primitive}{p}

\newcommand{\params}{\theta}
\newcommand{\point}{\mathbf{p}}
\newcommand{\probability}{P}
\newcommand{\dataset}{\mathcal{D}}

\newtheorem{proposition}{\textbf{Proposition}}
\newtheorem{theorem}{\textbf{Theorem}}
\newtheorem{corollary}{\textbf{Corollary}}
\newtheorem{assumption}{\textbf{Assumption}}
\newtheorem{example}{\textbf{Example}}
\newtheorem{definition}{\textbf{Definition}}
\newtheorem{remark}{\textbf{Remark}}

%% file: chapters/HPNet/sections/01_intro.tex
\subsection{Introduction}
\label{Section:Introduction}

The geometry of man-made objects can be frequently analysed in terms of primitive surface patches (planes, spheres, cylinders, cones, and other simple parametric surfaces). Decomposing a 3D model into primitive surfaces is of fundamental importance with applications in reverse engineering, shape compression, shape understanding, shape editing, and robot learning. \emph{Primitive segmentation} is the task of grouping and labeling points on an object based on primitive shape, and is fundamentally challenging due to the large search space and the fact that primitive patches may only approximately fit the object.

\begin{figure}
\centering
\includegraphics[width=0.8\textwidth]{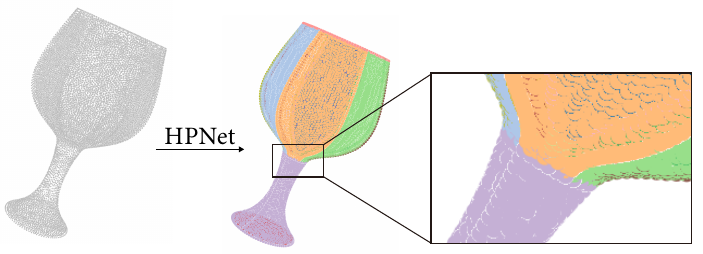}
\caption{HPNet takes a point cloud as input and outputs detected primitive patches. It can handle diverse primitives at different scales. The detected primitives have smooth boundaries.}
\label{Figure:Teaser}
\vspace{-0.1in}
\end{figure}

This section introduces a deep learning model called HPNet. It takes as input a point cloud (optionally including normals) and outputs a segmentation of the point cloud into primitives, with a type label for each primitive segment (see Figure~\ref{Figure:Teaser}). The main idea of HPNet is to combine traditional tried-and-true geometric heuristics for primitive detection (for instance, algebraic relations between points and shape primitives, and segmentation from sharp edges) with a deep primitive detection approach based on powerful feature learning. We achieve this union through the use of a hybrid point descriptor that combines a learned semantic descriptor and two spectral descriptors. The first spectral descriptor is derived from the adjacency matrix between the input points and predicted primitive parameters. The second spectral descriptor is built based on the adjacency matrix that models sharp edges. In both cases, the spectral approach unifies all primitive segmentation cues as point descriptors. It also rectifies the inputs to the spectral modules, e.g., incorrect predictions of primitive parameters. Given the point descriptors, HPNet employs a mean-shift clustering module to perform primitive segmentation. We also present an effective approach to train HPNet.

A key insight that allows HPNet to succeed on a wide diversity of input shapes is that different types of features have different power on different types of models. For example, when planar shapes are prominent, spectral descriptors are very effective, as eigenvectors of edge-aware adjacency operators localize strongly on the planar patches and segment them robustly.
In contrast, the semantic descriptors are more useful for complex curved shapes, such as patches of cones and cylinders, in which local curvature information is sufficient to determine the shape parameters. HPNet exploits this insight by using adaptive weights to combine different types of features. A weighting module automatically computes the relative weighting for different types of descriptors, based on an entropy measure of each descriptor which estimates the importance of that descriptor for clustering.  


We evaluate HPNet on two benchmark datasets, i.e., ANSI~\cite{li2019supervised} and ABCParts~\cite{Sharma_2018_CVPR}. With only points as input, HPNet improves the Segmentation Mean IoU (MIoU) score from 80.9/75.0 to 91.3/78.1 on the ANSI and ABCParts benchmarks, respectively. With both positions and normals as input, HPNet improves the MIoU score from 88.6/82.1 to 94.2/85.2. We also perform an ablation study to justify the effectiveness of different components of HPNet.

%% file: chapters/HPNet/sections/02_related.tex
\subsection{Related Work}
\label{Section:Related:Works}

Primitive segmentation from 3D shapes has been studied considerably in the past. It is beyond the scope of this paper to provide a comprehensive overview. We refer to~\cite{KaiserZB19} for a thorough survey.  

\paragraph{Non-deep learning based approaches.}
Traditional approaches leverage stochastic paradigms~\cite{fischler1981random,schnabel-2007-efficient,MOD3GP}, parameter spaces~\cite{RABBANI2007355} or clustering and segmentation techniques~\cite{yan2012variational,lafarge:hal-00759265,rtpsrd}. Among these early works, RANSAC~\cite{fischler1981random} and its variants~\cite{schnabel-2007-efficient,Li:2011:GF,MATAS2004837,10.1371/journal.pone.0117341} are the most widely used methods for primitive segmentation and fitting. RANSAC-based methods can estimate model parameters iteratively and showed state-of-the-art results. However, they also suffer from laborious parameter tuning for each primitive. They also do not fully utilize each shape's information (e.g., sharp edges) and prior knowledge about primitive segments, which can be incorporated via machine learning approaches.  

\paragraph{Deep learning based approaches.}
Several recent works~\cite{Sharma_2018_CVPR,abstractionTulsiani17,zou20173d, li2019supervised, SharmaLMKCM20, 8363617, yan2019recurrentfeedbackimprovesfeedforward, 10.1016/j.cag.2023.05.003} have studied how to develop deep learning models for primitive segmentation. \cite{zou20173d} and \cite{abstractionTulsiani17} proposed to detect cuboids as rough abstractions of the input shapes. However, their performance is limited on other types of primitives. CSGNet~\cite{sharma2018csgnet} can handle more variety of primitives, but it requires a labeled hierarchical structure for the underlying primitives. This hierarchical structure is not always well-defined.  

Our work is most relevant to SPFN~\cite{li2019supervised} and ParseNet~\cite{Sharma_2018_CVPR}. SPFN proposed a supervised primitive fitting method to predict per-point properties, including segmentation labels, type labels, and normals. Then they introduced a differential model estimation module to fit the primitive parameters. ParseNet proposed a complete model that can handle more primitives, including B-spline patches. However, in the segmentation part of their work, they mainly utilized semantic supervision, which ignored the importance of combining geometric features such as sharp edges. Another novelty of HPNet is that we design the prediction of per-point shape parameters and leverage spectral embedding to generate clearer dense point-wise descriptors.

\paragraph{Hybrid representations for 3D recognition.}
Our approach is motivated by recent methods that use hybrid geometric representations for solving 3D vision tasks. Examples include leveraging different types of keypoints for relative pose estimation~\cite{DBLP:conf/nips/GuibasHL19,yang2020extreme}, utilizing hybrid geometric primitives for 3D object detection and segmentation~\cite{DBLP:conf/cvpr/ZhangLWZH19,ZhangSYH20}, and geometric synthesis under hybrid representations~\cite{DBLP:journals/tog/ZhangYMLHVH20,PoursaeedFAK20,Yang_2021_ICCV}. Our work differentiates from prior methods by learning two spectral descriptors and a weighting sub-module that combines different geometric representations. In particular, the weighting sub-module models entropy for feature representations. This functionality is hard to achieve using alternative techniques, e.g., feature transformation networks. The weighting sub-module is also relevant to the line of work on feature selection~\cite{TangAL14,Liu:2007:CMFS,Li:2017:FS}. The resulting weights are derived from solving a quadratic program.   

\begin{figure*}
\includegraphics[width=\textwidth]{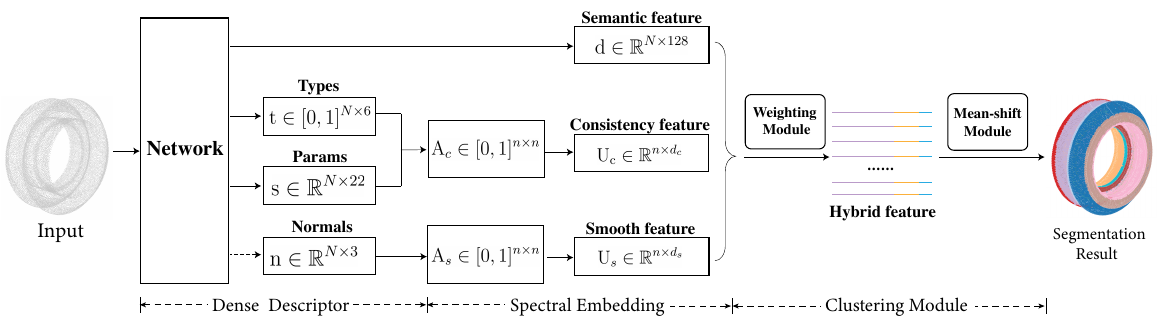}
\caption{Overview of our approach pipeline. HPNet consists of three modules: (1) Dense Descriptor takes a point cloud and optional normal as input and outputs a semantic feature descriptor, a type indicator vector, and a shape parameter vector. (2) Spectral Embedding Module takes dense descriptors as input and builds geometric consistency matrix $A_c$ and smoothness matrix $A_s$. Then it outputs consistency feature $U_c$ and smoothness feature $U_s$. (3) Clustering Module combines three features with adaptive weights and use mean-shift clustering to output the segmentation result.}
\label{Figure:Overview}
\end{figure*}


%% file: chapters/HPNet/sections/03_overview.tex
\subsection{Overview}
\label{Section:PS:AO}


\subsubsection{Problem Statement}
\label{Subsec:PS:AO}

We assume the input is given by a point cloud $\set{P} = \{p_i|1\leq i \leq n\}$. Each point $p_i$ has a position $\bs{p}_i\in \realnum^3$ and an optional normal $\bs{n}_i\in \realnum^3$. Our goal is to decompose $\set{P} = \set{P}_1\cup\cdots \cup \set{P}_{K}$ into $K$ primitives $\set{P}_k,1\leq k \leq K$. Each primitive has a type (i.e., plane, sphere, cylinder, cone, b-spline) and some type-specific shape parameters (normal; radius; control points; etc). We translate and scale each point cloud so that its mean is at the origin, and the diameter of the point cloud is $1$.




\subsubsection{Overview of HPNet}
\label{Subsec:Overview:StoPrimFit}

Figure~\ref{Figure:Overview} illustrates the pipeline of HPNet, which combines three types of modules: dense descriptor, spectral embedding, and clustering. 

\paragraph{Dense descriptor module.} The first module is trained to compute dense point-wise features. These features consist of a semantic descriptor trained to differentiate points from different primitives, a binary type vector identifying the primitive type each point likely belongs to, a parameter vector that predicts the shape parameters of the primitive fitting the point's neighborhood, and the point normal (in case normals were not provided in the input). Note that unlike standard approaches that either learn descriptors to separate different primitive points or detect shape parameters (c.f.~\cite{KaiserZB19,li2019supervised,SharmaLMKCM20}), HPNet combines both approaches. Incorporating both types of features promotes HPNet learning both semantic information (encoded in the type vectors) and local shape geometry (encoded in the shape parameters).

\paragraph{Spectral embedding modules.} The spectral embedding module converts relational cues useful for segmentation (algebraic relations between points that indicate consistency in surface shape parameters, or presence of sharp edges) into dense point-wise descriptors. Each type of relational cue is modeled using an adjacency graph with different weights. The resulting point descriptors are given by the leading eigenvectors of this adjacent matrix. We consider two relational cues that are complementary to the semantic point descriptors. The first module identifies sharp edges, which serve as boundaries between many pairs of primitive patches. Its adjacency matrix assigns high weights to neighboring vertices with similar normals. As the Euclidean distance in the spectral embedding space corresponds to the diffusion distance on the adjacency graph, points that fall on different sides of a sharp edge are far from each other in the embedding space, allowing robust identification of patches separated by creases.

The second module models the consistency between the input points and the predicted shape parameters.  It assigns high weight to an edge if one endpoint is consistent with the local geometry predicted by the other endpoint's shape parameters. The advantage of this formulation comes from the stability of leading eigenvectors against matrix perturbations. Even when a significant portion of the predicted primitive parameters is incorrect, the number of edges with wrong weights due to wrong primitive parameters is typically small, so that the correct predictions still yield an adjacency matrix where each primitive patch is a strongly connected sub-graph. The resulting descriptors are usually cleaner than the raw predicted shape parameters from the dense descriptor module. 

\paragraph{Clustering module.} The clustering module aggregates the point descriptors and applies mean-shift clustering to obtain the final primitive decomposition. A key observation is that the desired combination weights vary across different models. Instead of relying on hand-crafted weights, HPNet learns the combination weights by building on an entropy metric that assesses the underlying clustering structures in  high-dimensional point clouds.

\paragraph{Network training.} We present an effective strategy to train HPNet. The main idea, which has proven to be successful for keypoint-based 6D object pose estimation~\cite{PengLHZB19}, is to use a large-scale training set and a small-scale validation set. The training set is used in learning the dense descriptor module. In contrast, the validation set is used in learning the hybrid parameters of the spectral embedding and clustering modules. This approach alleviates over-fitting. 




%% file: chapters/HPNet/sections/04_approach.tex
\subsection{Our Method}
\label{Section:Approach}

This section introduces our approach in detail. Section~\ref{Subsec:Dense:Descriptor:Module} to Section~\ref{Subsec:Clustering:Module} present the three modules of HPNet. Section~\ref{Subsec:Network:Training} then introduces how to train HPNet in an end-to-end manner.

\subsubsection{Dense Descriptor Module}
\label{Subsec:Dense:Descriptor:Module}

As shown in Figure~\ref{Figure:Overview}, the first module of HPNet predicts dense point-wise attributes. The attributes associated with each point $p_i\in \set{P}$ includes a semantic feature descriptor $\bs{d}_i\in \realnum^{128}$, a binary type indicator vector $\bs{t}_i\in \{0,1\}^6$, and a shape parameter vector $\bs{s}_i\in \realnum^{22}$. HPNet considers six primitive types: \plane, \sphere, \cylinder, \cone, \bsplineo, and \bsplinec. The type indicator vector satisfies $t_i(j) = 1$ if and only if the index of the underlying primitive type of $p_i$ is $j$. $\bs{s}_i$ collects shape parameters for the \plane, \sphere, \cylinder, and \cone. HPNet uses the pre-trained SplineNet introduced in ~\cite{SharmaLMKCM20} to get the control points of open and closed b-spline patches. The prediction network is derived from DGCNN~\cite{wang2019dynamic} and Point Transformer~\cite{zhao2020point}, and the details are deferred to the supplemental material.

\subsubsection{Spectral Embedding Modules}
\label{Subsec:Spectral:Embedding:Module}

The spectral embedding modules take adjacency matrices of the input point cloud as inputs and output their leading eigenvectors. Each leading eigenvector is then a dense point descriptor function. HPNet considers two adjacency matrices. The first one models the consistency between the positions of the input points and the predicted shape parameters in a neighborhood around the point. The second one models presence of sharp edges. We provide the details for constructing these matrices in Sections~~\ref{Subsec:Consistency:Mat} and~\ref{Subsec:Smoothness:Mat} below.

\paragraph{Geometric Consistency Matrix}
\label{Subsec:Consistency:Mat}

We first define a distance metric $d(p_i,\bs{s}_j)$ between one point $p_i$ and the predicted shape parameter $\bs{s}_j$ associated with another point $p_j$~\cite{li2019supervised}. Let $t_j$ be the predicted primitive type of $s_j$. When $t_j\in \{\plane,\sphere,\cone,\cylinder\}$, we define $d(p_i, \bs{s}_j) = $
$$
\left\{
\begin{array}{cl}
|\bs{p}_i^T\bs{n}_j-d_j| & t_j = \plane \\
|\|\bs{p}_i-\bs{o}_j\|-r_j| & t_j = \sphere \\
|\|(I-\bs{a}_j\bs{a}_j^T)(\bs{p}_i-\bs{o}_j)\|-r_j| & t_j = \cylinder \\
{\scriptsize \|\bs{p}_i-\bs{o}_j\|\cos\left(\arccos\left(\bs{a}_j^T\frac{\bs{p}_i-\bs{o}_j}{\|\bs{p}_i-\bs{o}_j\|}\right)-\theta_j\right) } & t_j = \cone
\end{array}
\right.\
$$
where $\bs{n}_j$ and $d_j$ denote the normal and distance of a plane primitive; $\bs{o}_j$ and $r_j$ the center and radius of a sphere primitive; $\bs{a}_j$, $\bs{o}_j$, and $r_j$ the direction, center, and radius of a cylinder; and $\bs{o}_j$, $\bs{a}_j$, and $\theta_j$ the center, apex, and angle of a cone. There are $22$ total parameters.

When $t_j\in \{\bsplineo,\bsplinec\}$, we define $d(p_i, \bs{s}_j)$ as the closest distance between $p_i$ and the B-spline patch specified by $\bs{s}_j$. Please refer to the supplemental material for an efficient approach for computing $d(p_i, \bs{s}_j)$ approximately. 

We then define the signed weight between $p_i$ and $\bs{s}_j$ as
$$
w(p_i,\bs{s}_j):= \exp\left(-\frac{d^2(p_i,\bs{s}_j)}{2\sigma_{t_j}^2}\right),
$$
where $\sigma_{t_j} > 0$ is a hyperparameter associated with type $t_j$. Finally, we define the consistency adjacency matrix $A_{c}\in [0,1]^{n\times n}$, whose elements are given by
$$
A_{\cons}(i,j) = \big(w(p_i,\bs{s}_j)+w(p_j,\bs{s}_i)\big)/2, \quad 1\leq i,j \leq n.
$$
Let $\lambda_{\cons,i}$ and $\bs{u}_{\cons,i}$ denote the $i$-th eigenvalues and eigenvectors of $A_{\cons}$. We define the resulting descriptors as columns of $U_{\cons} = \big(\sqrt{\frac{\lambda_1}{\lambda_1}}\bs{u}_{\cons,1},\cdots, \sqrt{\frac{\lambda_1}{\lambda_{d_{\cons}}}}\bs{u}_{\cons,d_{\cons}}\big)\in \realnum^{n\times d_{\cons}}$. Depending on the number of primitives, $d_{\cons}$ varies across different shapes in our experiments. 

\paragraph{Discussion.}
Below we provide an analysis to show that the spectral descriptor is superior to the predicted primitive parameters. 
Denote $A_{\cons}^{\good}$ as the matrix $A_{\cons}$ with all entries $(i,j)$ set to zero when $p_i,p_j$ belong to different primitives. Without losing generality, we can reorder the vertices so that
$A_{\cons}^{\good} = \textup{diag}(A_{\cons,1}^{\good},\cdots,A_{\cons,K}^{\good})$ is a block diagonal matrix, where $A_{\cons,k}^{\good}$ contains the weights for pairs of points belonging to the $k$-th primitive. Decompose
$$
A_{\cons} = A_{\cons}^{\good} + E.
$$
To simplify the discussion, we further assume the weights are binary, i.e., $1$ for consistent pairs and $0$ for inconsistent pairs. We also assume $d_c = K$ in this analysis for convenience. 

Let $U_{\cons}^{\good}\in R^{n\times K}$ be the counterpart of $U_{\cons}$, whose columns are the re-scaled eigenvectors of $A_{\cons}^{\good}$. A variant of the Davis-Kahan theorem~\cite{yu2014useful} provides the difference between $U_{\cons}^{\good}$ and $U_{\cons}$:
\begin{equation}
\min\limits_{R\in O(K)}\|U_{\cons}R-U_{\cons}^{\good}\|_{\set{F}} \leq \frac{\sqrt{\lambda_{1}(A_{\cons}^{\good}})\|E\|_{\set{F}}}{\lambda_{K}(A_{\cons}^{\good})-\lambda_{K+1}(A_{\cons}^{\good})},
\label{Eq:DK}
\end{equation}
where $\|\cdot\|_{\set{F}}$ denotes the Frobenius norm. 

Applying (\ref{Eq:DK}), we argue that when primitive sizes are comparable, i.e., the size of the $k$-th primitive $n_k = \frac{n}{K}$, 
\begin{equation}
\min\limits_{R\in O(K)}\frac{\|U_{\cons}R-U_{\cons}^{\good}\|_{\set{F}}}{\|U_{\cons}^{\good}\|_{\set{F}}} = O\left({\scriptstyle \frac{2K\sqrt{\rho}}{(1-\rho)+\sqrt{1-\rho}}} n^{-\frac{1}{4}}\right),
\label{Eq:Spectral:Embedding:Error}
\end{equation}
where $\rho$ is the fraction of outliers of the predicted shape parameters. Moreover, it is easy to see that corresponding error of the predicted shape parameters scales as $O(\sqrt{\rho})$. It follows when $n$ is sufficiently large, the spectral descriptor is superior to the predicted shape parameters. Due to space constraint, we leave the proof to the supplemental material.

\vspace{-0.1in}
\paragraph{Smoothness Matrix}
\label{Subsec:Smoothness:Mat}

Consider a $k$-nearest neighbor graph $\set{G} = (\set{P},\set{E})$ whose vertices are taken from the input point cloud ($k = 50$ in this paper). For each edge $(i,j)\in \set{E}$, we define the corresponding weight as
$$
w_{(p_i,p_{j})} = \exp\left(-\frac{\|\bs{n}_i-\bs{n}_{j}\|^2}{2\sigma_{\edge}^2}\right),
$$
where $\sigma_{\edge}$ is a hyperparameter, and where $\bs{n}_i$ and $\bs{n}_{j}$ are the normals at $p_i$ and $p_{j}$, respectively.
Let $A_{\smooth}$ be the weighted adjacency matrix of $\set{G}$. The spectral descriptors $U_{\smooth}$ associated with $A_{\smooth}$ are then defined identically to $U_{\cons}$ (columns are scaled leading eigenvectors of $A_{\smooth}$).

\vspace{-0.1in}
\paragraph{Discussion.} The usefulness of $U_{\smooth}$ comes from the fact the Euclidean distance between $p_i$ and $p_j$ defined by the rows of $U_{\smooth}$ is identical to the diffusion distance on the weighted graph $\set{G}_{\smooth}$ specified by $A_{\smooth}$ (c.f.~\cite{Sun:2009:HKS}).
Note that points that are close to each other on the input model, but are on different sides of sharp edges, tend to have large diffusion distances (the paths connecting them have to detour around sharp edges). Therefore, these points have large distances in the embedding space.
\input{chapters/HPNet/tables/main_table.tex}
\subsubsection{Clustering Module}
\label{Subsec:Clustering:Module}

The clustering module concatenates the different types of point descriptors defined above, namely, the semantic descriptors $\bs{d}_i$ and the two spectral features specified by $U_{\cons}^{T}\bs{e}_i$ and $U_{\smooth}^T\bs{e}_i$. As mentioned in the introduction, the relative importance of the different descriptors varies dramatically based on the object geometry; so rather than using fixed weights to combine the descriptors, we introduce an approach that learns how to combine them.

\paragraph{Weighting sub-module.} To make the notations uncluttered, we describe our weighting scheme in a generic settings where there are $L$ features $F_l\in \realnum^{n\times m_l}, 1\leq l \leq L$, with $m_l$ the dimension of the $l$-th feature. Our goal is to compute a weight $w_l \in (0,1)$ for each feature, with $\sum_l w_l^2 = 1$. In the context of this paper, $L = 1 + d_{\cons} + d_{\smooth}$, i.e., $F_1$ corresponds the semantic descriptors $\bs{d}_i$, and each remaining $F_l$ corresponds to one spectral descriptor. Note that we weight each spectral descriptor individually because the desired number of spectral descriptors is dependent on the number of underlying primitives, which varies across different shapes.

HPNet uses the criterion that a feature $F_l$ should have large weights if $F_l$ reveals articulated cluster structure. Motivated from the feature selection for clustering approach described in~\cite{Dash:2000:FSC}, HPNet applies an entropy score to define the feature weight. Specifically, we first model the multivariate probability function of the feature space $F_l$ as
$$
P_{F_l}^{\sigma_l}(\bs{x}):= \frac{1}{n}(2\pi)^{-\frac{m_l}{2}}\sigma_l^{-m_l}\sum\limits_{i=1}^{n}\exp\left(-\frac{\|\bs{x}-F_l^T\bs{e}_i\|^2}{2\sigma_l^2}\right),
$$
where $\sigma_l$ is a hyperparameter associated with $F_l$. The entropy of $F_l$ is then
\begin{equation}
H_{\sigma_l}(F_l):= -\sum\limits_{i=1}^{n}P_{F_l}^{\sigma_l}(F_l^T\bs{e}_i)\log\Big(P_{F_l}^{\sigma_l}(F_l^T\bs{e}_i)\Big).
\label{Eq:Entropy:Fl}
\end{equation}
Intuitively, features where points form clusters (versus random point distributions) tend to have low entropy values. We model the weight of each feature $F_l$ so that it is inversely proportional to $H_{\sigma_l}(F_l)$:
\begin{equation}
w_l := \overline{w}_l/\sqrt{\sum\limits_{l}\overline{w}_l^2}, \quad \overline{w}_l:= \frac{1}{H_{\sigma_l}(F_l)}.    
\end{equation}


\paragraph{Mean-shift clustering sub-module.}
Since the primitive number varies between different models, we apply a mean-shift clustering procedure~\cite{comaniciu2002mean} to obtain the primitive segmentation result.

\input{chapters/HPNet/figures/figure2}

\subsubsection{Network Training}
\label{Subsec:Network:Training}

The network training of HPNet consists of two stages. The prediction module is trained in the first stage, while the hyperparameters of HPNet is learned in the second stage.

\paragraph{Training of the Prediction Module}

We train the prediction module using a training dataset, in which each point cloud has ground-truth primitive segmentation and associated primitive parameters. In the following, we focus on defining the loss for one point cloud $\set{P}$. The total loss consists of three terms:
\begin{equation}
\set{L}(\set{P}) = \set{L}_{\emb}(\set{P}) + \alpha \set{L}_{\type}(\set{P}) + \beta \set{L}_{\param}(\set{P}),
\end{equation}
where $\set{L}_{\emb}(\set{P})$ is the embedding loss that trains the semantic descriptor; $\set{L}_{\type}(\set{P})$ is the type loss that trains the type vector; $\set{L}_{\param}(\set{P})$ is the parameter loss that trains the primitive parameters. We set $\alpha = 1.0$ and $\beta = 0.1$ in our experiments. Network training employs ADAM~\cite{KingmaB14}. In the following, we define each loss term. 

\vspace{-5pt}
\paragraph{Embedding loss.} Similar to ~\cite{debrabandere2017semantic,yu2019single}, the embedding loss seeks to pull the semantic descriptors close to each other in the same primitive patch and push semantic descriptors of different primitive patches far from each other. Specifically, the loss consists of two terms: $\set{L}_{\pull}$ and $\set{L}_{\push}$. Denote $\set{P}_{k}^{\gt}$ as the ground-truth primitives. $\set{L}_{\pull}$ pulls each descriptor to the mean of the descriptors of the underlying primitive: 
\begin{equation}
\set{L}_{\pull} = \frac{1}{K}\sum^K_{k=1}\frac{1}{|\set{P}_k^{\gt}|}\sum_{p_i\in \set{P}_k^{\gt}}\max\left({\scriptstyle \left\|\bs{d}_i - \bs{d}_{\set{P}_k^{\gt}}\right\| - \delta_1, 0 } \right),
\label{Eq:L:pull}
\end{equation}
where $\bs{d}_{\set{P}_k^{\gt}} = \sum_{p_i\in \set{P}_k^{\gt}}\bs{d}_i/|\set{P}_k^{\gt}|$. $L_{\push}$ pushes the embedding centers away from each other:
\begin{equation}
\set{L}_{\push} = \frac{1}{K(K-1)}\sum_{k < k'}\max\left({\scriptstyle \delta_2 - \left\|\bs{d}_{\set{P}_k^{\gt}} - \bs{d}_{\set{P}_{k'}^{\gt}}\right\|, 0} \right).
\label{Eq:L:push}
\end{equation}
Combing (\ref{Eq:L:pull}) and (\ref{Eq:L:push}), we define the embedding loss as
$$
\set{L}_{\emb} = \lambda \set{L}_{\pull} + \nu \set{L}_{\push}.
$$
In our experiments, $\lambda=1$, $\nu=1$, $\delta_v = 0.5$, and $\delta_d = 1.5$.

\vspace{-10pt}
\paragraph{Type loss.} We employ the cross-entropy loss $H_{ce}$ to define the type loss $$
\set{L}_{\type} = \frac{1}{n}\sum_{i = 1}^n H_{ce}(\bs{t}_i, \bs{t}_i^{\gt}),
$$
where $\bs{t}_i^{\gt}$ is the ground-truth of $\bs{t}_i$.

\vspace{-5pt}
\paragraph{Parameter loss.} The parameter loss employs the standard L2 loss on $\mathbb{R}^{22}$ between the predicted shape parameter $\bs{s}_i$ and the underlying ground-truth $\bs{s}_i^{\gt}$:
$$
\set{L}_{\param} = \frac{1}{n}\sum_{i = 1}^n \|\bs{s}_i - \bs{s}_i^{\gt}\|^2.
$$

\paragraph{Learning of the HyperParameters}

The second stage optimizes the hyperparameters of HPNet, including those used in defining the spectral modules and those used in defining the entropy terms (Equation~\ref{Eq:Entropy:Fl}). Since the total number of parameters is small, we use the established finite-difference approach for hyperparameter optimization from Song et al.~\cite{SongSH20}. This is done by optimizing the embedding loss on a validation dataset. Given the current hyperparameters, we compute the numerical gradient by sampling neighboring hyperpachapters/HPNet/rameter configurations. The best-fitting linear function gives the numerical gradient. We then apply a backtracking line search to determine the step-size. This gradient descent procedure terminates when the step size is smaller than $10^{-3}$, which typically occurs within 10-30 iterations.


%% file: chapters/HPNet/tables/main_table.tex
\begin{landscape}
\begin{table*}
  \footnotesize
  \centering
  \begin{tabular}{l|c|c|c|c|c|c|c|c|c}
  \toprule
  &  & \multicolumn{4}{c|}{ANSI} & \multicolumn{4}{c}{ABCParts}\\
  \hline
  & Input & seg iou & type iou & res error & P coverage & seg iou & type iou & res error & P coverage \\
  \hline
  NN       & p   & 81.92 & 95.00 & 0.014 & 91.90 & 54.10 & 61.10 & - & - \\
  RANSAC~\cite{schnabel2007efficient}  & p+n & 70.10 & 93.13 & 0.029 & 78.79 & 67.21 & -     & 0.022 & 83.40 \\
  \hline
  SPFN~\cite{li2019supervised}     & p   & 77.61 & 95.43 & 0.014 & 92.10 & 58.15 & 73.88 & 0.023 & 87.58 \\
  SPFN~\cite{li2019supervised} & p+n & 88.05 & 98.10 & 0.011 & 92.94 & 73.41 & 80.04 & 0.020 & 89.40 \\
  ParseNet~\cite{SharmaLMKCM20} & p   & 80.91 & 97.49 & 0.013 & 90.91 & 75.01 & 81.16 & 0.014 & 87.95 \\
  ParseNet~\cite{SharmaLMKCM20} & p+n & 88.57 & 98.26 & 0.010 & 92.72 & 82.14  & 88.60 & 0.011 & 92.97 \\  
  \hline
  Ours-wc  & p   & 90.12 & 98.21 & 0.012 & 92.00 & 76.71 & 82.14 & 0.013 & 88.21 \\
  Ours-wc  & p+n & 92.41 & 98.87 & 0.010 & 93.58 & 83.20 & 89.54 & 0.010 & 93.15 \\
  Ours     & p & 91.34 & 98.66 & 0.011 & 93.02 & 78.12 & 85.32 & 0.012 & 90.54  \\
  Ours     & p+n & \textbf{94.15} & \textbf{98.90} & \textbf{0.008} & \textbf{94.02} & \textbf{85.24} & \textbf{91.04} & \textbf{0.009} & \textbf{94.31} \\
  \bottomrule
  \end{tabular}
{%
 \caption{Benchmark evaluation on our approach and baseline approaches. We provide different input to the model:points(p) and points+normals(p+n). Here, Ours-wc stands for our method without combining two spectral descriptors.}
 \label{tab:baseline-comparison}
}
\end{table*}
\end{landscape}

%% file: chapters/HPNet/figures/figure2.tex
\begin{landscape}
\begin{figure*}
\centering
\footnotesize
\def\imh{0.073\textwidth}
\def\imw{0.27\textwidth}
\newcommand{\TT}[1]{\raisebox{-0.5\height}{#1}}
\setlength{\tabcolsep}{1pt}
\begin{tabular}{cccccccc}

\rotatebox[origin=c]{90}{G.T} & 
\TT{\includegraphics[width=\imw]     {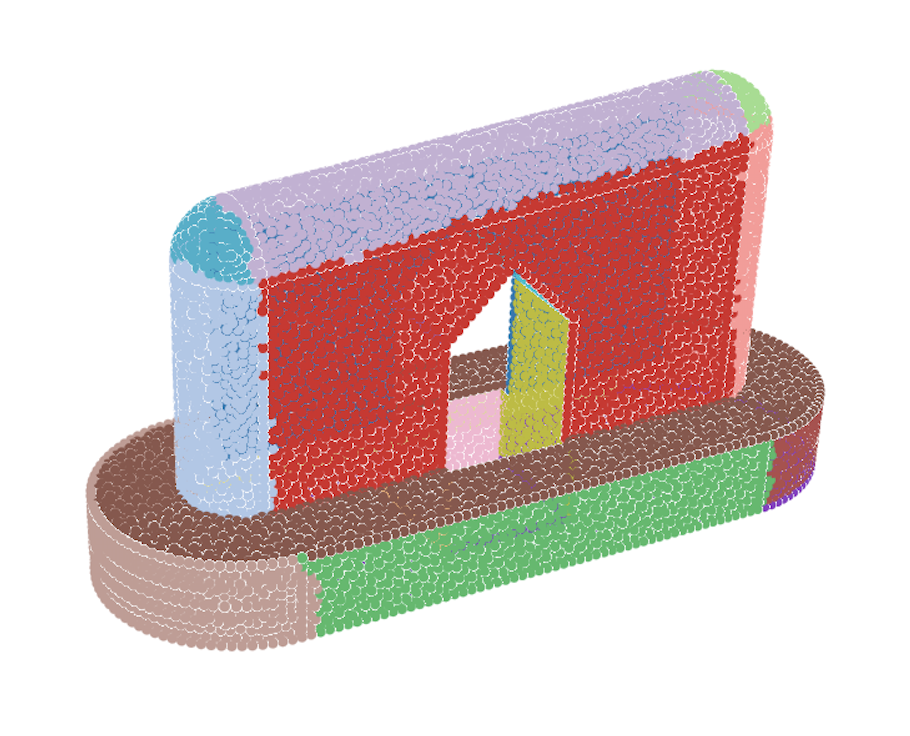}} & 
\TT{\includegraphics[width=\imw]   {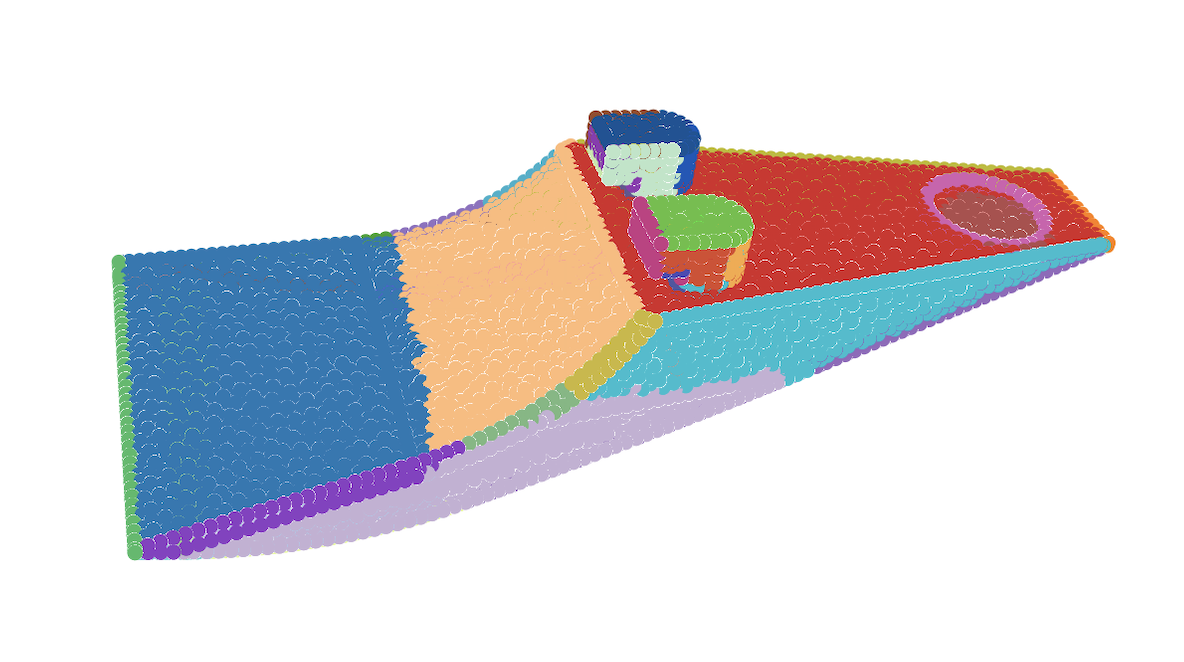}} & 
\TT{\includegraphics[height=0.11\textwidth]      {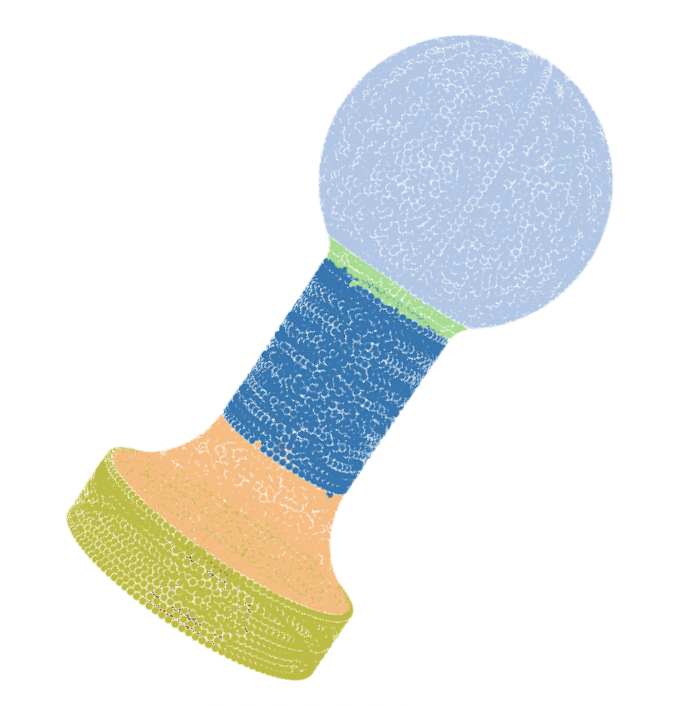}} & 
\TT{\includegraphics[width=\imw]    {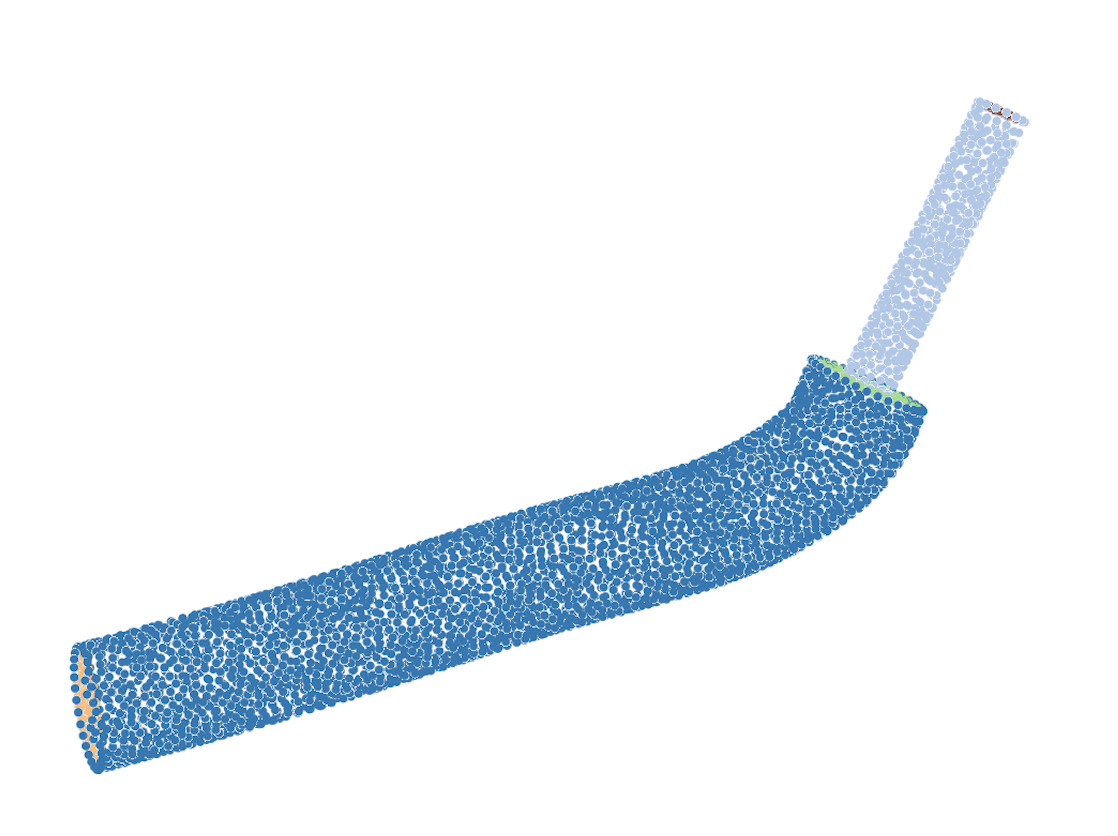}} & 
\TT{\includegraphics[width=0.12\textwidth]    {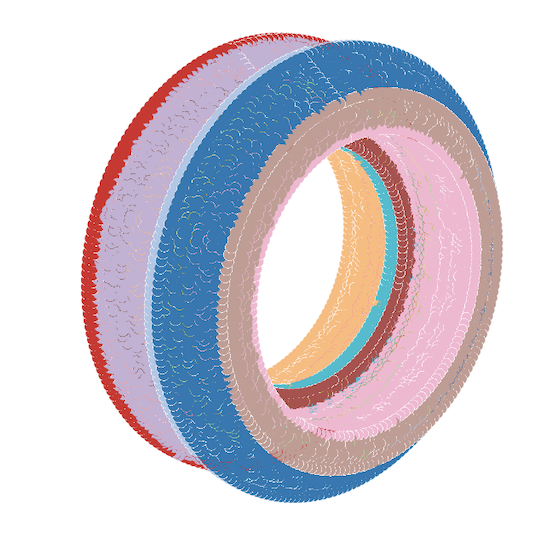}} & 
\TT{\includegraphics[width=0.13\textwidth]      {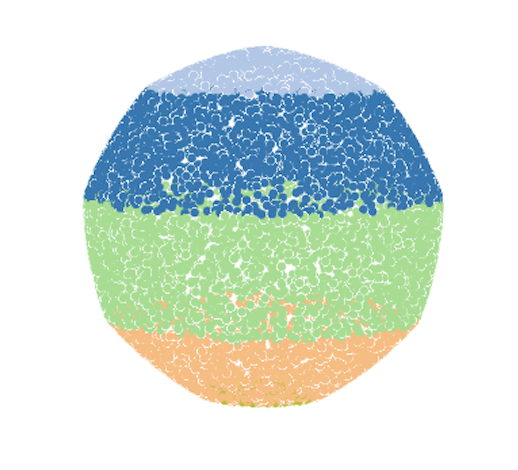}} &
\TT{\includegraphics[width=0.13\textwidth]      {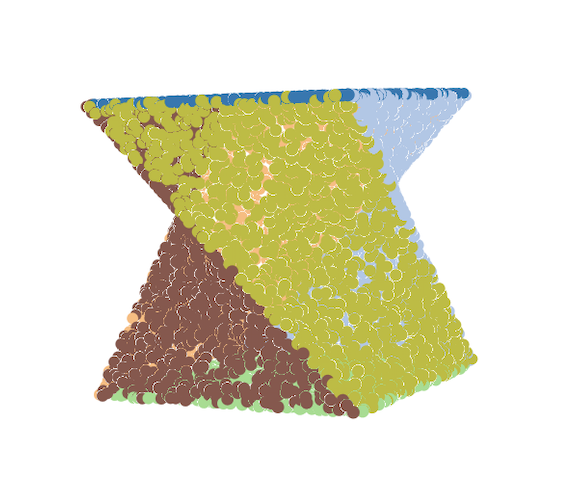}} \\
\hline
\rotatebox[origin=c]{90}{SPFN} & 
\TT{\includegraphics[width=\imw]     {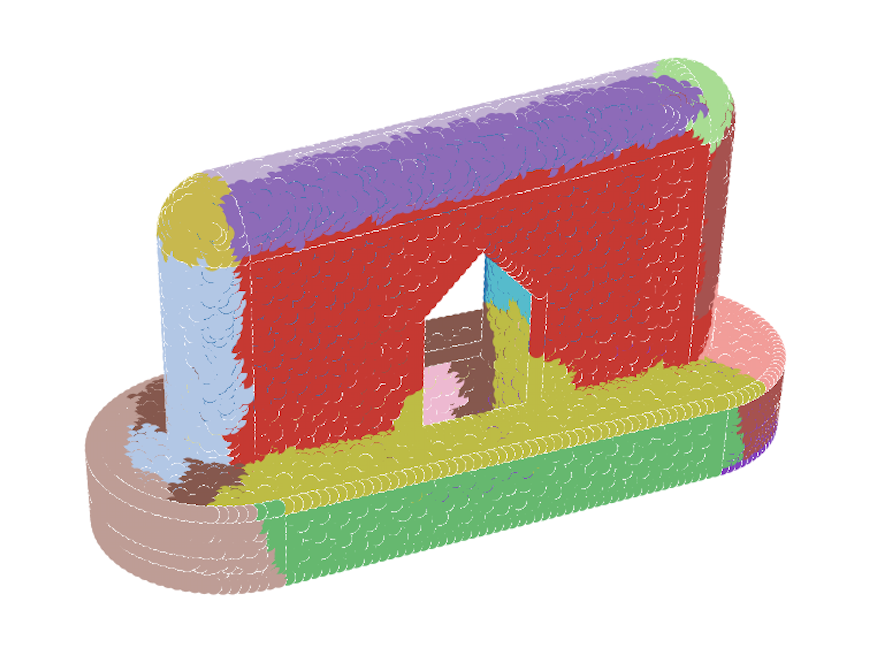}} &
\TT{\includegraphics[width=\imw]   {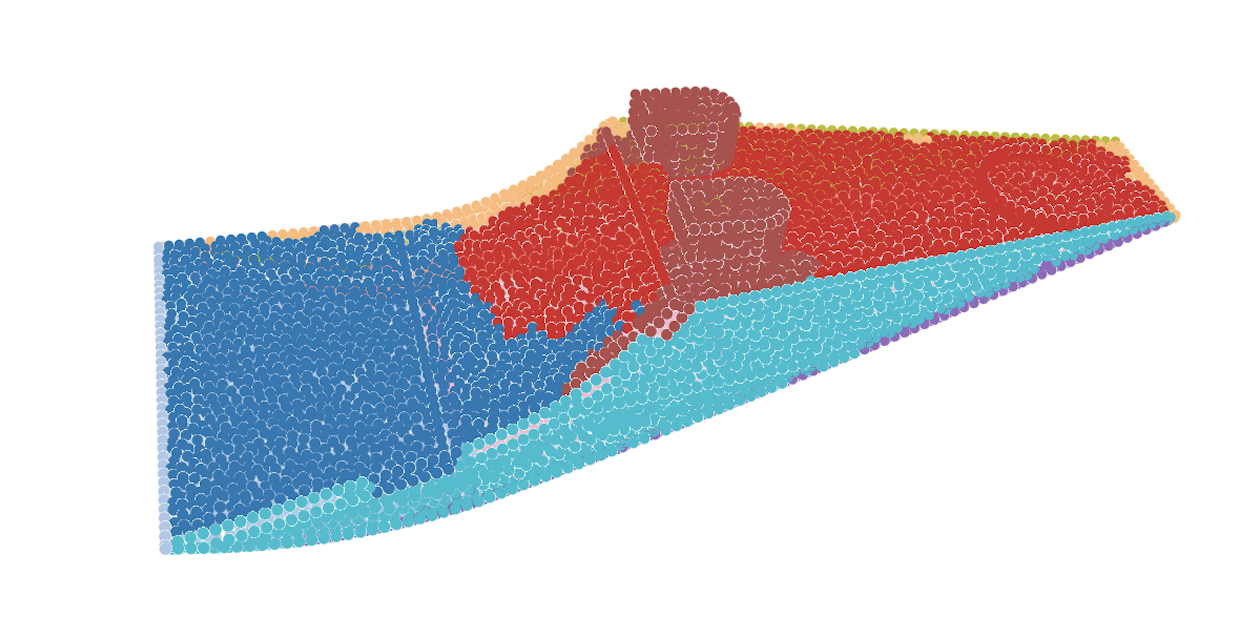}} & 
\TT{\includegraphics[height=0.11\textwidth]      {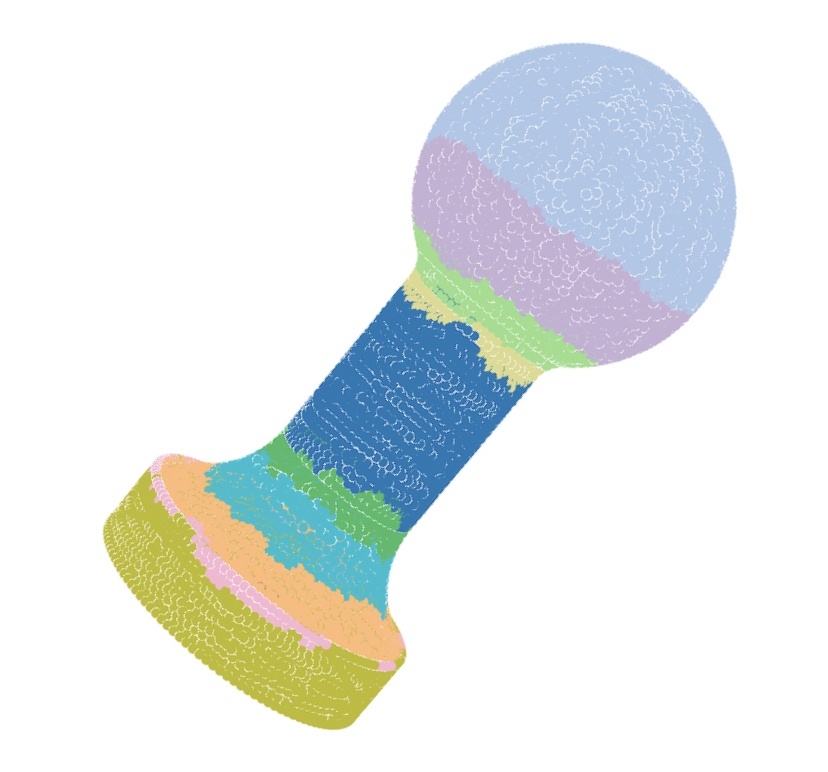}} & 
\TT{\includegraphics[width=\imw]    {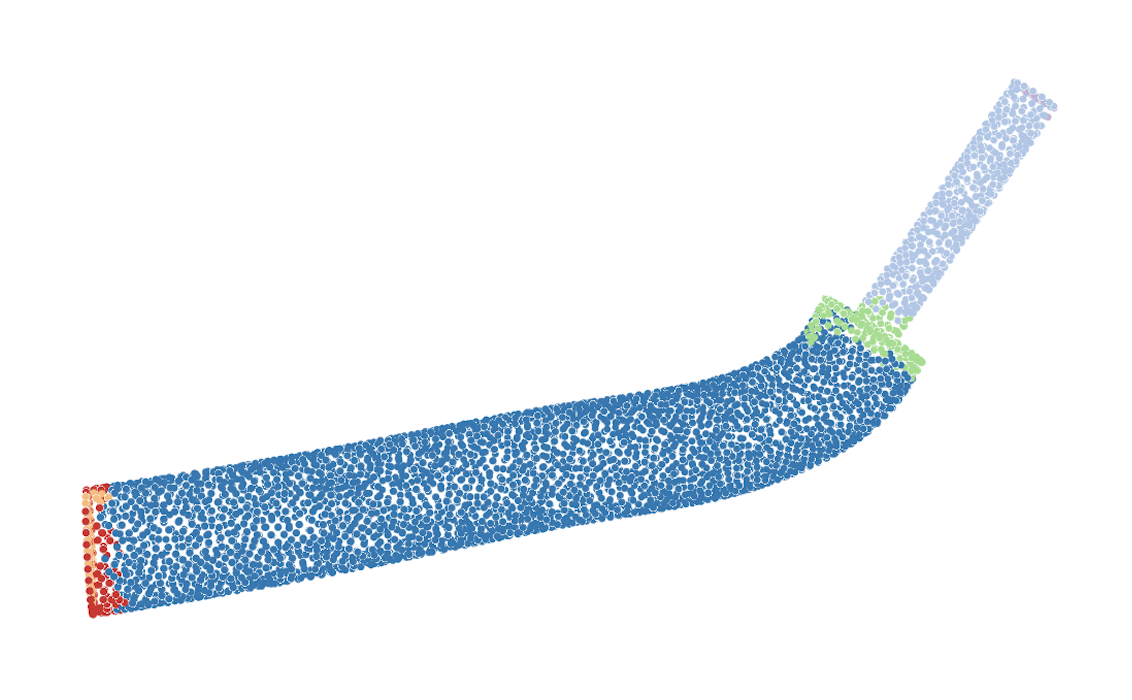}} & 
\TT{\includegraphics[width=0.115\textwidth]    {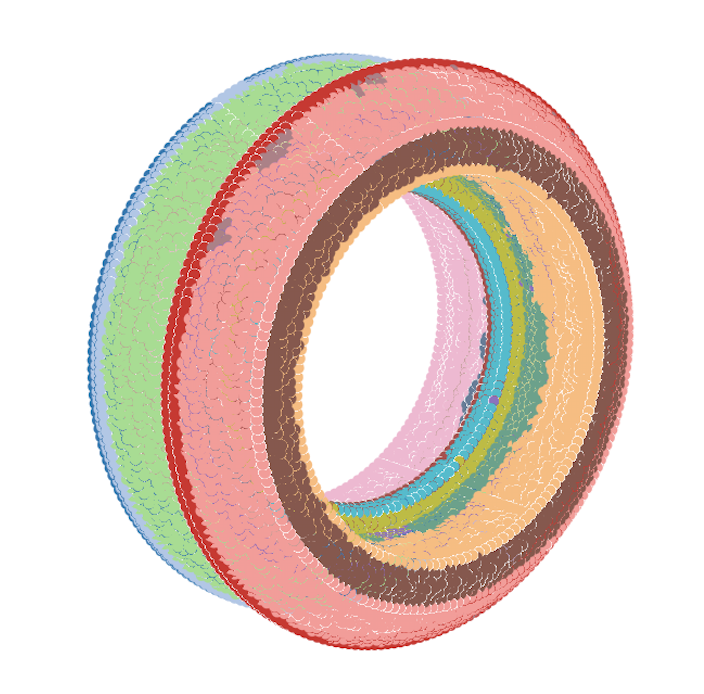}} &
\TT{\includegraphics[width=0.13\textwidth]      {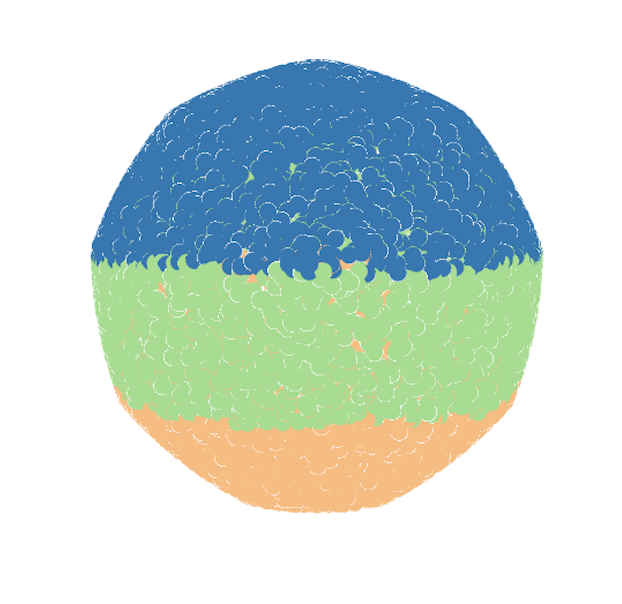}} &
\TT{\includegraphics[width=0.13\textwidth]      {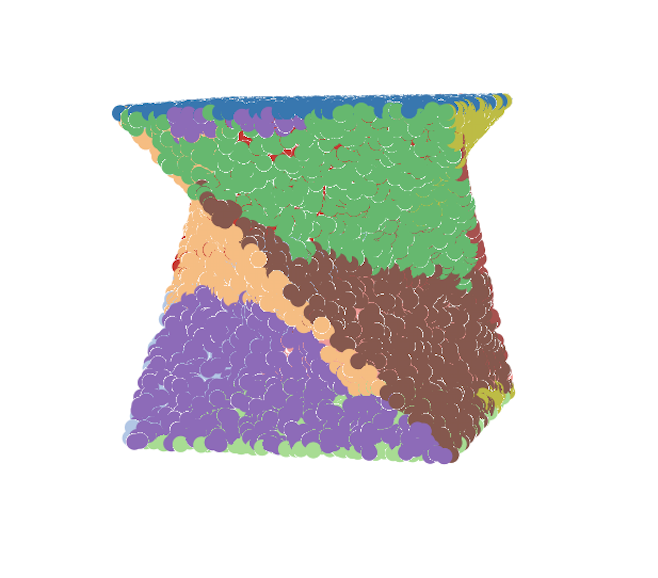}}\\

\rotatebox[origin=c]{90}{Parsenet} & 
\TT{\includegraphics[width=\imw]     {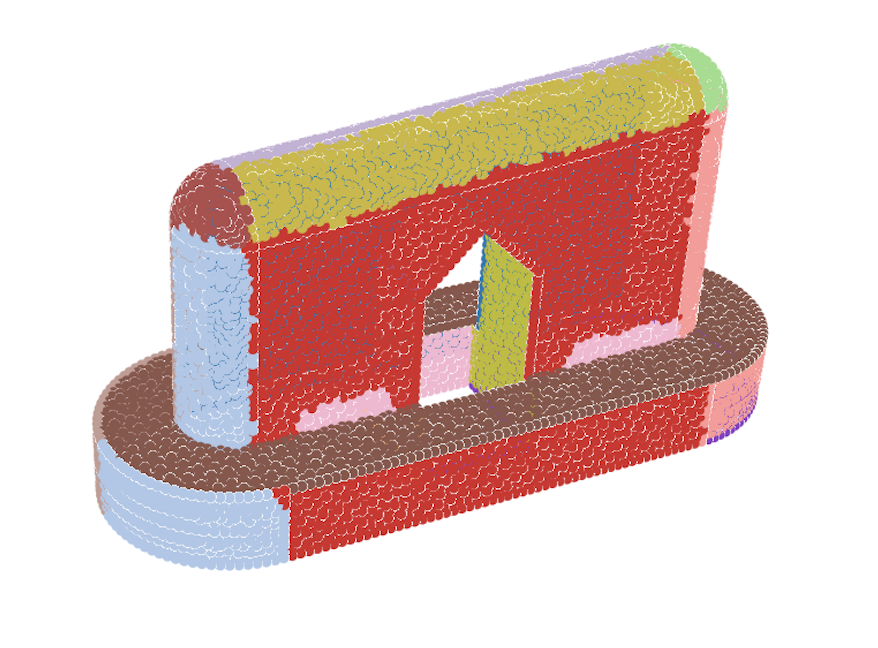}} & 
\TT{\includegraphics[width=\imw]   {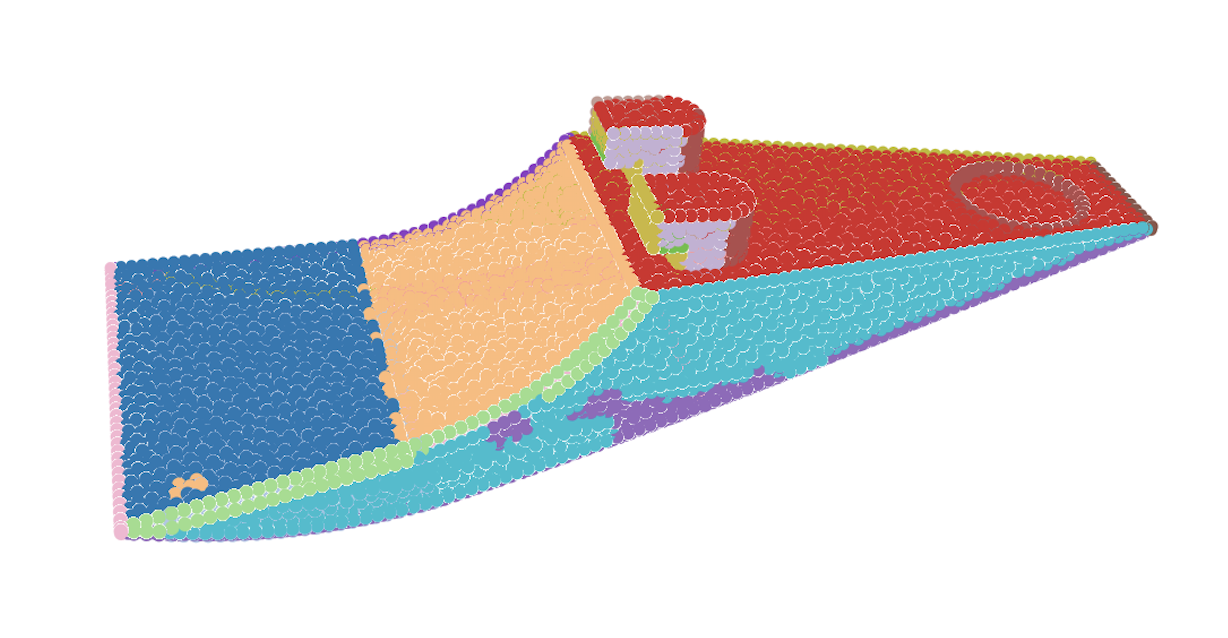}} & 
\TT{\includegraphics[height=0.11\textwidth]      {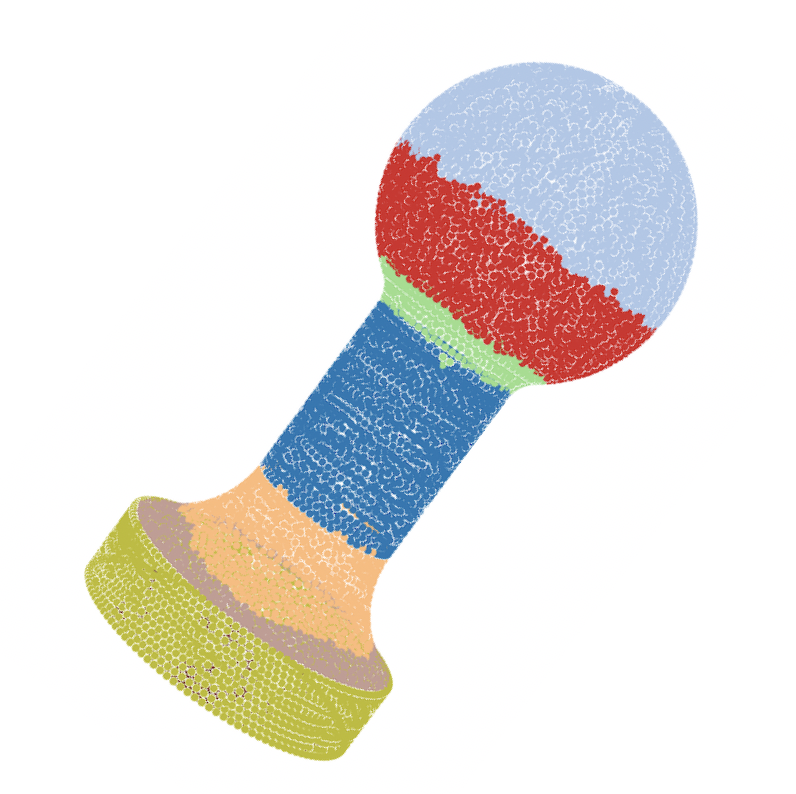}} & 
\TT{\includegraphics[width=\imw]    {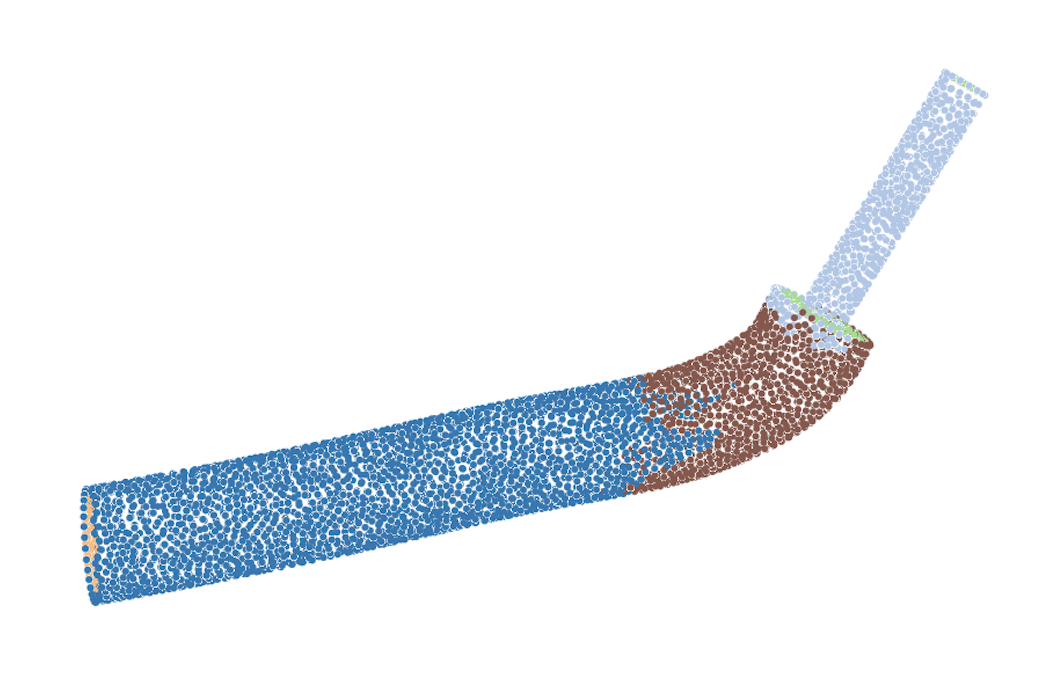}} & 
\TT{\includegraphics[width=0.125\textwidth]    {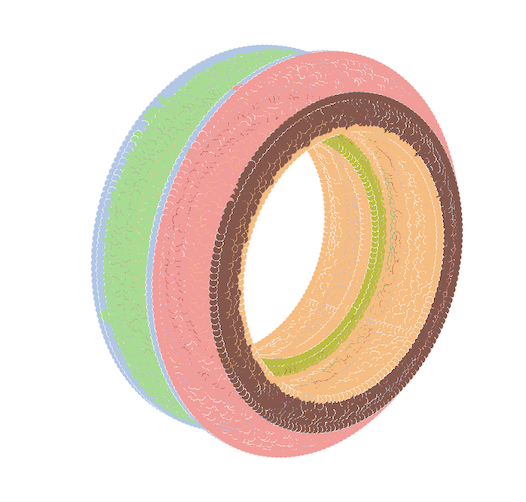}} &
\TT{\includegraphics[width=0.13\textwidth]      {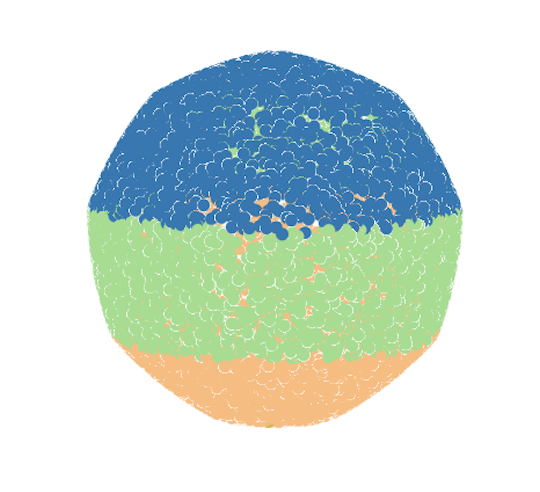}} &
\TT{\includegraphics[width=0.13\textwidth]      {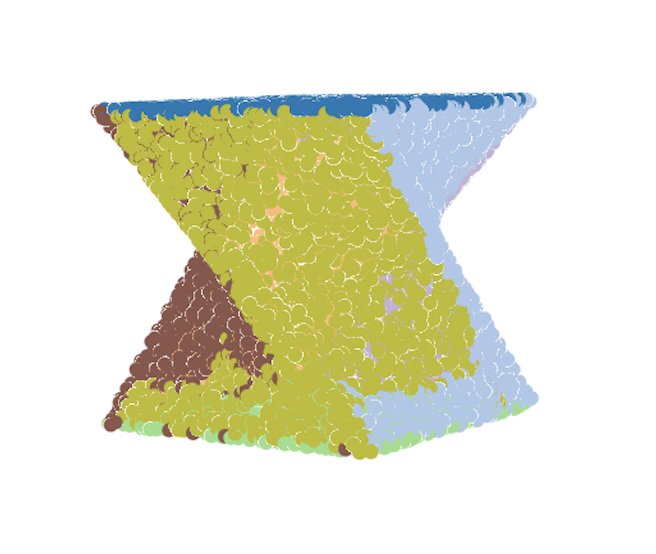}}\\

\rotatebox[origin=c]{90}{Ours} & 
\TT{\includegraphics[width=\imw]     {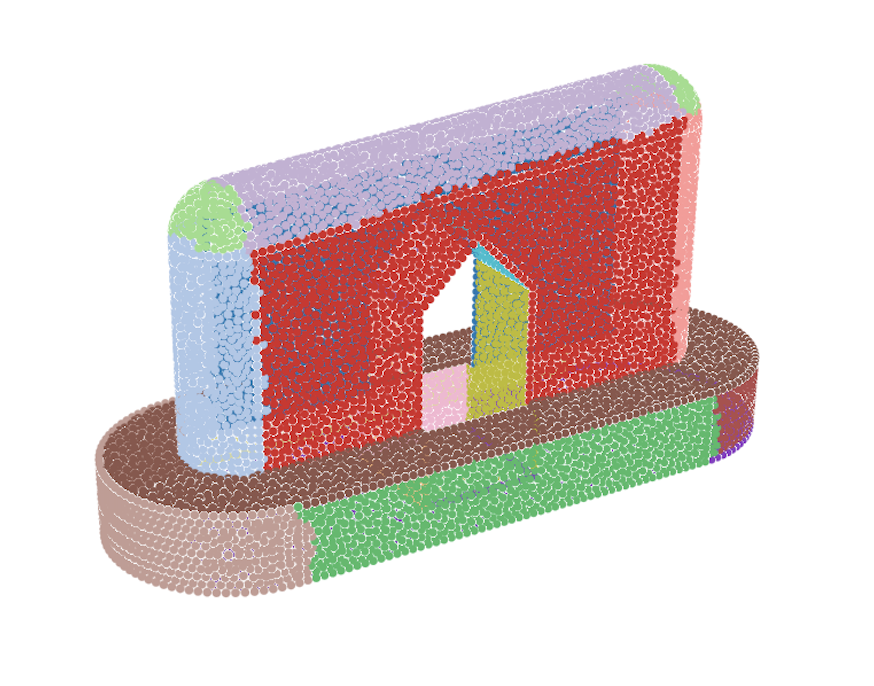}} & 
\TT{\includegraphics[width=\imw]   {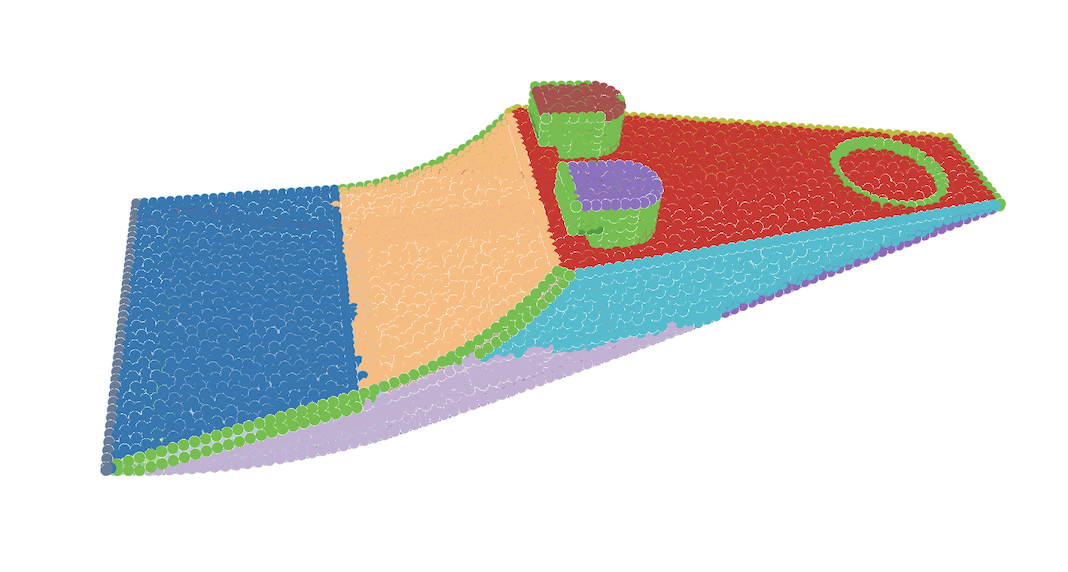}} & 
\TT{\includegraphics[height=0.11\textwidth]      {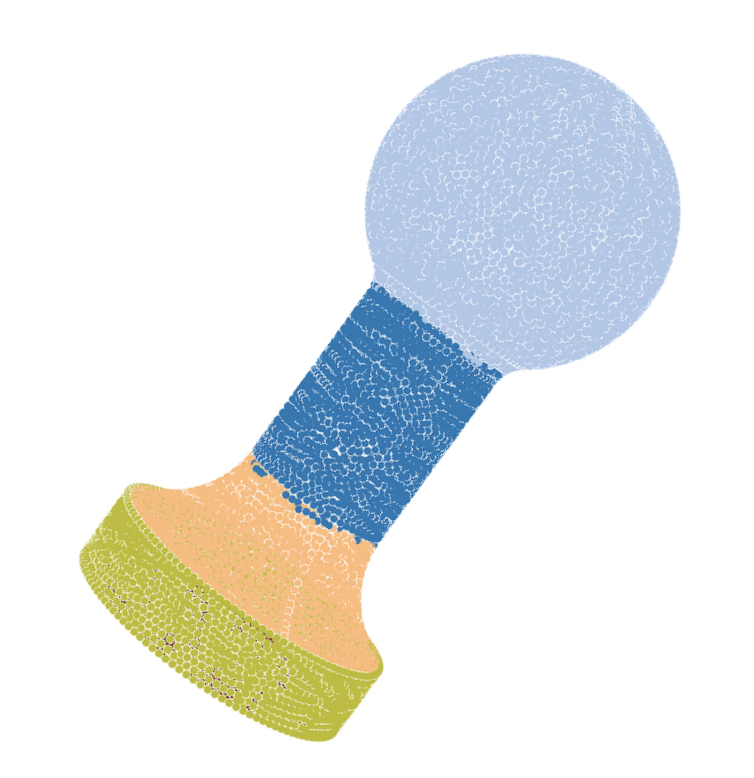}} & 
\TT{\includegraphics[width=\imw]    {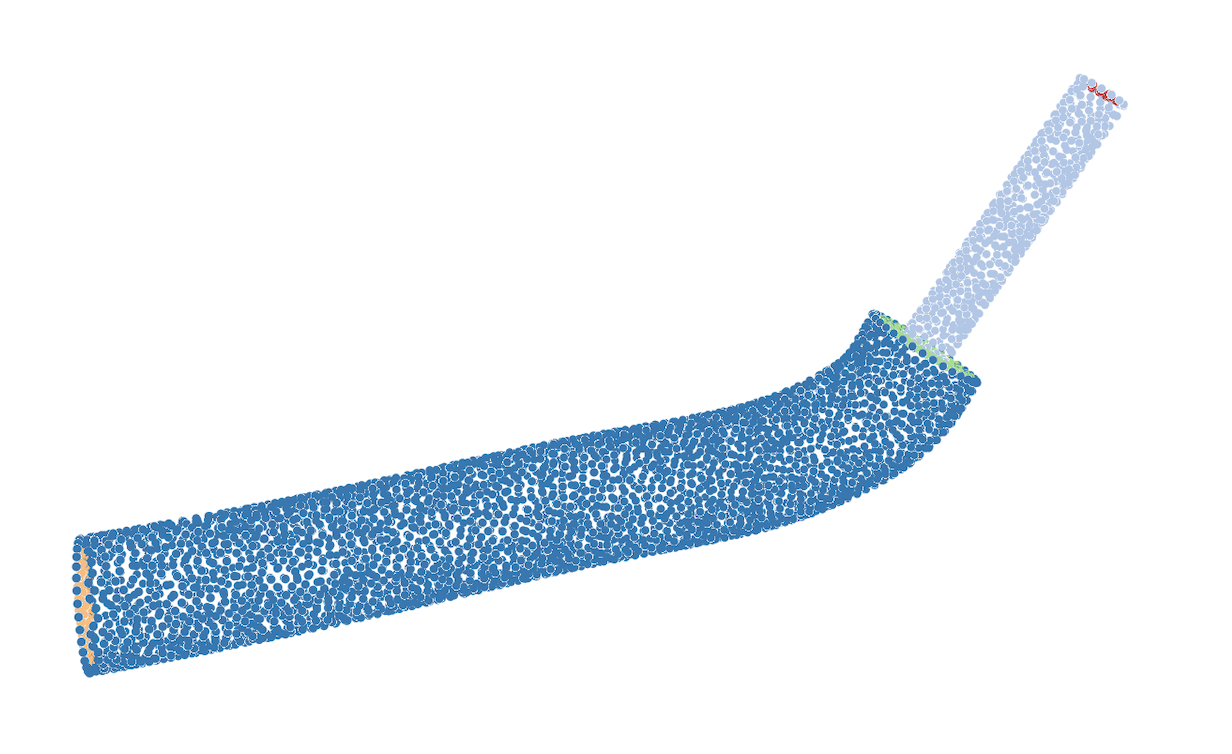}} & 
\TT{\includegraphics[width=0.125\textwidth]    {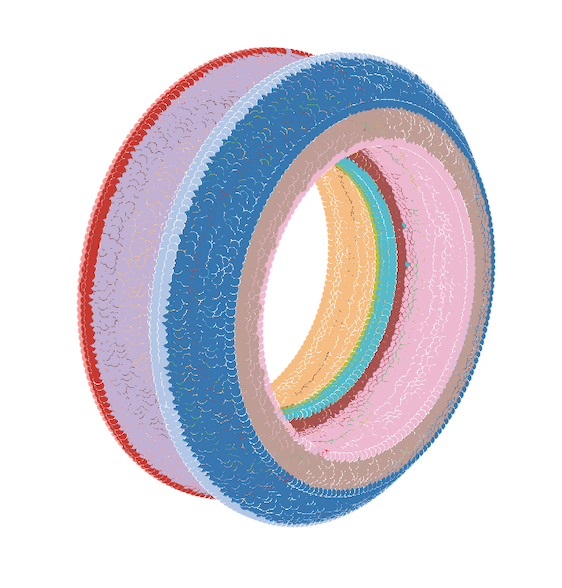}} & 
\TT{\includegraphics[width=0.13\textwidth]      {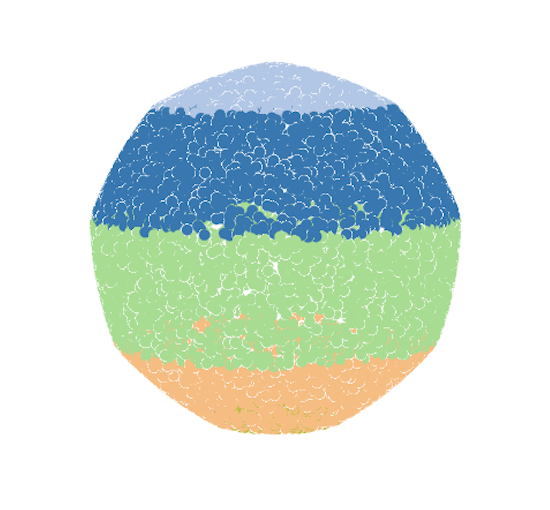}} &
\TT{\includegraphics[width=0.13\textwidth]      {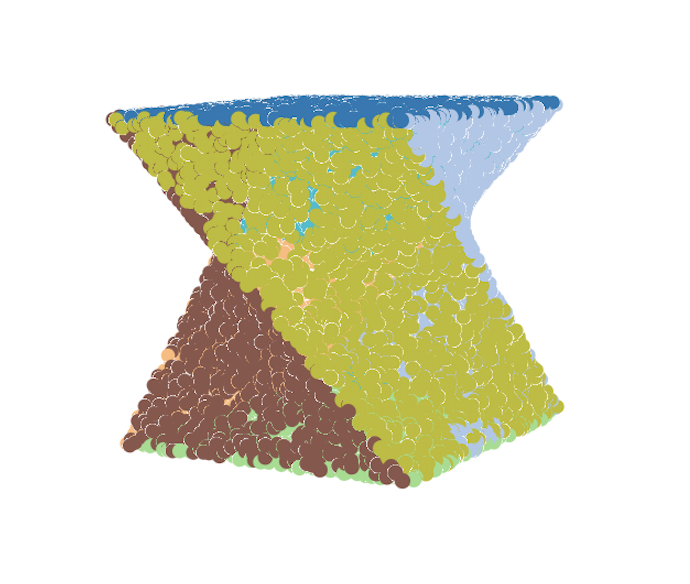}}\\

\end{tabular}
\caption{Primitive segmentation results with different methods. From top to down, we show the results of ground truth, SPFN~\cite{li2019supervised}, ParseNet~\cite{SharmaLMKCM20}, and Our approach.}
\label{Figure:Comparison}
\vspace{-0.15in}
\end{figure*}
\end{landscape}

%% file: chapters/HPNet/sections/05_results.tex
\subsection{Experimental Results}
\label{Section:Experimental:Results}


\subsubsection{Experimental Setup}
\label{Section:Experimental:Setup}
\paragraph{Datasets.}
We show experimental evaluation on two popular primitive segmentation datasets, i.e., ANSI Mechanical Component Dataset~\cite{li2019supervised} and ABCParts Dataset~\cite{SharmaLMKCM20}. ANSI mainly contains diverse mechanical components provided by TraceParts. Most of the objects in this dataset are composed by four basic primitives(plane, sphere, cylinder, and cone). We have 13k/3k/3k models on train/test/validation sets respectively. Each model contains 8192 points. ABCParts is derived from the ABC dataset~\cite{koch2019abc}, which provides a large source of 3D CAD models. In ABCParts, the objects are more complicated than those in ANSI, and each of them contains at least one B-spline surface patch. We have 24k/4k/4k models on train/test/validation sets on ABCParts and each model contains 10000 points. Please refer to supplementary materials for more details.

\input{chapters/HPNet/figures/fig_primitive_type}

\paragraph{Evaluation metrics.}
We follow the standard metrics proposed by Li \etal~\cite{li2019supervised} for quantitative evaluation.
\begin{itemize}[leftmargin=*,topsep=3pt]
\setlength\itemsep{1pt}
\item \textit{Seg-IoU:} Denoting $K$ as the number of ground-truth patches, this metric evaluates the segmentation mean IoU score: $\frac{1}{K}\sum^{K}_{k=1}\text{IoU}(\mathbf{W}_{:,k}, \hat{\mathbf{W}}_{:,k})$,
where $\mathbf{W}\in \{0,1\}^{n\times K}$ is the predicted segmentation membership matrix; $\hat{\mathbf{W}}\in \{0,1\}^{n\times K}$ is the ground truth.
\item \textit{Type-IoU:} $\frac{1}{K}\sum^K_{k=1}\mathcal{I}[t_k=\hat{t}_k]$, where $t_k$ is the predicted primitive type for the $k$-th segment and $\hat{t}_k$ is the ground truth. $\mathcal{I}$ is an indicator function.
\item \textit{Res-Error:} $\sum^K_{k=1}\frac{1}{m_k}\sum_{\hat{\textbf{s}}_k \in\hat{\set{P}}_k}d_{\set{P}_k}(\hat{\textbf{s}}_k)$, where $\set{P}_k$ is the predicted primitive path, $M_k$ is the number of sampled points $\hat{\textbf{s}}_k$ from ground truth primitive patch $\hat{\set{P}}_k$, $d_{\set{P}_k}(\hat{\textbf{s}}_k)$ is the distance between $\hat{\textbf{s}}_k$ and $\set{P}_k$.
\item \textit{P-coverage:} $\frac{1}{n}\sum^n_{i=1}\delta(\text{min}^K_{k=1}d_{\set{P}_k}(\bs{p}_i) < \epsilon)$, where $\epsilon = 0.01$.
\end{itemize}

\input{chapters/HPNet/tables/ablation_table}
\subsubsection{Analysis of Results}
\label{Section:Result:Analysis}

Table~\ref{tab:baseline-comparison} and Figure~\ref{Figure:Comparison} present quantitative and qualitative results of HPNet. We can see that HPNet produces results that are close to the underlying ground-truth. Thanks to the sharp-edge module, the segment boundaries are smooth. Moreover, HPNet can even rectify small over-segmented patches in the training data. Quantitatively, HPNet offers state-of-the-art results under all error metrics on these two datasets.

\paragraph{Baseline comparison.}
Our experimental study considers four baseline approaches. These include two state-of-the-art non-deep learning methods: nearest neighbor (NN)~\cite{Sharma_2018_CVPR} and Efficient RANSAC~\cite{schnabel2007efficient}, and two state-of-the-art deep learning methods: Supervised Primitive Fitting (SPFN)~\cite{li2019supervised} and ParseNet~\cite{SharmaLMKCM20}. Note that to build a fair comparison, we replace the network backbones in SPFN and ParseNet to keep the same as HPNet.

Quantitatively (see Table~\ref{tab:baseline-comparison}), HPNet leads to considerable performance gains from all baseline approaches. Specifically, on ANSI, HPNet leads to salient 12.89\% and 6.30\% improvements in Seg-IOU under the point and point+normal settings, respectively. The improvements under other metrics are also considerable. For example, the Res-Error numbers decrease from 0.013 to 0.011 and from 0.010 to 0.008 under the point and point+normal settings, respectively. The improvements on ABCParts are also considerable, the improvements on Seg-IOU are 4.15\% and 3.77\% under the point and point+normal settings, respectively. HPNet also exhibits consider improvements under other metrics.


Qualitatively (see Figure~\ref{Figure:Comparison}), HPNet leads to better results in the sense that it provides accurate segmentation for both large and small primitive patches. The segmentation boundaries are also smooth. In contrast, baseline approaches ParseNet and SPEN possesses non-smooth boundaries and the issues of over-segmentation and under-segmentation are noticeable.

The relative improvements on ANSI are bigger than those on ABCParts. As illustrated in Figure~\ref{fig:primitive_type}(a), we can understand this behavior from the fact that predictions of shape parameters of B-spline patches are less accurate than the other four primitive types.





\input{chapters/HPNet/figures/fig_sharp_edge}
\subsubsection{Ablation Study}
\label{Section:Ablation:Study}

We proceed to study the impacts of different components of HPNet. Table~\ref{tab:ablation} provides the overall Seg-IoU and Type-IoU scores after dropping each component of HPNet. Figure~\ref{fig:primitive_type}(b) shows Seg-IoU scores for different types of primitives.

\paragraph{No descriptor module.} Table~\ref{tab:ablation} shows that the descriptor module is critical for HPNet. Without this module, the Seg-IoU scores drop by 14.9\% and 17.5\% on ANSI and ABCParts, respectively. The Type-IoU scores drop by 6.5\% and 13.4\% on ANSI and ABCParts, respectively. Figure~\ref{fig:primitive_type}(b) shows that except for plane and sphere, the relative performance drops are glaring across other primitive types. This is expected as predicting accurate shape parameters requires global knowledge of the underlying primitive, which can be more difficult than predicting a semantic descriptor. The more significant drop on ABCParts than ANSI can be explained as the fact that ABCParts contain fewer primitives of types sphere and plane.

\vspace{-5pt}
\paragraph{No smoothness module.} As illustrated in Table~\ref{tab:ablation}, dropping the smoothness module leads to 0.67\% and 1.34\% decrements in the Seg-IoU scores and 0.17\% and 1.13\% decrements in the Type-IoU scores on ANSI and ABCParts, respectively. Figure~\ref{fig:primitive_type}(b) shows that the smoothness module contributes most to primitive types of cone, cylinder, and b-spline patches, which are likely to have sharp edges with other primitives. This also explains the smoothness module is slightly more effective on ABCParts than ANSI.

Figure~\ref{fig:boundary} illustrates the effects of the smoothness module. Note that the smoothness module promotes smooth primitive boundaries. It also utilizes the sharp edges to merge and split primitives that are not possible when dropping this module.

\paragraph{No consistency module.} As illustrated in Table~\ref{tab:ablation}, dropping the consistency module leads to 1.31\% and 1.69\% decrements in the Seg-IoU scores and 0.37\% and 2.18\% decrements in the Type-IoU scores on ANSI and ABCParts, respectively. Figure~\ref{fig:primitive_type}(b) shows that the consistency module contributes most to primitive types of \sphere, \cone, and \cylinder patches. One reason is that predictions of shape parameters tend to more accurate on these patches. On the other hand, the accuracy for \plane is already very high, leaving a small margin for improvements.

\input{chapters/HPNet/figures/fig_weight_illustration}

\paragraph{No weight learning.} Finally, we study the importance of learning combination weights. Table~\ref{tab:ablation} shows that using fixed combination weights leads to 0.08\% and 0.46\% decrements in the Seg-IoU scores and 0.00\% and 0.48\% decrements in the Type-IoU scores on ANSI and ABCParts, respectively. These statistics show that using adaptive weights is effective. Moreover, this strategy has larger impact on ABCParts than ANSI because the diversity of the primitives from ABCParts is larger than that from ANSI.

As shown in Figure~\ref{Figure:Weights}, the learned weights are adaptive for different models. When the model contains small and complex primitives (shape(a)), the semantic descriptor is more useful. If the shape parameter and type predictions are accurate (shape(b)), the consistency descriptor places a more important role. Finally, when sharp edges are prominent (shape(c)), the smoothness descriptor becomes critical.

%% file: chapters/HPNet/figures/fig_primitive_type.tex
\begin{figure}
\centering
\includegraphics[width=\columnwidth]{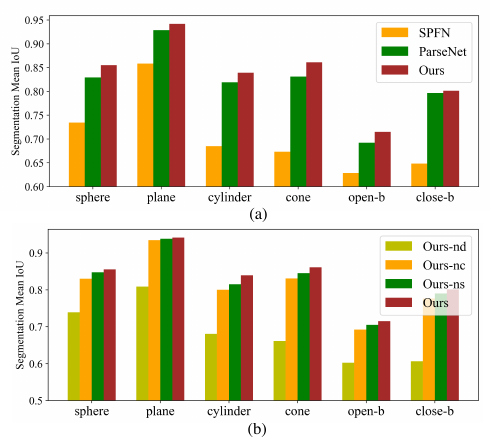}
\caption{Mean IoU of segmentation results on different primitive types. Here, open-b and close-b represent open and closed B-spline patches. (a): comparison between HPNet and baseline methods. (b): comparison between different components of HPNet.}
\label{fig:primitive_type}
\end{figure}

%% file: chapters/HPNet/tables/ablation_table.tex
\begin{table}

  \centering
  \begin{tabular}{l|c|c|c|c}
  \toprule
  &  \multicolumn{2}{c|}{ANSI} & \multicolumn{2}{c}{ABCParts}\\
  \hline
  &  Seg-IoU & Type-IoU & Seg-IoU & Type-IoU \\
  \hline
  Ours-nd   & 80.10 & 92.45 & 70.33 & 78.87 \\
  Ours-nc  & 92.92 & 98.53 & 83.80 & 89.41 \\
  Ours-ns  & 93.52 & 98.73 & 84.10 & 90.01 \\
  Ours-nw  & 94.07 & \textbf{98.90} & 84.78 & 90.56 \\
  Ours     & \textbf{94.15} & \textbf{98.90} & \textbf{85.24} & \textbf{91.04} \\
  \bottomrule
  \end{tabular}
{%
 \caption{Evaluation on different combinations of our approach. `Ours-nd' denotes dropping the output of the descriptor module.
 `Ours-nc' denotes dropping the output of the consistency module.
 `Ours-ns' denotes dropping the output of the smoothness module.  `Ours-nw' means our approach without weight learning. `Ours' corresponds to our full algorithm.}
 \label{tab:ablation}
}
\end{table}

%% file: chapters/HPNet/figures/fig_sharp_edge.tex
\begin{figure}
\centering
\footnotesize
\def\imh{0.3\columewidth}
\def\imw{0.15\textwidth}
\newcommand{\TT}[1]{\raisebox{-0.5\height}{#1}}
\setlength{\tabcolsep}{1pt}
\begin{tabular}{ccccccc}

\TT{\includegraphics[width=\imw]{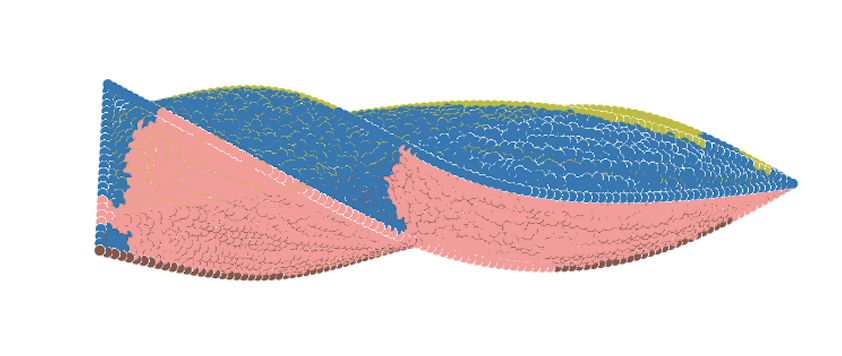}} & 
\TT{\includegraphics[width=\imw]{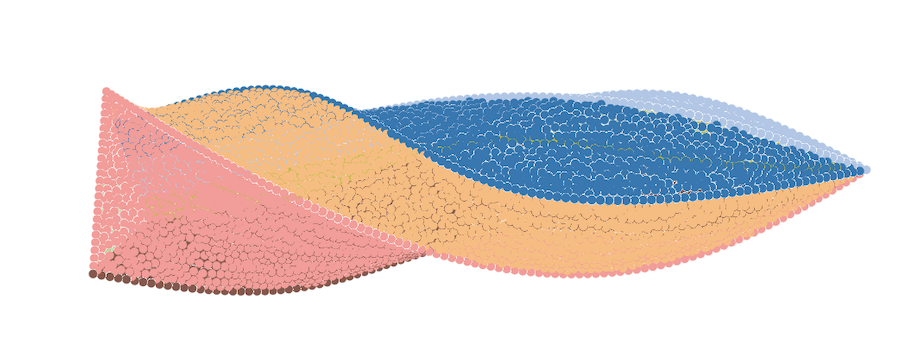}} & 
\TT{\includegraphics[width=\imw]{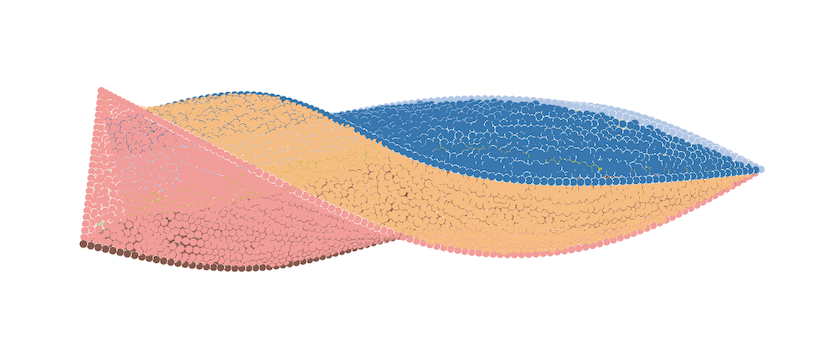}} \\
[+5pt]
\TT{\includegraphics[width=0.10\textwidth]{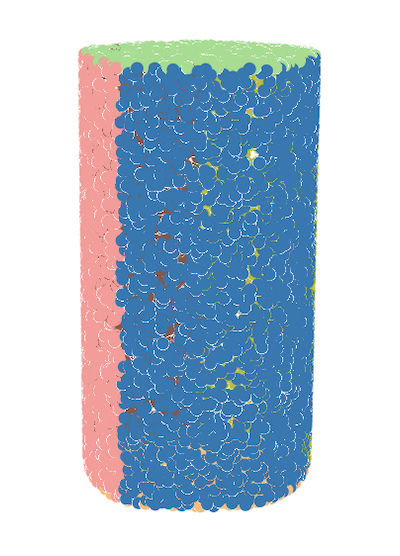}} & \TT{\includegraphics[width=0.11\textwidth]{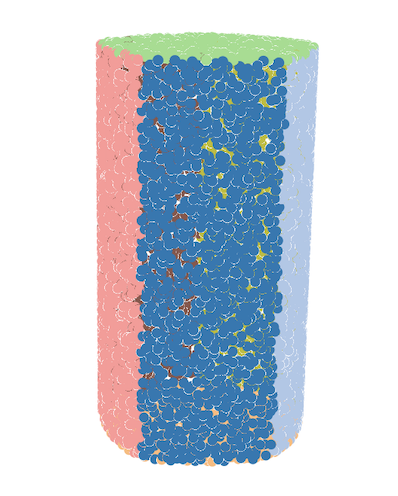}} & 
\TT{\includegraphics[width=0.12\textwidth]{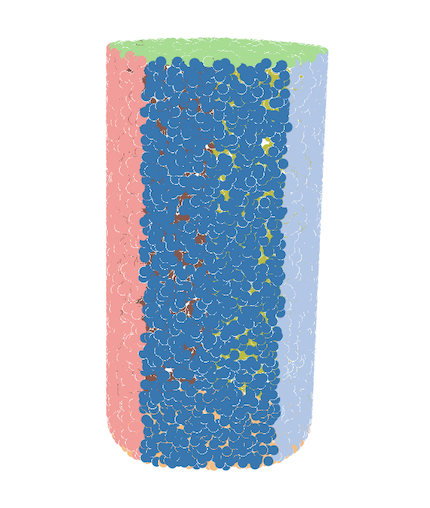}} \\
[+5pt]
\TT{\includegraphics[width=0.1\textwidth]{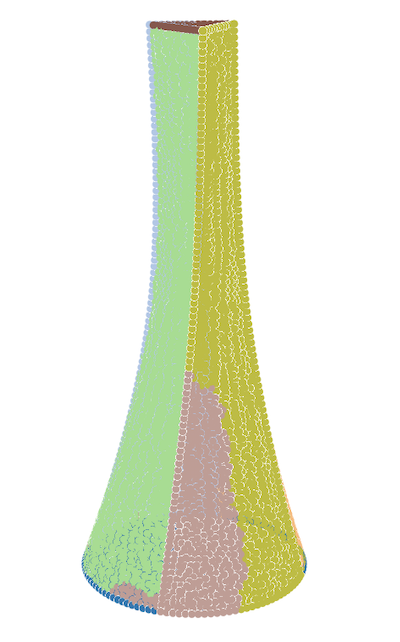}} & 
\TT{\includegraphics[width=0.11\textwidth]{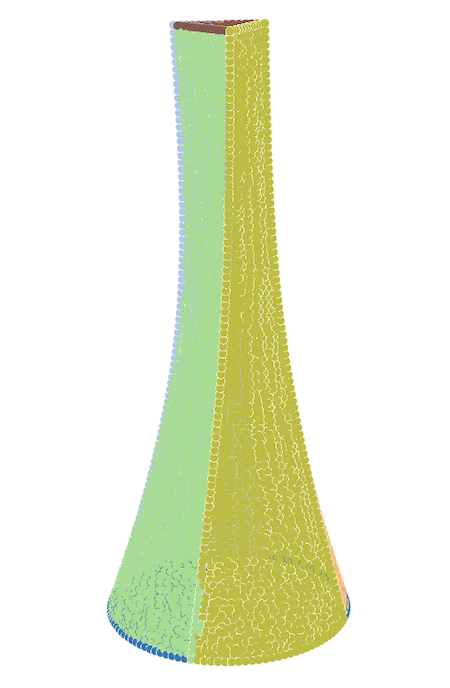}} & 
\TT{\includegraphics[width=0.11\textwidth]{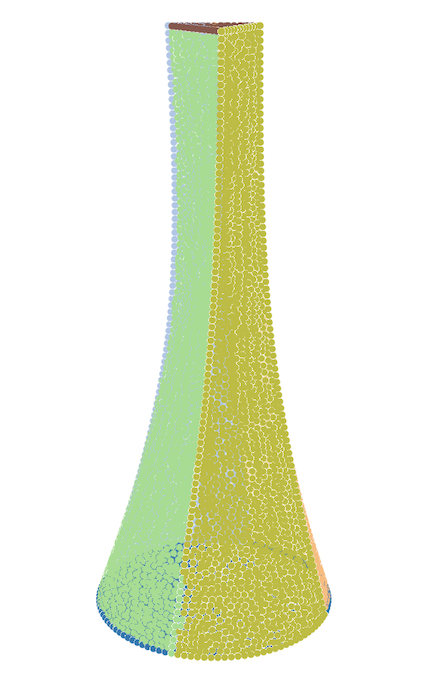}} \\

Ours-ns & Ours & G.T \\
\end{tabular}
\caption{Examples on comparison between with and without sharp edge descriptor. Here, `Ours-ns' represents our model without combining sharp edge descriptor. We notice that adding sharp edge descriptor helps model to capture boundary better.}
\label{fig:boundary}
\vspace{-0.15in}
\end{figure}

%% file: chapters/HPNet/figures/fig_weight_illustration.tex
\begin{figure}
\centering
\includegraphics[width=0.9\textwidth]{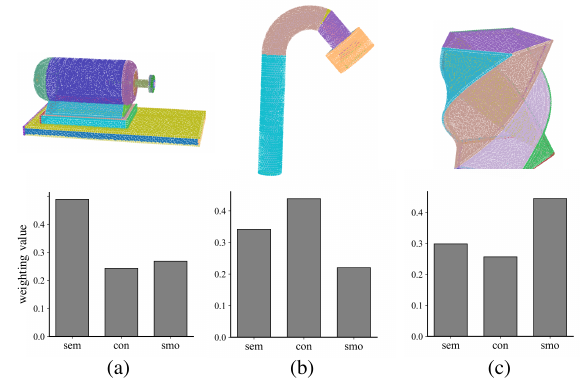}
\caption{Illustration on the learned weights for three typical shapes. For simplicity, we use the L2 norm of the weighting vector of each type of spectral descriptors to represent each spectral embedding space. `sem' denotes semantic descriptor. `con' denotes geometric consistency descriptor. `smo' denotes smoothness descriptor.}
\label{Figure:Weights}
\end{figure}

%% file: chapters/HPNet/sections/06_conclusions.tex
\subsection{Conclusion}
\label{Section:Conclusions}

In this section, we have presented HPNet which combines multiple segmentation cues for primitive shape segmentation. Experimental results show that HPNet leads to considerable performance gains compared to previous approaches that leverage a single segmentation cue. Moreover, making the combination weights adaptive to the input models leads to additional performance improvement. Our ablation study further justifies different components of HPNet.

One limitation of HPNet is that it does not utilize symmetric relations among geometric primitives (c.f~\cite{Li:2011:GF}). In the future, we plan to study novel graph neural networks to detect and enforce structural relations among primitives. Detecting such relations is essential for many models such as architectures.

%% file: chapters/pc_ssl.tex
\chapter{Point Cloud Self-supervised Representation Learning}\label{ch:pc-ssl}

\renewcommand{\thefootnote}{\color{red}\fnsymbol{footnote}}
\footnote[6]{The content of this chapter includes my publications ``Implicit AutoEncoder for Point-Cloud Self-supervised Representation Learning''~\cite{yan2023implicit} in International Conference on Computer Vision (ICCV) 2023 and ``3D Feature Prediction for Masked-AutoEncoder-Based Point Cloud Pretraining''~\cite{yan20233d} in International Conference on Learning Representations (ICLR) 2024. I am the first author of the publications.}

In the previous chapter we study the supervised learning methods for point cloud representation learning. The present work demonstrated an effective approach for the primitive segmentation task. However, a limitation of supervised learning methods is that they are typically tailored to specific tasks. In this chapter, we further study the self-supervised learning methods for point cloud representation learning. 
Point cloud self-supervised representation learning is a method for training models to understand 3D point cloud data without the need for explicit labels. Instead, models learn to extract meaningful features by solving pretext tasks that generate supervisory signals from the data itself. These tasks might include predicting point cloud transformations, reconstructing occluded parts, or distinguishing between different augmented views of the same point cloud. Architectures such as PointNet are often employed for these tasks, utilizing techniques like contrastive learning or autoencoding to learn robust and generalizable features. By learning these intrinsic properties of point clouds, models can later be fine-tuned on various downstream tasks such as classification, segmentation, or object detection with minimal labeled data, making self-supervised learning a powerful approach for leveraging vast amounts of unlabeled 3D data.
In Section~\ref{sec:iae}, we propose a new point cloud encoder that incorporates the advantages of implicit functions. In Section~\ref{sec:maskfeat3d}, we explore the optimal paradigm of masked autoencoders for point cloud data. I am the main author of the presented works.

\input{chapters/IAE/IAE}

\pagebreak

\input{chapters/MaskFeat3D/MaskFeat3D}

\pagebreak

%% file: chapters/IAE/IAE.tex
\section{Implicit AutoEncoder for Point-Cloud Self-supervised Representation Learning}
\label{sec:iae}

\renewcommand{\thefootnote}{\color{red}\fnsymbol{footnote}}
\footnote[6]{The content of this section is based on my publication ``Implicit AutoEncoder for Point-Cloud Self-supervised Representation Learning''~\cite{yan2023implicit} in International Conference on Computer Vision (ICCV) 2023. I am the first author of the publication.}

This section advocates the use of implicit surface representation in autoencoder-based self-supervised 3D representation learning. The most popular and accessible 3D representation, i.e., point clouds, involves discrete samples of the underlying continuous 3D surface. This discretization process introduces sampling variations on the 3D shape, making it challenging to develop transferable knowledge of the true 3D geometry. In the standard autoencoding paradigm, the encoder is compelled to encode not only the 3D geometry but also information on the specific discrete sampling of the 3D shape into the latent code. This is because the point cloud reconstructed by the decoder is considered unacceptable unless there is a perfect mapping between the original and the reconstructed point clouds. This paper introduces the Implicit AutoEncoder (IAE), a simple yet effective method that addresses the sampling variation issue by replacing the commonly-used point-cloud decoder with an implicit decoder. The implicit decoder reconstructs a continuous representation of the 3D shape, independent of the imperfections in the discrete samples. Extensive experiments demonstrate that the proposed IAE achieves state-of-the-art performance across various self-supervised learning benchmarks.

\input{chapters/IAE/sections/01_intro}
\input{chapters/IAE/sections/02_related}
\input{chapters/IAE/sections/03_approach}
\input{chapters/IAE/sections/04_results}
\input{chapters/IAE/sections/08_theory}
\input{chapters/IAE/sections/06_conclusion}

%% file: chapters/IAE/sections/01_intro.tex
\subsection{Introduction}
\label{Section:Introduction}

The rapid advancement and growing accessibility of commodity scanning devices have made it easier to capture vast amounts of 3D data represented by point clouds. The success of point-based deep networks~\cite{qi2017pointnet, qi2017pointnet++} additionally enables many 3D vision applications to exploit this natural and flexible representation. Exciting results have emerged in recent years, ranging from object-level understanding, including shape classification~\cite{chang2015shapenet} and part segmentation~\cite{yi2016scalable, sharma2020parsenet, yan2021hpnet}, to scene-level understanding, such as 3D object detection~\cite{dai2017scannet, song2015sun, geiger2012we} and 3D semantic segmentation~\cite{armeni20163d}. 
\input{chapters/IAE/figures_tex/teaser_v3}

Annotating point-cloud datasets is a highly labor-intensive task due to the challenges involved in designing 3D interfaces and visualizing point clouds. As a result, researchers are motivated to explore self-supervised representation learning paradigms. The fundamental concept is to pre-train deep neural networks on large unlabeled datasets and fine-tune them on smaller datasets annotated based on specific task requirements. A carefully designed self-supervised representation learning paradigm effectively initializes the network weights, enabling fine-tuning on downstream tasks ({\em e.g.}, classification and segmentation) to avoid weak local minimums and achieve improved stability~\cite{erhan2010does}.

Research in self-supervised representation learning predates the history of 3D computer vision. Notably, significant effort has been dedicated to developing self-supervised learning methods for 2D images~\cite{pathak2016context, wu2018unsupervised, doersch2015unsupervised, masci2011stacked, chen2020simple, zhuang2021unsupervised, zhuang2019self}, with autoencoders being one of the most popular tools~\cite{pathak2016context,bengio2013representation,tschannen2018recent,vincent2008extracting,he2021masked}. An autoencoder-based self-supervised representation learning pipeline comprises an encoder that transforms the input into a latent code and a decoder that expands the latent code to reconstruct the input. Since the latent code has a much lower dimension than the input, the encoder is encouraged to summarize the input by condensed latent features with the help of the reconstruction loss.

There is a growing interest in developing self-supervised representation learning methods for point clouds by drawing inspiration from image-based autoencoders. For example, Yang et al.~\cite{yang2018foldingnet} propose a novel folding-based decoder, and Wang et al.~\cite{wang2020unsupervised} develop a denoising autoencoder based on the standard point-cloud completion model. However, to the best of our knowledge, prior works in point-cloud self-supervised representation learning have relied on the same design principles as image-based methods, where \textbf{both the encoder and decoder represent the 3D geometry in the same format} ({\em i.e.}, point clouds).

Point clouds are noisy, discretized, and unstructured representations of 3D geometry. As shown in Figure~\ref{fig:teaser}, a 3D shape can be represented by many different point clouds, all of which are valid representations of the same object. Different point cloud samples are subject to different noises, which are induced from various sources such as the intrinsic noises from the sensors and the interference from the environment. The unordered and noisy nature distinguishes point clouds from conventional structured data, such as pixel images defined on rectangular grids.  When training a point-cloud autoencoder, the encoder is forced to capture \textit{sampling variations}, limiting the model's ability to extract valuable information about the true 3D geometry.

This paper, for the first time, formalizes the concept of sampling variations and proposes to combine the implicit surface representation with point-cloud self-supervised representation learning. Specifically, we introduce an asymmetric point-cloud autoencoder scheme, where the encoder takes a point cloud as input, and the decoder uses the implicit function as the output 3D representation. The rest of the paper refers to this design as the Implicit AutoEncoder (IAE). IAE enjoys many advantages over existing methods. First, reconstruction under the implicit representation discourages latent space learning from being distracted by the imperfections brought by sampling variations and encourages the encoder to capture generalizable features from the true 3D geometry. Second, the reconstruction loss defined on implicit functions bypasses the expensive and unstable data association operation commonly used in point-based reconstruction losses such as the Earth Mover Distance~\cite{rubner2000earth} and the Chamfer Distance~\cite{fan2017point}. The added efficiency allows IAE to process up to 40k input points with a single Tesla V100 GPU, making it possible to capture very fine geometric details when existing methods can only work with sparse data with approximately 1k points.

To demonstrate the effectiveness of IAE, we conduct experiments to verify that the learned representation from our pre-trained model can adapt to various downstream tasks, both at the object level and the scene level, including shape classification, linear evaluation, object detection, and indoor semantic segmentation. IAE consistently outperforms state-of-the-art methods in all settings. 
Specifically, under the best setting, IAE achieves 88.2\% / 94.3\% classification accuracy on ScanObjectNN~\cite{uy-scanobjectnn-iccv19} / ModelNet40~\cite{wu20153d}.
IAE is also the first to support the autoencoding paradigm in scene-level pre-training for the application to various scene-level downstream tasks. For example, compare to training from scratch, IAE achieves +0.7\% and +1.1\% absolute improvements in object detection quality evaluated by mAP@0.5 on ScanNet~\cite{dai2017scannet} and SUN RGB-D~\cite{song2015sun}, respectively.

We summarize the contributions of our paper as follows:

\begin{enumerate}[leftmargin=*]\setlength\itemsep{0mm}
    \item[-] We propose an asymmetric point-cloud autoencoder called Implicit AutoEncoder (IAE). The IAE takes a point cloud as input and uses an implicit function as the output 3D representation. We combine the implicit surface representation with point-cloud self-supervised representation learning for the first time.
    \item[-] We formalize the concept of
sampling variations in point clouds and demonstrate that IAE is more effective at capturing generalizable features from true 3D geometry than standard point-cloud autoencoders.
    \item[-] We conduct experiments to demonstrate the effectiveness of IAE on various downstream tasks, including shape classification, object detection, and indoor semantic segmentation. IAE consistently outperforms state-of-the-art methods in all settings.
    
\end{enumerate}

%% file: chapters/IAE/figures_tex/teaser_v3.tex
\begin{figure}[t]
    \centering
    \includegraphics[width=0.7\columnwidth]{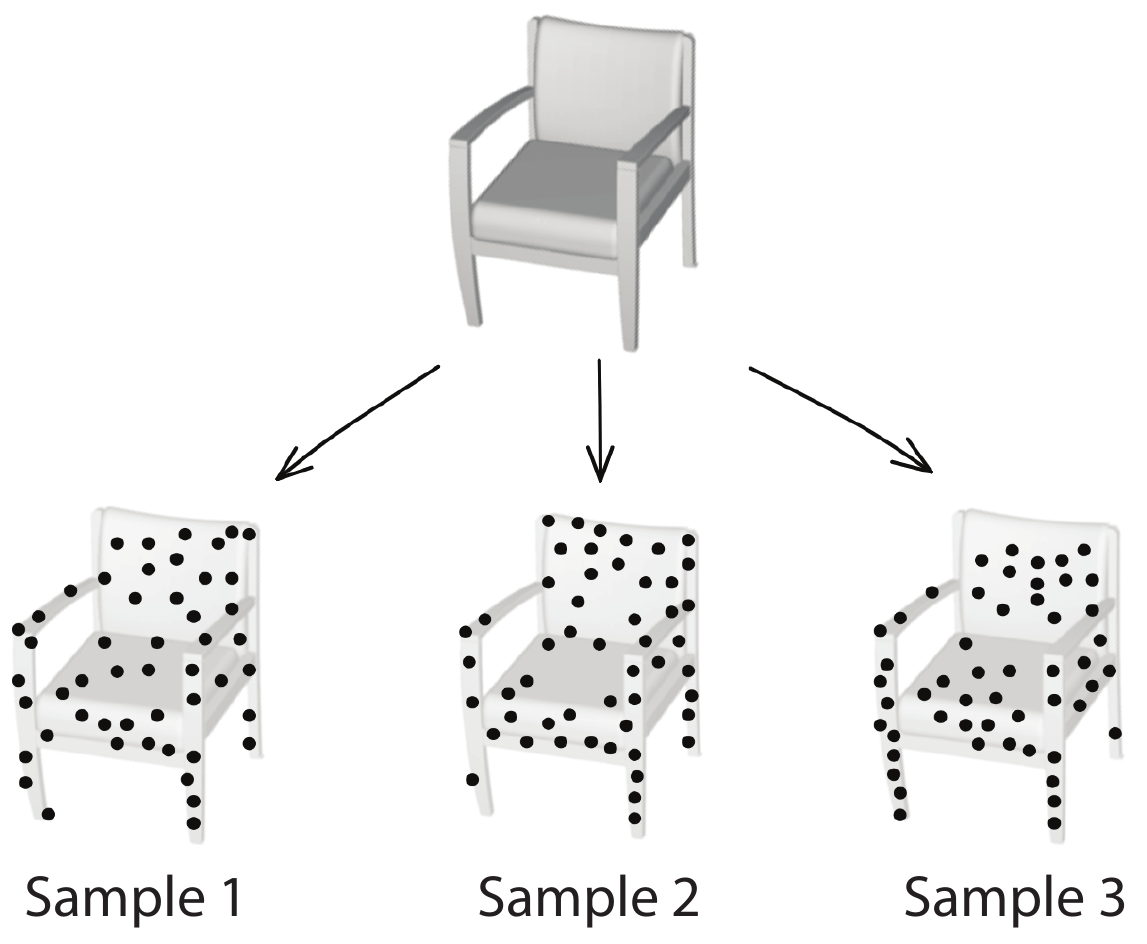}
    \caption{The Sampling Variation Problem. Given a continuous 3D shape, there are infinitely many ways to sample a point cloud. The proposed Implicit AutoEncoder (IAE) learns a latent representation of the true 3D geometry independent of the specific discrete sampling process. By alleviating the sampling variation problem, IAE improves existing point-cloud self-supervised representation learning methods in various downstream tasks.}
    \label{fig:teaser}
    \vspace{-10pt}
\end{figure}

%% file: chapters/IAE/sections/02_related.tex
\subsection{Related Work}
\noindent\textbf{Point-Cloud Self-Supervised Representation Learning}. The unordered nature of 3D point clouds poses a unique challenge to representation learning, which is known to be effective on conventional structured data representations like images. Several point-cloud self-supervised methods have been developed to learn representations from individual 3D objects~\cite{wang2020unsupervised, sauder2019self, yang2018foldingnet, poursaeed2020self, achlioptas2018learning, hassani2019unsupervised, li2018so, chen2021shape, yu2021point}, with the synthetic ShpapeNet~\cite{chang2015shapenet} being the most popular dataset used in pre-training. However, existing methods exhibit limited generalizability to real scene-level datasets due to a significant domain gap~\cite{xie2020pointcontrast}.

The remarkable success achieved in 2D has sparked a growing interest in extending the contrastive learning paradigm to self-supervised representation learning on point clouds~\cite{xie2020pointcontrast, rao2021randomrooms, zhang2021self}. These approaches rely on the contrastive loss to separate positive and negative instance pairs in the latent space, necessitating meticulous design efforts to define similar and dissimilar examples. They also consume a substantial amount of computational resources, particularly when working with large datasets. While 3D contrastive learning is a popular field of research, this paper addresses the sampling variation problem under the simple autoencoder paradigm discussed below.

\noindent\textbf{Point-Cloud Autoencoders}. Several existing works exploit the autoencoder paradigm for point-cloud self-supervised representation learning. They simultaneously train an encoder and a decoder, with the encoder mapping the input point cloud to a latent code, and the decode reconstructing another point cloud from the latent code~\cite{yang2018foldingnet, wang2020unsupervised, pang2022masked, zhang2022point}. While these approaches have demonstrated promising performance improvements, the sampling variation problem remains a significant challenge when the decoder output is represented as point clouds. This is because the network must encode not only the geometry but also the imperfections of the specific point cloud sampling into the latent code. Additionally, training a decoder under a point-based network architecture is not as straightforward as training a regular convolutional or fully-connected model, which limits the quality of the output. This paper proposes the use of implicit representation as the decoder output in the autoencoder-based self-supervised representation learning paradigm. As we will see, the implicit decoder alleviates the sampling variation issue and allows for a more efficient pipeline than existing approaches. Closely related to our design, ParAE~\cite{eckart2021self} learns the 3D data distribution through discrete generative models. While ParAE builds the supervision upon a point-wise partitioning matrix, the proposed IAE utilizes implicit representations that are conceptually more relevant to the 3D geometry and easier to train.

\noindent\textbf{Implicit Representations}. A 3D shape can be represented implicitly by a deep network that maps 3D coordinates to signed distances~\cite{michalkiewicz2019implicit, park2019deepsdf} or occupancy grids~\cite{chen2019learning, mescheder2019occupancy, peng2020convolutional}. Unlike explicit representations such as point clouds, voxels, or triangle meshes, implicit representations offer a continuous shape parameterization and do not suffer from discretization errors. Implicit representations have been successfully applied in various 3D tasks, including 3D reconstruction from images~\cite{liu2020dist, liu2019learning, niemeyer2020differentiable}, primitive-based 3D reconstruction~\cite{genova2020local, genova2019learning, paschalidou2020learning}, 4D reconstruction~\cite{niemeyer2019occupancy}, and continuous texture representation~\cite{oechsle2019texture}. However, to the best of our knowledge, no existing research has investigated the integration of implicit representations into the self-supervised representation learning paradigm, making IAE the pioneer in the field.

%% file: chapters/IAE/sections/03_approach.tex
\subsection{Method}
\label{sec:method}

\subsubsection{Problem Setup}
Assume we have access to a large dataset of 3D point clouds $\mathbf{D} = \{\mathcal{P} | \mathcal{P} \in \mathbb{R}^{n \times 3}\}$ without any annotations. The goal of autoencoder-based point-cloud self-supervised learning is to train an encoder network $f_{\Theta}$ and a decoder network $g_{\Phi}$, where the encoder maps the point cloud to an m-dimensional latent code:
\begin{equation}
\small
    f_{\Theta}: \mathbb{R}^{n \times 3} \rightarrow \mathbb{R}^{m},  m \ll n
\end{equation}
and the decoder maps the latent back to the 3D geometry:
\begin{equation}
\small
    g_{\Phi}: \mathbb{R}^{m} \rightarrow \mathbb{S}
\end{equation}
Here, $\mathbb{S}$ refers to the space of 3D shapes under the representation of choice. In practice, a popular choice is to use a sub-sampled version of the point cloud as the output shape representation: $\mathbb{S} = \mathbb{R}^{n' \times 3}$, where $n' \leq n$. The parameters $\Theta$ and $\Phi$ are jointly trained by minimizing the distance metric $d$ between the input point cloud and the reconstructed shape:
\begin{equation}
\small
    \Theta^*, \Phi^* = \argmin_{\Theta, \Phi} d((g_{\Phi} \circ f_{\Theta})(\mathcal{P}), \mathcal{P})
\end{equation}
After training the autoencoder, the encoder $f_{\Theta^*}$ is fine-tuned on a small dataset with task-specific annotations ({\em e.g.}, bounding boxes and segmentation labels).

\subsubsection{Challenges}
\label{sec:challenge}
\noindent\textbf{Sampling Variations}. As shown in Figure~\ref{fig:teaser}, there are infinitely many valid point clouds representing the same 3D shape. This is because point clouds are discrete samples of the continuous 3D geometry. Due to the finite sampling size and noisy conditions, each point cloud contains a unique set of imperfections from the sampling process. Ideally, an autoencoder is expected to encode the information just about the true 3D geometry into the latent code. In practice, the encoder is compelled to capture the imperfections in the latent code since the decoder output is represented as another point cloud with a distance metric in place enforcing the two discrete samples to be identical. Sampling variation causes distractions to the autoencoder learning and leads to suboptimal performance in downstream tasks. 

\noindent\textbf{Efficient Distance Calculation}. The Earth Mover Distance (EMD)~\cite{rubner2000earth} in Eq.~\ref{eq:emd} and the Chamfer Distance (CD)~\cite{fan2017point} in Eq.~\ref{eq:cd} are commonly used metrics to calculate the distance between the input and the reconstructed point clouds. For each point $\mathbf{x}$ in point cloud $\mathcal{P}$, EMD and CD find the nearest point in the other cloud $\hat{\mathcal{P}}$ for distance calculation. The correspondence discovery operation makes both EMD and CD expensive metrics to use, especially when the point clouds are dense. 

\begin{equation}
\small
    d_\text{EMD}(\mathcal{P}, \hat{\mathcal{P}}) =  \min_{\phi: \mathcal{P} \rightarrow \hat{\mathcal{P}}}\sum_{\mathbf{x} \in \mathcal{P}}||\mathbf{x} - \phi(\mathbf{x})||_2
    \label{eq:emd}
\end{equation}
\vspace{-4pt}
\begin{equation}
    \small
    d_\text{CD}(\mathcal{P}, \hat{\mathcal{P}}) = \sum_{\hat{\mathbf{x}} \in \hat{\mathcal{P}}} \min_{\mathbf{x} \in \mathcal{P}} ||\hat{\mathbf{x}} - \mathbf{x}||_2 +  \sum_{\mathbf{x} \in \mathcal{P}}\min_{\hat{\mathbf{x}} \in \hat{\mathcal{P}}} ||\mathbf{x} - \hat{\mathbf{x}}||_2
    \label{eq:cd}
\end{equation}

\input{chapters/IAE/figures_tex/main_figure}

\noindent\textbf{Data Density}. Many large-scale open 3D datasets provide very dense point clouds. For example, the popular ShapeNet dataset~\cite{chang2015shapenet} contains 50k points per shape. Due to the inefficiency discussed above, existing explicit autoencoders lack the ability to exploit such density and are limited to a sparely sub-sampled version of the original dense point cloud. Sub-sampling introduces information loss and additional sampling variations. If it \textit{could} use dense point clouds during training, an explicit autoencoder \textit{would} potentially be as good as the proposed IAE. The problem, however, is that it is non-trivial to use dense data in explicit autoencoders. A naive adoption of the EMD~\cite{rubner2000earth} and CD~\cite{fan2017point} losses leads to an unaffordable computation cost. We show more analysis in Section~\ref{sec:analysis}.

\footnotetext{For simplification, we denote $f$ as $f_{\Theta}$, $g$ as $g_{\Phi}$, and do not show the query point $\mathbf{x}$ in Figure~\ref{fig:main_figure}.} 

\subsubsection{The Implicit AutoEncoder (IAE)}
\noindent\textbf{Design}.
As illustrated by Figure~\ref{fig:main_figure}, we propose the Implicit AutoEncoder (IAE), which elegantly addresses the challenges discussed above. In IAE, while the encoder still takes a point cloud as input, the decoder outputs an implicit function: $\mathbb{S} = \{\lambda | \lambda(\mathbf{x}) \in \mathbb{R}, \forall \mathbf{x} \in \mathbb{R}^3 \}$. Let $\lambda_\text{gt}$ be the ``ground-truth'' implicit function derived analytically from the input shape, the training objective of IAE is to minimize the distance between two implicit surfaces through the distance metric $d_\text{imp}$:
\begin{equation}
\small
    \Theta^*, \Phi^* = \argmin_{\Theta, \Phi} d_\text{imp}((g_{\Phi}\circ f_{\Theta})(\mathbf{x}|\mathcal{P}), \lambda_\text{gt}(\mathbf{x}))
    \label{eq:imp}
\end{equation}
Our experiments use three popular variants of the implicit function, including the signed distance function~\cite{michalkiewicz2019implicit, park2019deepsdf}, the unsigned distance function~\cite{chibane2020neural}, and the occupancy grid~\cite{chen2019learning, mescheder2019occupancy, peng2020convolutional}. When the output shape is represented by either the signed (SDF) or the unsigned (UDF) distance function, $d_\text{imp}$ is implemented as the mean $\mathcal{L}_1$ distance between the ground-truth and reconstructed distance functions on $N$ uniformly sampled points inside the 3D bounding box:

\begin{equation}
\small
    d_\text{sdf} = \frac{1}{N}\sum_{\mathbf{x} \in X} |(g_{\Phi}\circ f_{\Theta})(\mathbf{x}|\mathcal{P}) - s(\mathbf{x})|
    \label{eq:sdf}
\end{equation}
\begin{equation}
\small
    d_\text{udf} = \frac{1}{N}\sum_{\mathbf{x} \in X} ||(g_{\Phi}\circ f_{\Theta})(\mathbf{x}|\mathcal{P})| - u(\mathbf{x})|
    \label{eq:sdf}
\end{equation}
where $s$ denotes SDF value of point $\mathbf{x}$, and $u$ denotes UDF value of point $\mathbf{x}$.

On the other hand, when the output shape is represented by the occupancy grid, $d_\text{imp}$ becomes the cross-entropy loss:

\begin{equation}
\small
    d_\text{occ} = \frac{1}{N}\sum_{\mathbf{x} \in X}CE((g_{\Phi}\circ f_{\Theta})(\mathbf{x}|\mathcal{P}), o(\mathbf{x}))
    \label{eq:occ}
\end{equation}
where $CE$ denotes cross entropy, $o$ denotes occupancy.


%% file: chapters/IAE/figures_tex/main_figure.tex
\begin{figure*}[h]
\centering
\includegraphics[width=\textwidth]{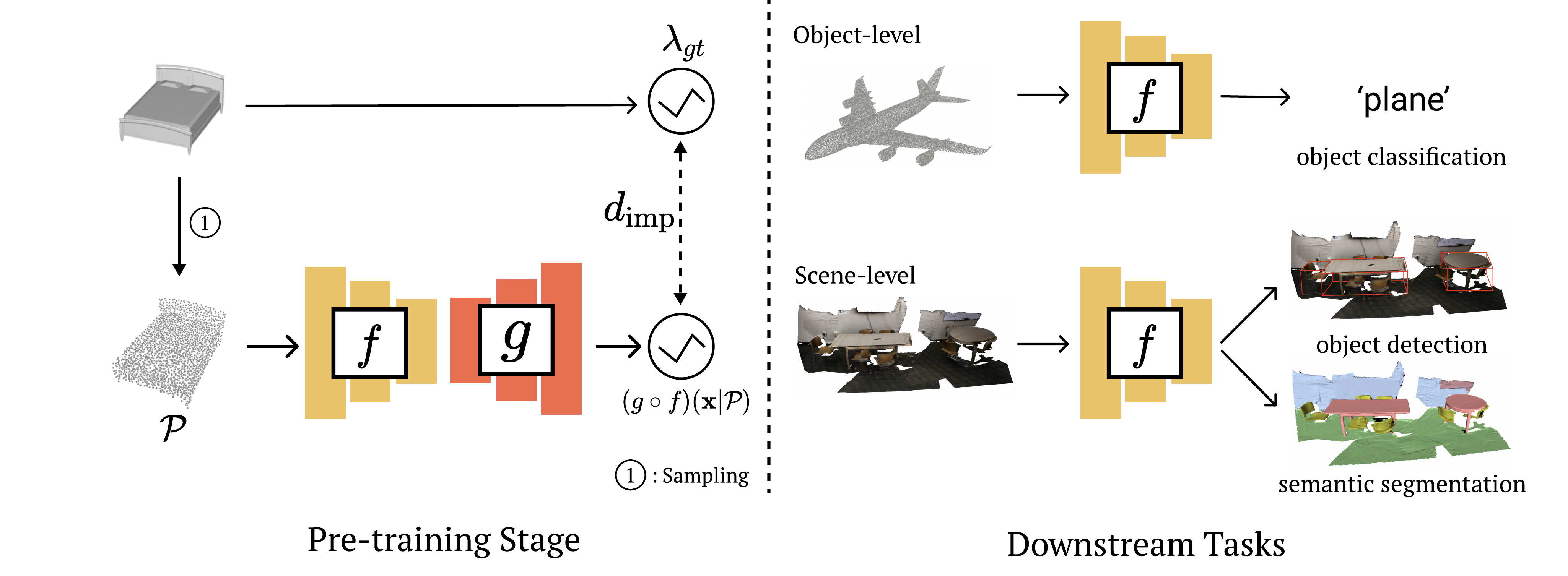}
\caption{The Implicit AutoEncoder. In the pre-training stage, we jointly train an encoder $f_{\Theta}$ and a decoder $g_{\Phi}$ to reconstruct the underlying 3D shape from the input\protect\footnotemark. While the encoder takes the point cloud as input, the output of the decoder is represented by an implicit function. The implicit function maps a 3D query point $\mathbf{x} = (x, y, z)$ to the signed~\cite{michalkiewicz2019implicit, park2019deepsdf}, unsigned~\cite{chibane2020neural}, or occupancy~\cite{chen2019learning, mescheder2019occupancy, peng2020convolutional} representation of the shape. After pre-training, we fine-tune the encoder $f_{\Theta}$ on different downstream tasks. At the object level, we evaluate IAE on object classification. At the scene level, we evaluate IAE on object detection and semantic segmentation.}
\label{fig:main_figure}
\vspace{-0.1in}
\end{figure*}

%% file: chapters/IAE/sections/04_results.tex
\subsection{Experiments}

\input{chapters/IAE/tables/cls_2}

\subsubsection{Pre-Training Datasets}
\label{sec:setup}
\noindent\textbf{ShapeNet}~\cite{chang2015shapenet} is a synthetic dataset commonly used in 3D vision, consisting of 57,748 shapes from 55 object categories. In addition to point clouds, ShapeNet offers high-quality water-tight triangular meshes as a more accurate description of 3D geometry. Following~\cite{park2019deepsdf,mescheder2019occupancy}, we generate ground-truth signed distance functions and occupancy grids from the provided mesh data. While ShapeNet is able to provide such high-quality 3D geometric information thanks to its synthetic nature, water-tight meshes are rarely available in real-world settings. To assess the effectiveness of IAE under more realistic conditions, we further use the nearest neighbor strategy~\cite{peterson2009k} to obtain unsigned distance functions from the point clouds in ShapeNet (50k points). 

\noindent\textbf{ScanNet}~\cite{dai2017scannet} is a real-world dataset with over 1,500 point-cloud scans. Unlike ShapeNet~\cite{chang2015shapenet}, which focuses on individual objects, ScanNet collects indoor scene data with a significantly larger perceptive range. We apply the sliding window strategy and crop each instance into $d \times d \times d$ cubes, where $d=3.0$ m. Following Qi et al.~\cite{qi2019deep}, we divide the dataset into training (8k point clouds) and validation (2.6k point clouds) splits. We randomly sub-sample 10,000 points from each point cloud as the input to the encoder. Due to the lack of high-quality water-tight meshes, we use the nearest neighbor strategy~\cite{peterson2009k} to calculate ground-truth unsigned distance functions. We refer interested readers to the supplementary material for implementation details. Importantly, we point out that IAE is the first method that supports the autoencoding paradigm in real-world scene-level training. As observed by Xie et al.~\cite{xie2020pointcontrast}, pre-training on synthetic datasets exhibits limited generalizability to real-world tasks. Real-world scene-level pre-training is extremely challenging due to noises, incomplete scans, and the complication of indoor scene contents. As a result, existing self-supervised methods fail to achieve any noticeable performance improvement~\cite{sauder2019self, poursaeed2020self, wang2020unsupervised}.

\subsubsection{Pre-Training Setting}
\label{sec:setting}
As outlined in Section~\ref{sec:method}, we propose a general framework for Implicit Autoencoder. The IAE consists of an encoder part that maps a given point cloud into a high-dimensional latent code, and a decoder part that fits an implicit function to the latent code and query point to output the implicit function value. Our framework can easily incorporate any backbone architecture for point cloud understanding into the encoder part. In order to ensure a fair comparison, we adopt the same backbone architecture as previous approaches.

The decoder part of the IAE fits an implicit function that takes both the latent code obtained from the encoder and the query point as input, and outputs the implicit function value at that point. We explore two different designs for the decoder part based on Occupancy Network-style~\cite{mescheder2019occupancy} and Convolutional Occupancy Network-style~\cite{peng2020convolutional}. The pretrained models used for downstream tasks in our experiments are based on the Convolutional Occupancy Network-style decoder. Further implementation details are provided in the supplementary material.

\footnotetext{DepthContrast used a slightly larger model than Votenet. For a fair comparison, we reproduce DepthContrast with the original Votenet model.} 

\subsubsection{Downstream Tasks}
\label{sec:tasks}
After pre-training IAE on ShapeNet~\cite{chang2015shapenet} and ScanNet~\cite{dai2017scannet}, we discard the decoder in pre-training and append different network heads onto the pre-trained encoder for different downstream tasks, at both the object level and the scene level.

\paragraph{Shape Classification}
\label{sec:tasks:shape}
We assess the quality of object-level feature learning on ModelNet40~\cite{wu20153d} and ScanObjectNN ~\cite{uy2019revisiting} datasets. ModelNet40 consists of 9,832 training objects and 2,468 testing objects across 40 different categories. We additionally pre-process the dataset following the procedure developed by Qi et al~\cite{qi2017pointnet}. ScanObjectNN contains approximately 15,000 objects from 15 categories, which are scanned from real-world indoor scenes with cluttered backgrounds. We conduct experiments on three variants: OBJ-BG, OBJ-ONLY, and PBT50-RS. Details are provided in supplementary materials.

\noindent\textbf{Linear SVM Evaluation.} 
To evaluate the 3D representation learning ability of IAE, we first conduct linear classification experiments on ModelNet40~\cite{wu20153d} and ScanObjectNN~\cite{uy2019revisiting} after pre-training on ShapeNet. We use our frozen pre-trained encoder to extract features from 1,024 points sampled from each 3D shape and train a linear SVM classifier on top of them. The classification accuracies are reported in Table~\ref{tab:cls} LINEAR section. To establish a fair comparison with previous approaches, we construct two different encoders, DGCNN~\cite{wang2019dynamic} and M2AE~\cite{zhang2022point}. Under M2AE backbone, IAE achieves 93.2\%/84.4\% classification accuracy on the ModelNet40/ScanObjectNN testing split, representing a 0.3\%/1.3\% absolute improvement from the state-of-the-art method. Furthermore, we highlight that IAE is capable of surpassing state-of-the-art performance even without high-quality water-tight mesh labels. In the second-to-last row of Table~\ref{tab:cls} LINEAR section, we present the results when the ground-truth implicit representation is directly generated from the dense point cloud on ShapeNet using the nearest neighbor strategy. The 1.2\% absolute improvement on ScanObjectNN indicates that even though point clouds are discrete samples of the true 3D geometry, converting point clouds to the implicit format is an effective way of alleviating sampling variations.

\noindent\textbf{Supervised Fine-Tuning.} 
In addition to fitting an SVM in the latent space, we fine-tune the encoder pre-trained on ShapeNet using ModelNet40~\cite{wu20153d} and ScanObjectNN~\cite{uy2019revisiting} labels. We again adopt both DGCNN~\cite{wang2019dynamic} and M2AE~\cite{zhang2022point} architecture to construct the encoder network for a fair comparison, whose outputs are utilized as class predictions in this downstream task. As shown in Table~\ref{tab:cls} FULL section, IAE shows consistent improvements and achieves state-of-the-art performance with 94.3\%/88.2\% classification accuracy on ModelNet40/ScanObjectNN respectively. IAE also surpasses previous state-of-the-art with 94.2\%/88.0\% classification accuracy when the ground-truth implicit function is computed without meshes. 

\paragraph{Indoor 3D Object Detection}

\input{chapters/IAE/tables/detection}

By leveraging the convolutional occupancy network~\cite{peng2020convolutional}, IAE demonstrates a robust ability to handle complex 3D scenes. We build the encoder with the VoteNet~\cite{qi2019deep} architecture and fine-tune the weights using ScanNet~\cite{dai2017scannet} detection labels after self-supervised pre-training on the same dataset. As shown in Table~\ref{tab:detection}, IAE outperforms random weight initialization with 18.8\% and 4.9\% relative improvements in mAP@0.5 and mAP@0.25, respectively. In addition to ScanNet, we further evaluate IAE on the more challenging SUN RGB-D dataset~\cite{song2015sun} and observe similar performance improvements.

Recently, Wang et al.~\cite{wang2022cagroup3d} propose the CAGroup3D model, which outperforms the commonly used VoteNet~\cite{qi2019deep} by a significant margin. Table~\ref{tab:detection} demonstrates that incorporating CAGroup3D as the IAE encoder further boosts the performance over the primitive CAGroup3D model.

\paragraph{Indoor 3D Semantic Segmentation}

\input{chapters/IAE/tables/seg}

On the Stanford Large-Scale 3D Indoor Spaces (S3DIS) dataset~\cite{armeni20163d}, we assess how well IAE can transfer its knowledge in self-supervised pre-training to scene-level semantic segmentation. S3DIS consists of 3D point clouds collected from 6 different large-area indoor environments, annotated with per-point categorical labels. Following Qi et al.~\cite{qi2017pointnet}, we divide each room instance into 1m $\times$ 1m blocks. We randomly sub-sample 4,096 points from the original point cloud as the encoder input and use 6-fold cross-validation during fine-tuning. As shown in Table~\ref{tab:semseg}, IAE significantly improves baseline approaches when the encoder is constructed from the same DGCNN~\cite{wang2019dynamic} architecture, with an 0.8\% absolute improvement in overall accuracy and 2.2\% improvement in mIoU. When integrated with PointNeXt~\cite{qian2022pointnext}, the latest model design in the field, IAE is still able to outperform the from-scratch training.

\paragraph{Cross-Domain Generalizability}

\input{chapters/IAE/tables/cross}

Up to this point, our discussion on scene-level fine-tuning assumes the pre-training process uses scene-level data as well. Whether synthetic object-level pre-training helps real-world scene-level fine-tuning remains an open debate in 3D vision. Xie et al.~\cite{xie2020pointcontrast} suggest pre-training on ShapeNet~\cite{chang2015shapenet} may have a detrimental effect on ScanNet~\cite{dai2017scannet} fine-tuning. In contrast, Huang et al.~\cite{huang2021spatio} demonstrate that with a sufficiently powerful encoder, object-level pre-training is actually beneficial regardless of whether the downstream task is at the object level or the scene level. As shown in Table~\ref{tab:cross}, with ShapeNet pre-training, IAE achieves a mAP@0.25 of 67.1 on the SUN RGB-D dataset~\cite{song2015sun} for object detection. Albeit not as good as scene-level pre-training, object-level pre-training demonstrates a noticeable improvement over training-from-scratch (66.8 mAP@0.25). On ModelNet40~\cite{wu20153d}, we use linear evaluation to determine the effect on scene-level classification. While it is meaningless to fit an SVM using an encoder with randomly initialized weights, Table~\ref{tab:cross} demonstrates that pre-training the IAE encoder on both ScanNet and ShapeNet leads to state-of-the-art classification accuracy under DGCNN~\cite{wang2019dynamic} backbone ({\em c.f.}, Table~\ref{tab:cls}). Empirically, we present additional evidence to support Huang et al.'s claim in the open debate.

\paragraph{Label Efficiency Training}
Pre-training helps models to be fine-tuned with a small amount of labeled data. We study the labeling efficiency of our model on 3D object detection by varying the portion of supervised training data. Results can be found in Figure~\ref{fig:effi}.
We take the VoteNet~\cite{qi2019deep} as the training network and use 20\%, 40\%, 60\%, and 80\% of the training data from ScanNet and SUN RGB-D datasets. We observe that our pre-training method gives larger gains when the labeled data is less. With only ~60\% training data on ScanNet/SUN RGB-D, our model can get similar performance compared with using all training data from scratch. This suggests our pre-training can help the downstream task to obtain better results with less data.

\input{chapters/IAE/figures_tex/efficient}

\input{chapters/IAE/figures_tex/tsne}
\paragraph{Embedding Visualization}
We visualize the learned features of our model and baseline approaches in Figure~\ref{fig:tsne}. We compare with FoldingNet~\cite{yang2018foldingnet}, OcCo~\cite{wang2020unsupervised}, and a sanity-check baseline, random initialization. Random initialization use randomly initialized network weights to obtain the embedding, and its performance explains the network prior. The embeddings for different categories in the ModelNet10 dataset are shown using t-SNE dimension reduction. Empirically, we observe that our pre-trained model provides a cleaner separation between different shape categories than FoldingNet~\cite{yang2018foldingnet}, OcCo~\cite{wang2020unsupervised}, and random initialization.

\input{chapters/IAE/figures_tex/chamfer_time}

\subsubsection{Ablation Study}
\label{sec:ablation}
We proceed with an ablation study to understand the different design choices we made in our experiments.

\noindent\textbf{Explicit vs. Implicit Decoders.} 
Under otherwise identical settings, we conduct two sets of parallel experiments to compare conventional explicit decoders with our proposed implicit decoders. Specifically, we choose three state-of-the-art explicit models, including FoldingNet~\cite{yang2018foldingnet}, OcCo~\cite{wang2020unsupervised}, and SnowflakeNet~\cite{xiang2021snowflakenet}, and compare them with two different implicit models, including Occupancy Network~\cite{mescheder2019occupancy}, and Convolutional Occupancy Network~\cite{peng2020convolutional}. We pre-train all models on ShapeNet~\cite{chang2015shapenet} and report the linear SVM classification accuracies on ModelNet40~\cite{wu20153d} under DGCNN~\cite{wang2019dynamic} backbone. As shown in Table~\ref{tab:ablation}~(a), implicit decoders consistently improve all explicit methods. Furthermore, we performed a comparison on real data by pre-training the models on ScanNet and reporting the SUN RGB-D detection result. As shown in Table~\ref{tab:ablation}~(b), the implicit decoder outperforms the explicit methods.

\input{chapters/IAE/tables/ablation_3table.tex}

\noindent\textbf{Choice of Implicit Representations.}
Among the signed distance function~\cite{michalkiewicz2019implicit, park2019deepsdf}, unsigned distance function~\cite{chibane2020neural}, and occupancy grid~\cite{chen2019learning, mescheder2019occupancy, peng2020convolutional}, what is the implicit representation of choice in self-supervised representation learning? As shown in Table~\ref{tab:ablation}~(c), while they all outperform the explicit representation, the signed distance function demonstrates the highest linear evaluation accuracy on ModelNet40~\cite{wu20153d} classification with ShapeNet~\cite{chang2015shapenet} pre-training. When ground-truth signed distance functions are unavailable due to the lack of high-quality meshes, the unsigned distance function becomes the second-best choice.

\input{chapters/IAE/figures_tex/analysis2}

\subsubsection{Analysis}
\label{sec:analysis}
In this section, we show more experiment analysis between explicit autoencoders and implicit autoencoders.

\noindent\textbf{Computational Cost for Loss Function.}
As mentioned in Section~\ref{sec:challenge}, existing explicit autoencoders lack the ability to utilize dense point clouds from open 3D datasets, but sub-sampling inevitably introduces information loss and additional sampling variation. To verify this issue, we implemented the OcCo~\cite{wang2021unsupervised} framework and tested its computation time and GPU memory utilization as the ground truth point cloud density increased.
In Figure~\ref{fig:chamfer}, we observe that by applying Chamfer loss and training the model on a single Tesla V100 GPU with a batch size of 1, the training time increases to over 10.53 hours per epoch, and the GPU memory usage increases to 26.8 GiB as the number of points for the ground-truth point cloud increased to 32k. Since pre-training usually takes at least 200 epochs, this cost is unacceptable.
In comparison, IAE only takes 0.3 hours per epoch with 6 GiB GPU memory.

\noindent\textbf{Sampling Variation Problem.}
We conduct additional experiments to gain insights into the internal mechanism of IAE. Specifically, we pre-train both IAE and an explicit autoencoder, as discussed in Section~\ref{sec:tasks:shape}, on the ShapeNet dataset~\cite{chang2015shapenet}. Next, we randomly sample 100 different point clouds from 10 characteristic shapes and use the pre-trained encoders to predict latent codes.
We use T-SNE~\cite{van2008visualizing} to obtain embeddings of the latent codes in $\mathbb{R}^2$. As visualized in Figure~\ref{fig:analysis}~(a), both the implicit and explicit models are successful in mapping the shapes into 10 distinctive clusters. However, the implicit clusters are noticeably smaller in size, with an average radius of only 56.3\% of the explicit clusters (Figure~\ref{fig:analysis}~(b)). This result supports our claim that IAE is less sensitive to sampling variations and is able to learn more generalizable features about the true 3D geometry. 

%% file: chapters/IAE/tables/cls_2.tex
\begin{landscape}
\begin{table*}[h]
    \centering
    \setlength\tabcolsep{11pt}
    \scriptsize
    \captionof{table}
    {Classification results on ScanObjectNN and ModelNet40 datasets. The model parameters number (\#Params), GFLOPS, and overall accuracy (\%) are reported. To establish a fair comparison, we present the performance of IAE using two distinct architectures: DGCNN and Transformer-like Point-M2AE. `w/o mesh' indicates that pre-training does not involve mesh data.}
    \begin{tabular}{lccccccc}
        \toprule
        \multirow{2}{*}{Method} & \multirow{2}{*}{\#Params(M)} & \multirow{2}{*}{GFLOPS} & \multicolumn{3}{c}{ScanObjectNN} & \multicolumn{2}{c}{ModelNet40} \\ 
     \cmidrule(lr){4-6}   \cmidrule(lr){7-8}   & & & OBJ\_BG & OBJ\_ONLY & PB\_T50\_RS & 1k P & 8k P \\
        \midrule
        \multicolumn{8}{c}{\textit{Supervised Learning Only}} \\
        \midrule
         PointNet~\cite{qi2017pointnet++} & 3.5 & 0.5 & 73.3 & 79.2 & 68.0 & 89.2 & 90.8 \\
        PointNet++~\cite{qi2017pointnet} & 1.5 & 1.7 & 82.3 & 84.3 & 77.9 & 90.7 & 91.9 \\
        PointCNN~\cite{li2018pointcnn} & 0.6 & - & 86.1 & 85.5 & 78.5 & 92.2 & - \\
        KPConv~\cite{thomas2019kpconv} & - & - & 86.4 & 84.1 & 80.2 & 92.9 & - \\
        GBNet~\cite{qiu2021geometric} & 8.8 & - & - & - & 81.0 & 93.8 & -\\
        PointMLP~\cite{ma2022rethinking} & 12.6 & 31.4 & - & - & 85.7 & 94.1 & -\\
        PointNeXt~\cite{qian2022pointnext} & 1.4 & 3.6 & - & - & 88.1 & 93.2 & -\\
        \midrule
        \multicolumn{8}{c}{\textit{with Self-Supervised Representation Learning} (FULL)} \\
        \midrule
        DGCNN~\cite{wang2019dynamic} & 1.8 & 2.4 & 82.8 & 86.2 & 78.1 & 92.9 & 93.1 \\
        JigSaw~\cite{sauder2019self} & 1.8 & 2.4 & - & - & 83.5 & 92.6 & - \\
        STRL~\cite{huang2021spatio} & 1.8 & 2.4 & - & - & - & 93.1 & -\\
        OcCo~\cite{wang2021unsupervised} & 1.8 & 2.4 & 88.2 & 87.5 & 84.3 & 93.0 & -\\
         IAE (DGCNN) & 1.8 & 2.4 & \textbf{90.2} & \textbf{89.0} & \textbf{85.6} & \textbf{94.2} & \textbf{94.2} \\
        \midrule
        Transformer~\cite{yu2021point} & 22.1 & 4.8 & 83.0 & 84.1 & 79.1 & 91.4 & 91.8\\
        Point-BERT~\cite{yu2021point} & 22.1 & 4.8 & 89.3 & 88.1 & 84.3 & 93.2 & 93.8\\
        Point-MAE~\cite{pang2022masked} & 22.1 & 4.8 & 90.0 & 88.3 & 85.2 & 93.8 & 94.0 \\
        Point-M2AE~\cite{zhang2022point} & 15.3 & 3.6 & 91.2 & 88.8 & 86.4 & 94.0 & - \\      
        IAE (M2AE) w/o mesh & 15.3 & 3.6 & 92.3 & 91.2 & 88.0 & 94.1 & 94.2 \\
        IAE (M2AE) & 15.3 & 3.6 & \textbf{92.5} & \textbf{91.6} & \textbf{88.2} & \textbf{94.2} &\textbf{ 94.3} \\
        \midrule
        \multicolumn{8}{c}{\textit{with Self-Supervised Representation Learning} (LINEAR)} \\
        \midrule
        JigSaw~\cite{sauder2019self} & 1.8 & 2.4 & - & - & 59.5 & 84.1 & - \\
        STRL~\cite{huang2021spatio} & 1.8 & 2.4 & - & - & 77.9 & 90.9 & -\\
        OcCo~\cite{wang2021unsupervised} & 1.8 & 2.4 & - & - & 78.3 & 89.7 & -\\
        CrossPoint~\cite{afham2022crosspoint} & 1.8 & 2.4 & - & - & 81.7 & 91.2 & -\\
        IAE (DGCNN) & 1.8 & 2.4 & - & - & \textbf{83.0} & \textbf{92.1} & -\\
        \midrule
        Point-M2AE~\cite{zhang2022point} & 15.3 & 3.6 & - & - & 83.1 & 92.9 & - \\
        IAE (M2AE) w/o mesh & 15.3 & 3.6 & - & - & 84.3 & 93.1 & - \\
        IAE (M2AE) & 15.3 & 3.6 & - & - & \textbf{84.4} & \textbf{93.2} & - \\
        \bottomrule
    \end{tabular}
    \label{tab:cls}
\vspace{-3mm}
\end{table*}
\end{landscape}

%% file: chapters/IAE/tables/detection.tex
\begin{table*}[h]
    \centering
   \setlength\tabcolsep{15pt}
    \begin{tabular}{lcccc}
    \toprule
     Method  & \multicolumn{2}{c}{ScanNet} & \multicolumn{2}{c}{SUN RGB-D} \\
        \cmidrule(lr){2-3}   \cmidrule(lr){4-5}  & $\text{AP}_{25}$ & $\text{AP}_{50}$ & $\text{AP}_{25}$ & $\text{AP}_{50}$ \\
        \midrule
        VoteNet~\cite{qi2019deep}  &  58.6 & 33.5 & 57.7 & 32.9 \\
        3D-MPA~\cite{engelmann20203d} & 64.2 & 49.2 & - & - \\
        MLCVNet~\cite{xie2020mlcvnet} & 64.5 & 41.4 & 59.8 & - \\
        BRNet~\cite{cheng2021back} & 66.1 & 50.9 & 61.1 & 43.7 \\
        3DETR~\cite{misra2021end} & 65.0 & 47.0 & 59.1 & 32.7 \\
        GroupFree~\cite{liu2021group} & 69.1 & 52.8 & 63.0 & 45.2 \\
        FCAF3D~\cite{rukhovich2021fcaf3d} & 71.5 & 57.3 & 64.2 & 48.9 \\
        CAGroup3D~\cite{wang2022cagroup3d} & 75.1 & 61.3 & 66.8 & 50.2 \\
        \midrule
        STRL~\cite{huang2021spatio} & 59.5 & 38.4 & 58.2 & 35.0 \\
        RandomRooms~\cite{rao2021randomrooms} & 61.3 & 36.2 & 59.2 & 35.4 \\
        PointContrast~\cite{xie2020pointcontrast} & 59.2 & 38.0 & 57.5 & 34.8 \\
        DepthContrast\protect\footnotemark~\cite{zhang2021self} & 62.1 & 39.1 & 60.4 & 35.4 \\
        Point-M2AE~\cite{zhang2022point} & 66.3 & 48.3 & - & - \\
        IAE (VoteNet) & 61.5 & 39.8 & 60.4 & 36.0 \\
        IAE (CAGroup3D) & \textbf{76.1} & \textbf{62.0} & \textbf{67.6} & \textbf{51.3}\\
        \bottomrule
    \end{tabular}
    \caption{{3D Object Detection Results. We fine-tune our pre-trained model on ScanNetV2~\cite{dai2017scannet} and SUN-RGBD~\cite{song2015sun} validation sets using VoteNet~\cite{qi2019deep} and CAGroup3D~\cite{wang2022cagroup3d}. We show mean Average Precision~(mAP) across all semantic classes with 3D IoU thresholds of 0.25 and 0.5. Methods in the second section denote self-supervised methods.}}
    \label{tab:detection}
\end{table*}

%% file: chapters/IAE/tables/seg.tex
\begin{table}[t]
    \setlength\tabcolsep{15pt}
    \centering
    \begin{tabular}{lcccc}
        \toprule
     Method  & \multicolumn{2}{c}{ScanNet} & \multicolumn{2}{c}{SUN RGB-D} \\
        \cmidrule(lr){2-3}   \cmidrule(lr){4-5}  & $\text{AP}_{25}$ & $\text{AP}_{50}$ & $\text{AP}_{25}$ & $\text{AP}_{50}$ \\
        \midrule
        VoteNet~\cite{qi2019deep}  &  58.6 & 33.5 & 57.7 & 32.9 \\
        3D-MPA~\cite{engelmann20203d} & 64.2 & 49.2 & - & - \\
        MLCVNet~\cite{xie2020mlcvnet} & 64.5 & 41.4 & 59.8 & - \\
        BRNet~\cite{cheng2021back} & 66.1 & 50.9 & 61.1 & 43.7 \\
        3DETR~\cite{misra2021end} & 65.0 & 47.0 & 59.1 & 32.7 \\
        GroupFree~\cite{liu2021group} & 69.1 & 52.8 & 63.0 & 45.2 \\
        FCAF3D~\cite{rukhovich2021fcaf3d} & 71.5 & 57.3 & 64.2 & 48.9 \\
        CAGroup3D~\cite{wang2022cagroup3d} & 75.1 & 61.3 & 66.8 & 50.2 \\
        \midrule
        STRL~\cite{huang2021spatio} & 59.5 & 38.4 & 58.2 & 35.0 \\
        RandomRooms~\cite{rao2021randomrooms} & 61.3 & 36.2 & 59.2 & 35.4 \\
        PointContrast~\cite{xie2020pointcontrast} & 59.2 & 38.0 & 57.5 & 34.8 \\
        DepthContrast\protect\footnotemark~\cite{zhang2021self} & 62.1 & 39.1 & 60.4 & 35.4 \\
        Point-M2AE~\cite{zhang2022point} & 66.3 & 48.3 & - & - \\
        IAE (VoteNet) & 61.5 & 39.8 & 60.4 & 36.0 \\
        IAE (CAGroup3D) & \textbf{76.1} & \textbf{62.0} & \textbf{67.6} & \textbf{51.3}\\
        \bottomrule
    \end{tabular}
    \captionof{table}{Semantic Segmentation Results on S3DIS~\cite{armeni20163d} 6-Fold. We show Overall Accuracy (OA) and mean Intersection over Union~(mIoU) across six folds.}
    \label{tab:semseg}
\end{table}

%% file: chapters/IAE/tables/cross.tex
\begin{table}[h]
    \setlength\tabcolsep{10pt}
    \centering
    \begin{tabular}{lcc}
        \toprule
        Task & Pre-train & Acc/$\text{AP}_{25}$ \\
        \midrule
        \multirow{2}{*}{Object detection} & ScanNet & \textbf{65.0} \\
        & ShapeNet & 64.6 \\
        \midrule
        \multirow{2}{*}{MN40 Linear} & ScanNet & 91.3\% \\
        & ShapeNet & \textbf{92.1\%} \\
        \bottomrule
    \end{tabular}
    \captionof{table}{Cross-Domain Generalizability between ShapeNet~\cite{chang2015shapenet} and ScanNet~\cite{dai2017scannet}. For the 3D object detection task, we report mAP at IoU=0.25 on the SUN RGB-D dataset~\cite{song2015sun}. For ModelNet40~\cite{wu20153d} linear evaluation, we report the classification accuracy.}
    \label{tab:cross}
\vspace{-10pt}
\end{table}

%% file: chapters/IAE/figures_tex/efficient.tex
\begin{figure}[t]
    \centering
    \includegraphics[width=\columnwidth]{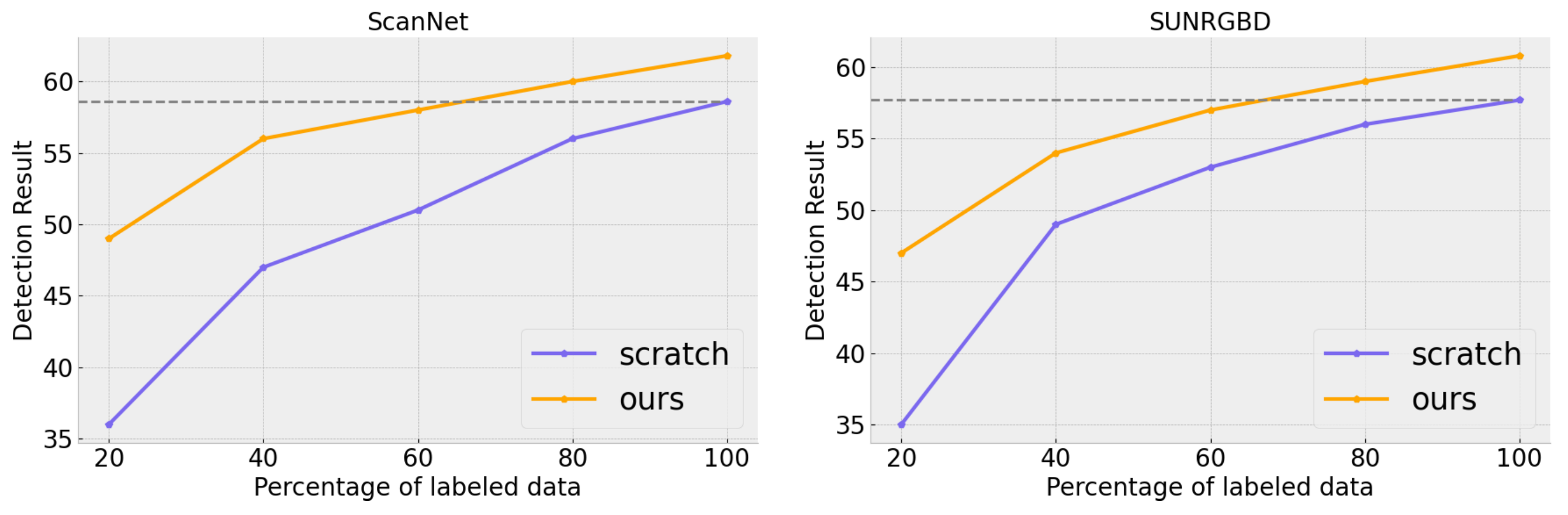}
    \caption{Label efficiency training. We pre-train our model on ScanNet and then fine-tune on ScanNet and SUN RGB-D separately. During fine-tuning, different percentages of labeled data are used. Our pre-training model outperforms training from scratch and achieves nearly the same result with only 60\% labeled data.}
    \label{fig:effi}
\end{figure}

%% file: chapters/IAE/figures_tex/tsne.tex
\begin{figure}[h]
    \centering
    \includegraphics[width=\columnwidth]{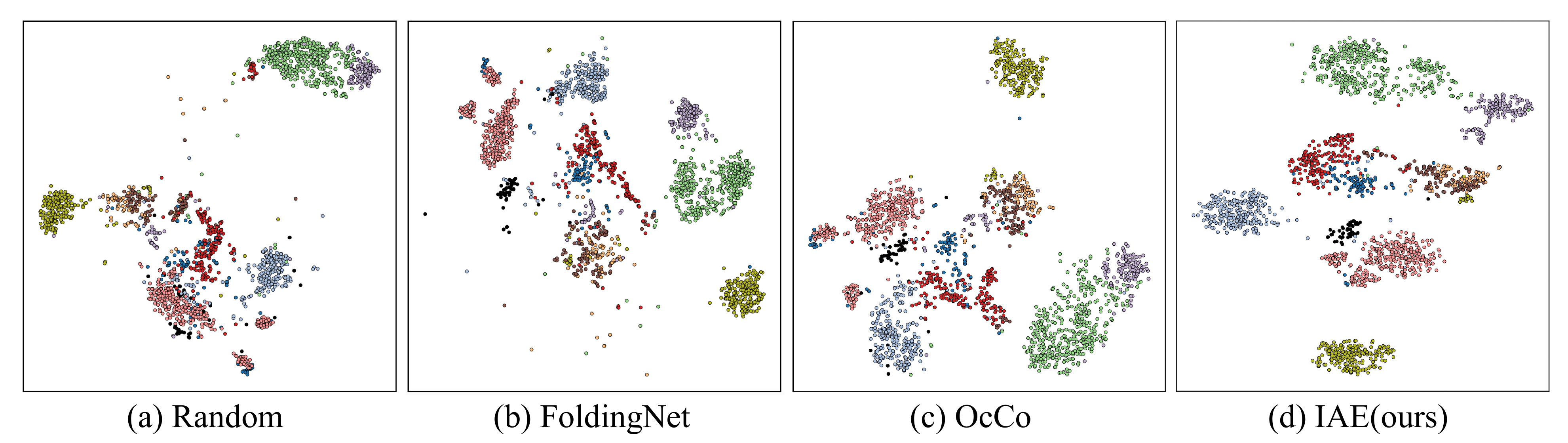}
    \caption{Visualization of learned features. We visualize the learned features for each sample in ModelNet10 using t-SNE. All the models use DGCNN as the encoder backbone. (a) uses random initialization. (b), (c), (d) are pre-trained on ShapeNet.}
    \label{fig:tsne}
\end{figure}

%% file: chapters/IAE/figures_tex/chamfer_time.tex
\begin{figure}[t]
    \centering
    \includegraphics[width=1.\columnwidth]{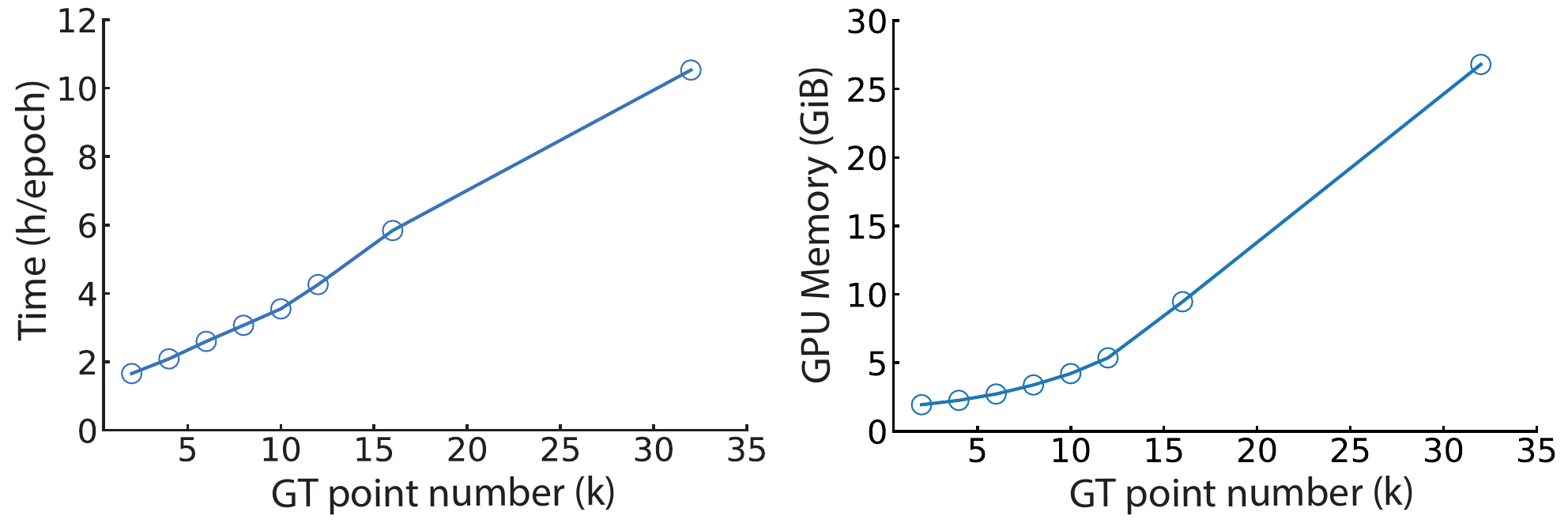}
    \caption{Computation time and GPU memory utilization of explicit autoencoder. We use the standard explicit autoencoder framework, OcCo~\cite{wang2021unsupervised}, and take DGCNN~\cite{wang2019dynamic} as the encoder backbone. We use Chamfer Distance as the loss function.}
    \label{fig:chamfer}
\end{figure}


%% file: chapters/IAE/tables/ablation_3table.tex
\begin{table*}[h]
\begin{minipage}[b]{.5\linewidth}
    \centering
    \begin{tabular}{l|cc}
        \toprule
         & Method & MN40 \\
        \midrule
        \multirow{3}{*}{EAE} &  FoldingNet & 90.1\% \\
        & OcCo & 89.7\%  \\
        & Snowflake &  89.9\% \\
        \midrule
        \multirow{2}{*}{IAE} & OccNet & 91.5\% \\
        & ConvONet &  \textbf{92.1\%} \\
        \bottomrule
    \end{tabular}
    \caption*{{(a) Different decoders}}
\end{minipage}
\begin{minipage}[b]{.5\linewidth}
    \centering
    \begin{tabular}{l|cc}
        \toprule 
         & Function & MN40 \\
        \midrule
       EAE & PC & 90.1\% \\
        \midrule
        \multirow{4}{*}{IAE} & Occ & 91.3\% \\
        & UDF & 91.7\% \\
        & SDF & \textbf{92.1\%} \\
        \bottomrule
    \end{tabular}
    \caption*{{(b) Different implicit functions}}
\end{minipage}
\begin{minipage}[b]{\linewidth}
    \centering
    \begin{tabular}{l|cc}
    \toprule
     & Method & SUN RGB-D \\
        \midrule
        \multirow{3}{*}{EAE} &  FoldingNet & 58.2 \\
        & OcCo & 58.4  \\
        & Snowflake & 58.1 \\
        \midrule
        IAE & ConvOccNet &  \textbf{60.4} \\
        \bottomrule
    \end{tabular}
    \caption*{{(c) Different decoders on real data task}}
\end{minipage}
\caption{{(a) Ablation Study on Different Decoder Models on shape-level ModelNet40~\cite{wu20153d} linear evaluation. Our implicit autoencoders (IAE) consistently outperform their explicit counterparts (EAE). (b) Ablation Study on Implicit Functions. For explicit representation, we use FoldingNet~\cite{yang2018foldingnet} as the decoder and point cloud (PC) at the output. Three implicit representations show consistent improvements over the explicit representation. (c) Ablation Study on Different Decoder Models on real data task, mAP at IoU=0.25 on SUN RGB-D~\cite{song2015sun} detection. Our approach using the convolutional occupancy network shows consistent improvements.}}
\vspace{-10pt}
\label{tab:ablation}
\end{table*}

%% file: chapters/IAE/figures_tex/analysis2.tex
\begin{figure}[t]
    \centering
    \begin{overpic}[width=\linewidth]{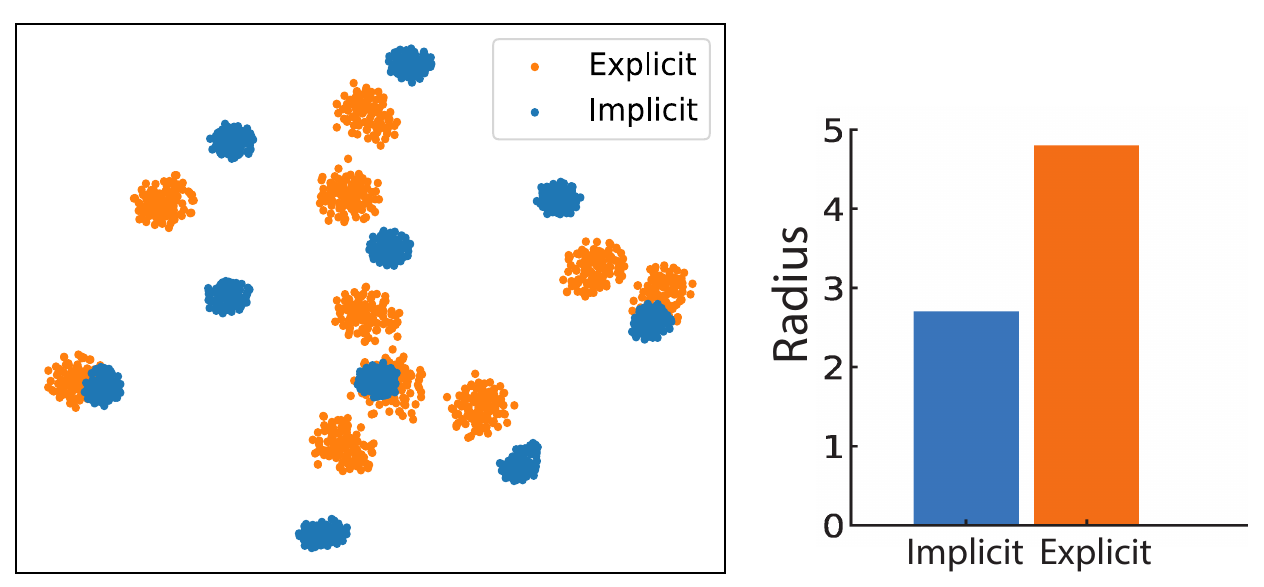}
    \put(27,-3.5){(a)}
    \put(80,-3.5){(b)}
    \end{overpic}
    \vspace{1pt}
    \caption{Experiment analysis on Explicit and Implicit Autoencoders. (a) T-SNE visualization for encoder latent codes of IAE (blue) and EAE (orange). The IAE clusters are noticeably smaller. (b) The average cluster radius for IAE and EAE in (a).}
    \label{fig:analysis}
    \vspace{-10pt}
\end{figure}

%% file: chapters/IAE/sections/08_theory.tex
\subsection{Analysis of IAE under a Linear AE model}
\label{Section:Motivation}

We proceed to gain more insights of IAE under a linear AE model. 


Specifically, suppose we have $N>n$ points $\bs{x}_i\in R^n, 1\leq i \leq N$. Without losing generality, we assume $\bs{x}_i$ lies on a low-dimensional linear space $\set{L}$ of dimension $m < n$. These data points are used to model the underlying 3D models. Now let us perturb each point $\bs{x}_i' = \bs{x}_i + \epsilon_i$, where $\epsilon_i$ is used to model sampling variations. We assume $\epsilon_i\in \set{L}^{\perp}$, meaning they encode variations that are orthogonal to variations among the underlying 3D models. Denote $X:=(\bs{x}_1,\cdots, \bs{x}_N)$ and $X':=(\bs{x}_1',\cdots, \bs{x}_{N}')$.

We consider two linear autoencoding models. The first one, which is analogous to IAE, takes $\bs{x}_i'$ as input and seeks to reconstruct $\bs{x}_i$:
\begin{align*}
A^{\star},Q^{\star} &= \underset{A', Q'\in \R^{n\times m}}{\textup{argmin}} \ \sum\limits_{k=1}^{N} \|A'{Q'}^T\bs{x}_{k}'-\bs{x}_k\|^2 \quad \\
s.t. \ {Q'}^TQ'& = I_m, \ Q'\in \set{C}(X')
\end{align*}
Here $Q'$ is the encoder, and $A'$ is the decoder. $\set{C}(X')$ denotes the column space of matrix $X'$. The constraints $Q'^T Q' = I_m$ and $Q'\in \set{C}(X')$ ensure that the encoder-decoder pair is unique up to a unitary transformation in $O(m)$.  
Below we show that $Q^{\star}$ is independent of $\epsilon_k$. 
\begin{prop}
Let $Q\in \R^{n\times m}$ collect the top-$m$ eigenvectors of the convariance matrix $C = \sum\limits_{k=1}^{N}\bs{x}_k\bs{x}_{k}'$. Then 
under the assumption that $\epsilon_k\in \set{L}^{\perp},1\leq k \leq N$,
$Q^{\star} = Q.$
\label{Prop:Derivative:Reformulated:AE}
\end{prop}
Prop.~\ref{Prop:Derivative:Reformulated:AE} indicates that $Q^{\star}$ does not encode sampling variations. 

Now consider the second model where we force the encoder-decoder pair to reconstruct the original inputs, which is analogous to standard autoencoding:
\begin{align*}
{\hat{A}^{\star}, \hat{Q}^{\star}} &= \underset{A',Q'\in \R^{n\times m}}{\textup{argmin}} \ \sum\limits_{k=1}^{N} \|A'{Q'}^T\bs{x}_k'-\bs{x}_k'\|^2_{\mathcal{F}} \quad \\
s.t. \ {Q'}^TQ' &= I_m, \ Q'\in \set{C}(X')
\end{align*}
In this case, $\hat{A}^{\star}=\hat{Q}^{\star}$, and both of them are given by the top $m$ eigenvectors of $C' = \sum\bs{x}_k'{\bs{x}_{k}'}^T$. 

To quantitatively compare encoders $\hat{Q}^{\star}$ and $Q$, we need the following definition.
\begin{defn}
Consider two unitary matrices $Q_1, Q_2 \in \R^{n\times m}$ where $Q_i^TQ_i = I_m$. We define the deviation between them as 
\begin{align*}
\mathcal{D}(Q_1, Q_2) &:= Q_1 -Q_2 R^{\star}, \\ \qquad R^{\star} &= \underset{R\in O(m)}{\textup{argmin}}\|Q_1 - Q_2R\|_{\mathcal{F}}^2
\label{Eq:Matrix:Difference}    
\end{align*}
\label{Defn:Matrix:Difference}
\end{defn}
The following proposition specifies the derivatives between $\hat{Q}^{\star}$ and $\epsilon_{k}$.
\begin{prop} 
Under the assumption that $\epsilon_k\in \set{L}^{\perp},1\leq k \leq N$, we have
\begin{equation}
\frac{\partial \mathcal{D}(\hat{Q}^{\star},Q)}{\partial \epsilon_{ki}} = (I_n - QQ^T)(\bs{e}_i\bs{x}_k^T)Q\Lambda^{+}
\label{Eq:Derivative:Q:Hat}
\end{equation}
where $\Lambda =\textup{diag}(\lambda_1,\cdots, \lambda_m)$ is a diagonal matrix that collects the top eigenvalues of $C$ that correspond to $Q$. $\bs{e}_k$ is the $k$-th standard basis vector.
\label{Prop:Derivative:Standard:AE}
\end{prop}
In other word, $\hat{Q}^{\star}$ is sensitive to $\epsilon_k$. Therefore, it encodes sampling variations. This theoretical analysis under a simplified setting suggests that IAE may potentially learn more robust representations.

%% file: chapters/IAE/sections/06_conclusion.tex

\input{chapters/IAE/figures_tex/supp_compare}
\subsection{Discussion}
\label{sec:supp:discuss}

\subsubsection{What is the difference between data augmentation and sampling variation?}

Sampling variation refers to the fact that different point cloud samples of the same 3D shape contain different noises induced from various sources, such as intrinsic noises from sensors and interference from the environment. In the explicit autoencoding paradigm, the encoder is required to encode not only the 3D geometry but also information about the specific discrete sampling of the 3D shape into the latent code, which can lead to sampling variation. This is because the decoder must reconstruct a point cloud that matches the original point cloud perfectly. For example, as shown in Figure~\ref{fig:supp_compare}, given a 3D shape and two randomly selected samples, the explicit autoencoder forces the decoder to output the same sample as the input. If the target sample changes, the loss increases due to incorrect mapping, even though the different samples represent the same shape. Thus, the explicit autoencoding paradigm must learn the mapping that includes sampling variation. In contrast, the implicit autoencoder does not face this problem because different samples map to the same target, which is the implicit representation of the 3D shape.

Data augmentation is a standard technique for point-cloud learning, including rotation, translation, scaling, and sub-sampling. IAE uses all of these standard data augmentation methods during pre-training, as other approaches do. For instance, IAE uses different sub-samples of the same 3D shape as input, but all of these sub-samples share the same target, which is the implicit function of the 3D shape.

\subsubsection{Is it a fair comparison with existing self-supervised learning approaches?}

In this subsection, we discuss the fairness of the comparison between the Implicit Autoencoder (IAE) and existing self-supervised learning approaches. Our approach differs from existing methods in that we use dense point clouds (50k points) or mesh data from ShapeNet to generate the implicit function label, while existing approaches utilize sparse, sub-sampled point clouds (i.e., 1k $\sim$ 2k points) for pre-training. We argue that the point of pre-training is to extract as much information as possible from all available data. Therefore, it is natural to include dense point clouds or mesh data in the pre-training dataset. However, explicit autoencoders lack the ability to efficiently exploit the density, as demonstrated in the experiment analysis in Section 4.5 of the main paper. Hence, we consider our approach to have an advantage rather than an unfair comparison.

We also emphasize that we use the same setting as existing self-supervised learning approaches for downstream task training to ensure a fair comparison. Therefore, we are confident that the comparison between our approach and existing approaches is fair.

\subsection{Conclusion}
\label{Section:Conclusions}
We present Implicit AutoEncoder (IAE), a simple yet effective model for point-cloud self-supervised representation learning. Unlike conventional point-cloud autoencoders, IAE exploits the implicit representation as the output of the decoder. IAE prevents latent space learning from being distracted by sampling variations and encourages the encoder to capture generalizable features from the true 3D geometry. Extensive experiments demonstrate that IAE achieves considerable improvements over a wide range of downstream tasks, including shape classification, linear evaluation, object detection, and indoor semantic segmentation. In the future, we plan to extend IAE to support not only hand-crafted but also trainable implicit representations that can be jointly learned with the autoencoder pipeline.

%% file: chapters/IAE/figures_tex/supp_compare.tex
\begin{figure}[h]
    \centering
    \includegraphics[width=\columnwidth]{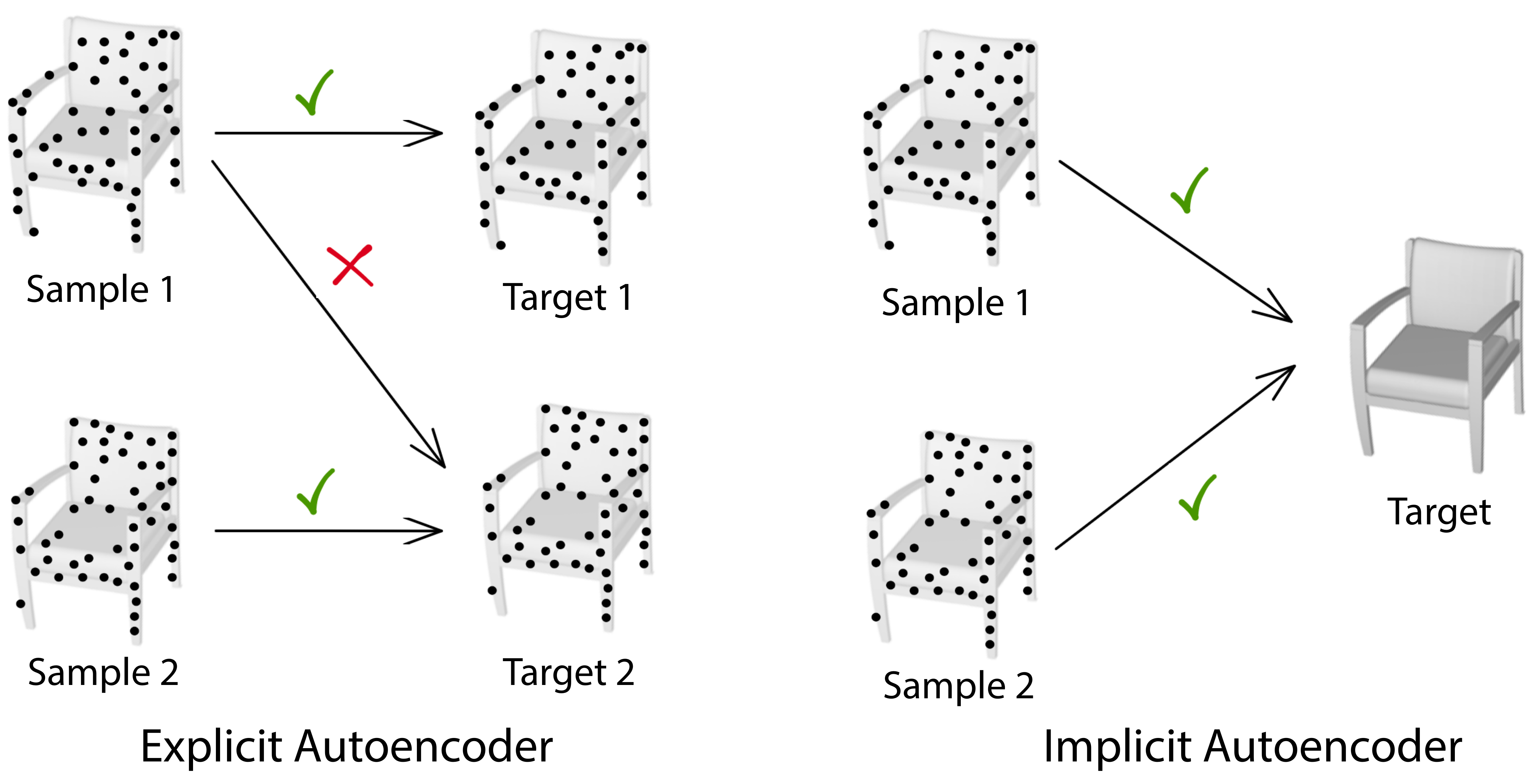}
    \caption{Comparison between Explicit Autoencoder and Implicit Autoencoder. Given a 3D shape, we randomly take two samples. For Implicit Autoencoder, we use 3D CAD model to represent the implicit function target for better visualization.}
    \label{fig:supp_compare}
\end{figure}

%% file: chapters/MaskFeat3D/MaskFeat3D.tex
\section{3D Feature Prediction for Masked-AutoEncoder-Based Point Cloud Pretraining}
\label{sec:maskfeat3d}

\renewcommand{\thefootnote}{\color{red}\fnsymbol{footnote}}
\footnote[6]{The content of this section is based on my publication ``3D Feature Prediction for Masked-AutoEncoder-Based Point Cloud Pretraining''~\cite{yan20233d} in International Conference on Learning Representations (ICLR) 2024. I am the first author of the publication.}

Masked autoencoders (MAE) have recently been introduced to 3D self-supervised pretraining for point clouds due to their great success in NLP and computer vision. Unlike MAEs used in the image domain, where the pretext task is to restore features at the masked pixels, such as colors, the existing 3D MAE works reconstruct the missing geometry only, i.e, the location of the masked points. In contrast to previous studies, we advocate that point location recovery is inessential and restoring intrinsic point features is much superior. To this end, we propose to ignore point position reconstruction and recover high-order features at masked points including surface normals and surface variations, through a novel attention-based decoder which is independent of the encoder design. We validate the effectiveness of our pretext task and decoder design using different encoder structures for 3D training and demonstrate the advantages of our pretrained networks on various point cloud analysis tasks.

\input{chapters/MaskFeat3D/Sections/01-Introduction}

\input{chapters/MaskFeat3D/Sections/02-RelatedWork}

\input{chapters/MaskFeat3D/Sections/03-Method}
\input{chapters/MaskFeat3D/Sections/04-Evaluation}

\input{chapters/MaskFeat3D/Sections/06-Conclusion}

%% file: chapters/MaskFeat3D/Sections/01-Introduction.tex
\subsection{Introduction}
\label{sec:intro}

\input{chapters/MaskFeat3D/Figures/fig-arch-compare}

Self-supervised pretraining has recently gained much attention. It starts from a pretext task trained on large unlabeled data, where the learned representation is fine-tuned on downstream tasks. This approach has shown great success in 2D images~\cite{Chen2020,Grill2020,He2020,bao2022beit,HeCXLDG22,zhou2021ibot, zhuang2021unsupervised, zhuang2019self} and natural language processing (NLP)~\cite{Devlin2018,Brown2020} . Recently, people started looking into self-supervised pretraining on point cloud data due to its importance in 3D analysis and robotics applications.  

An important self-supervised pretraining paradigm --- \emph{masked signal modeling} (MSM), including BERT~\cite{Bugliarello2021}, BEiT~\cite{bao2022beit}, and masked autoencoders (MAE)~\cite{HeCXLDG22}, has recently been adopted to 3D domains.  MSM has a simple setup: a randomly-masked input is fed to the encoder, and a decoder strives to recover the signal at the masked region. MSM is highly scalable and exhibits superior performance in many downstream vision and NLP tasks, outperforming their fully supervised equivalents. Additionally, it does not require extensive augmentation, which is essential and critical to another self-supervised pretraining paradigm --- contrastive learning. In images, a mask refers to a randomly selected portion of the pixels, and the pixel colors or other pixel features in the masked region are to be reconstructed by the decoder.

For 3D point clouds, the PointBERT approach~\cite{Yu_2022_CVPR} masks point patches and recovers patch tokens that are pretrained by a point cloud Tokenizer. As reconstruction features are associated with patches of points, the learned features at the point level are less competitive. MAE-based pretraining schemes~\cite{Pang2022MaskedAF,hess2022masked,zhang2022masked,Liu2022maskdis} tackle this problem by point-wise pretext tasks. However, their decoders are designed to recover the positions of the masked points in Cartesian coordinates or occupancy formats (Fig.~\ref{fig:arch-compare}-left). \textbf{These designs make an intrinsic difference from 2D MSMs, where there is no need to recover masked pixel locations.} This key difference makes MSM pay more attention to capturing the irregular and possibly noisy point distribution and ignore the intrinsic surface features associated with points, which are essential for 3D point cloud analysis.  

In the presented work, we propose to recover intrinsic point features, i.e., point normals, and surface variations~\cite{PC:Simp:2002} at masked points, where point normals are first-order surface features and surface variations are related to local curvature properties.  
We clearly demonstrate that the recovery of high-order surface point features, not point locations, is the key to improving 3D MSM performance. Learning to reconstruct high-order geometric features forces the encoder to extract distinctive and representative features robustly that may not be captured by learning to reconstruct point positions alone. Our study justifies the importance of designing signal recovery for 3D MSMs. It aligns 3D MSM learning with MSM development in vision, where feature modeling plays a critical role~\cite{wei2022masked}. 

To recover point signals, we design a practical attention-based decoder. This new decoder takes masked points as queries, and stacks several transformer blocks. In each block, self-attention is used to propagate context features over the masked points and cross-attention is applied to fabricate the point features with the encoder's output (As shown in Fig.~\ref{fig:arch-compare}-right and Fig.~\ref{fig:maskarch}).
This design is separable from the encoder design. Therefore, common 3D encoders, such as sparse CNNs, point-based networks, and transformer-based networks, can all be adopted to strengthen the pretraining capability. Another benefit of this decoder design is that the masked point positions are only accessible by the decoder, thus avoiding leakage of positional information in the early stage of the network, as suggested by~\cite{Pang2022MaskedAF}.

We conducted extensive ablation studies to verify the efficacy of our masked feature design and decoder. Substantial improvements over previous approaches and the generalization ability of our pretraining approach are demonstrated on various downstream tasks, including 3D shape classification, 3D shape part segmentation, and 3D object detection. 
We hope that our study can stimulate future research on designing strong MAE-based 3D backbones.

We summarize the contributions of our paper as follows:
\vspace{-5pt}
\begin{enumerate}[leftmargin=*]\setlength\itemsep{0mm}
    \item[-] We propose a novel masked autoencoding method for 3D self-supervised pretraining that predicts intrinsic point features at masked points instead of their positions.
    \item[-] We introduce a unique attention-based decoder that can generate point features without relying on any particular encoder architecture.
    \item[-] Our experiments demonstrate that restoring intrinsic point features is superior to point location recovery in terms of Point cloud MAE, and we achieve state-of-the-art performance on various downstream tasks.
\end{enumerate}

%% file: chapters/MaskFeat3D/Figures/fig-arch-compare.tex
\begin{figure}[h]
    \centering
    \begin{overpic}[width=1\linewidth]{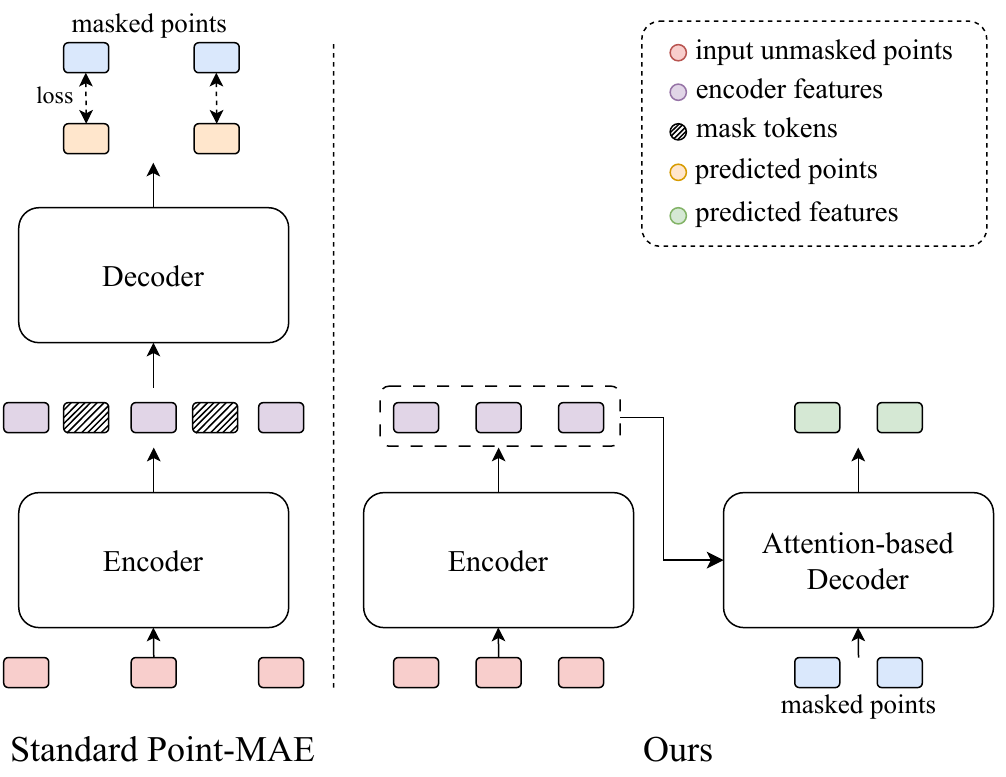}
    \end{overpic}
    \caption{Comparison of standard Point-MAE and our proposed method. Unlike standard Point-MAE that uses masked points as the prediction target, our method use a novel attention-based decoder to leverage masked points as an additional input and infer the corresponding features.}
    \label{fig:arch-compare} 
\end{figure}

%% file: chapters/MaskFeat3D/Sections/02-RelatedWork.tex
\subsection{Related Work}

\paragraph{Self-supervised pretraining in 3D} Self-supervised pretraining is an active research topic in machine learning~\cite{sslsurvey}. The early adoption of self-supervised pretraining for 3D is to use autoencoders~\cite{Yang2018a, yan2022implicit} and generative adversarial networks~\cite{Wu2016} to learn shape-level features, mainly for shape classification and retrieval tasks. Other self-supervised pretext tasks, such as clustering and registration, are also developed for 3D pretraining. Later, due to the great ability to learn features at both the instance and pixel levels in a self-supervised manner, contrastive learning~\cite{Wu2018a,Grill2020,He2020,Brown2020,Chen2021, yan2023multi} was introduced into the 3D domains to extract distinctive instance and point-wise features for various downstream tasks~\cite{Wang2021,Xie2020,Hou2021,Zhang2021}. However, contrastive learning requires data augmentation heavily to form positive or negative pairs for effective feature learning. 

\paragraph{Masked signal modeling in 3D} 
Masked signal modeling using transformer-based architectures for self-supervised learning (SSL) has shown great simplicity and superior performance. PointBERT~\cite{Yu_2022_CVPR} and PointMAE~\cite{Pang2022MaskedAF} are two such works that inherit from this idea. PointBERT partitions a point cloud into patches and trains a transformer-based autoencoder to recover masked patches' tokens. In contrast, PointMAE directly reconstructs point patches without costly tokenizer training, using Chamfer distance as the reconstruction loss.
Other works like \cite{zhang2022masked, Liu2022maskdis} and \cite{hess2022masked} explore different strategies for point cloud reconstruction or classification with masking. 
As discussed in Introduction Section, the pretext tasks of most previous works focus only on masked point locations.

\paragraph{Signal recovery in masked autoencoders}
Masked autoencoders for vision pretraining typically use raw color information in masked pixels as the target signal~\cite{HeCXLDG22}. However, Wei \etal~\cite{wei2022masked} have found that using alternative image features, such as HOG descriptors, tokenizer features, and features from other unsupervised and supervised pretrained networks, can improve network performance and efficiency. In contrast, existing 3D MAE methods have limited use of point features and struggle with predicting the location of masked points. Our approach focuses on feature recovery rather than position prediction, selecting representative 3D local features such as point normals and surface variation~\cite{PC:Simp:2002} as target features to demonstrate their efficacy. Our study allows for leveraging more advanced 3D features in 3D masked autoencoders, while further exploration of other types of 3D features~\cite{laga20183d} is left for future work.

%% file: chapters/MaskFeat3D/Sections/03-Method.tex
\subsection{Masked 3D Feature Prediction} \label{Section:Approach}

In this section, we present our masked 3D feature prediction approach for self-supervised point cloud pretraining. Our network design follows the masked autoencoder paradigm: a 3D encoder takes a point cloud whose points are randomly masked as input, and a decoder is responsible for reconstructing the predefined features at the masked points. The network architecture is depicted in Fig.~\ref{fig:maskarch}. In the following sections, we first introduce the masking strategy and 3D masked feature modeling in Sec.~\ref{subsec:mask} and~\ref{subsec:feature}, and then present our encoder and decoder design in Sec.~\ref{subsec:encoder} and~\ref{subsec:decoder}. Here, the key ingredients of our approach are the design of prediction targets and the decoder, which govern the quality of the learned features. 

\input{chapters/MaskFeat3D/Figures/fig-arch}

\subsubsection{3D Masking} \label{subsec:mask}

We follow the masking strategy proposed by PointBERT~\cite{Yu_2022_CVPR} to mask out some portions of an input point cloud and feed it to the encoder. 
Denote the input point cloud as $\mathcal{P} \in \mathbb{R}^{N \times 3}$, where $N$ is the number of points. We sample $K$ points using farthest point sampling (FPS). For each sample point, its $k$-nearest neighbor points form a point patch. For a given mask ratio $m_r, 0 < m_r < 1$, we randomly select $M$ patches and remove them from the input, where $M = \min(\lceil m_r \cdot K \rceil, K-1)$.




In the following, the masked points and the remaining points are denoted by $\mathcal{P}_M$ and $\mathcal{P}_U$, respectively.

\subsubsection{Target Feature Design} 
\label{subsec:feature}

As argued in Introduction Section, we advocate against using point locations as the reconstructed target. 
We choose to reconstruct normal and surface variation at each point, which reflect differential surface properties.
\input{chapters/MaskFeat3D/Figures/fig-vis-feature}
On the other hand, our decoder design (to be introduced in Sec.~\ref{subsec:decoder}) takes query points as input and output predicted point-wise features. Therefore, the decoder implicitly carries positional information for learning meaningful features through the encoder. 

Given a point cloud, both point normal and surface variations are defined using local principal component analysis (PCA).  We first define a covariance matrix $C_r$ over a local surface region around $\mathbf{p}$:
\begin{equation}
    C_r := \dfrac{\int_{\mathbf{x} \in \mathcal{S} \bigcap \mathbb{S}_r(\mathbf{p})}(\mathbf{p} - \mathbf{x}) (\mathbf{p} - \mathbf{x})^T \, \mathrm{d}\mathbf{x}}{\int_{\mathbf{x} \in \mathcal{S} \bigcap \mathbb{S}_r(\mathbf{p})} \mathbf{1} \cdot \,\mathrm{d}\mathbf{x}}, 
\end{equation}
where $\mathcal{S} \bigcap \mathbb{S}_r(\mathbf{p})$ is the local surface region at $\mathbf{p}$, restricted by a sphere centered at $\mathbf{p}$ with radius $r$. We set $r = 0.1$ in our case. The ablation details are shown in the supplement.

The normal $\bs{n}(\mathbf{p})$ at $\mathbf{p}$ is estimated as the smallest eigenvector of $C_r$. The sign of each normal is computed by using the approach of~\cite{DBLP:conf/siggraph/HoppeDDMS92}. 

Surface variation~\cite{PC:Simp:2002} at $\mathbf{p}$ is denoted by $\sigma_r(\mathbf{p})$, in the following form:
\begin{equation}
    \sigma_r(\mathbf{p}) = \dfrac{\lambda_1}{\lambda_1+\lambda_2+\lambda_3},
\end{equation}
where $\lambda_1\leq\lambda_2\leq\lambda_3$ are the eigenvalues of $C_r$.  Surface variation is a geometric feature that measures the local derivation at point $\mathbf{p}$ in a neighborhood of size $r$ on a given surface $\mathcal{S}$ . Its original and modified versions have been used as a robust feature descriptor for a variety of shape analysis and processing tasks, such as saliency extraction~\cite{SPPSG:SIG:2001}, curved feature extraction~\cite{10.1111:1467-8659.00675}, shape segmentation~\cite{Huang:2006:TOG, yan2021hpnet}, and shape simplification~\cite{PC:Simp:2002}.

In the limit, i.e., when $r\rightarrow 0$, $\sigma_r(\mathbf{p})$ is related to the mean curvature~\cite{Clarenz:2004:RFD}. By varying the radii of $\mathbb{S}_r$, multiscale surface variation descriptors can be constructed. In our work, we chose only single-scale surface variation for simplicity. 

Although both surface normal and surface variation are derived from local PCA, they are complementary to each other in the sense that surface normal carries first-order differential property while surface variation carries second-order differential property due to its relation to mean curvature. We visualize both features in Fig.~\ref{fig:variation} and show more examples in supplement. In Sec.~\ref{sec:ablation:study}, we show that reconstructing surface normal and surface variation leads to better learned features than reconstructing one of them. 


\paragraph{Loss function} Point normals and surface variations represent first- and second-order surface properties. Their value intervals are also bounded: \emph{surface normal has unit length; surface variation is non-negative and not greater than $\frac{1}{3}$}. Their value-bounded properties are suitable for easy minimizing the deviation from the prediction to their ground truths, compared to using unbounded features such as curvatures.  We denote the point normals and surface variations of $\mathcal{P}_M$ by $\mathcal{N}_M \in \mathbb{R}^{M \times 3}$ and $\mathcal{V}_M \in \mathbb{R}^{M}$, respectively.  The loss function for pretraining the masked autoencoders is composed of two terms:
\begin{align}
     L_n & = \| \mathcal{N}_M - \mathcal{\widehat{N}}_M \|_2^2; \\
       L_v & = \| \mathcal{V}_M - \mathcal{\widehat{V}}_M\|_1; 
\end{align}
where $\mathcal{\widehat{N}}_M$ and $\mathcal{\widehat{V}}_M$ are the predicted versions of $\mathcal{N}_M$ and $\mathcal{V}_M$, respectively. The total loss function $L = \lambda_1 L_n + \lambda_2 L_v$, where $\lambda_1 = 1, \lambda_2 = 1$ in our case.


\subsubsection{Encoder Design} \label{subsec:encoder}

Unlike most MAE-based approaches that are limited to ViT-based encoders, our approach is not restricted to any specific type of encoder. Common 3D encoders for point clouds are all supported, as long as the encoder outputs a set of learned features bind to spatial blocks, where spatial blocks could be point patches used for ViT-like transformer encoders~\cite{Yu_2022_CVPR,Pang2022MaskedAF,Liu2022maskdis,zhang2022masked}, set abstractions used by PointNet++-like encoders~\cite{qi2017pointnet++,PointNeXt}, or coarse voxels used by sparse CNN-based encoders~\cite{Wang:O:CNN,Graham2018,ChoyGS19}.

In the following, we briefly review these typical encoders and their adaption for our pretraining.

\paragraph{ViT-based encoders}  These encoders first embed point patches via PointNet~\cite{qi2017pointnet}, then send these patch tokens to a standard transformer that includes several multihead self-attention layers and feedforward layers. The transformer outputs the fabricated token features, corresponding to every input point patch. The token feature $\mathbf{f}_i$ and the patch center $\mathbf{c}_i$ form a block feature pair $B_i = \{\mathbf{f}_i, \mathbf{c}_i\}$, which is needed by our decoder.  Here we can call $\mathbf{f}_i$ \emph{block feature} and $\mathbf{c}_i$ \emph{block centroid}. 

\paragraph{PointNet++-like encoders} In these encoders, the network features are aggregated through a number of set abstraction levels. We take the learned features and the centroids at the coarsest set abstractions as block feature pairs. 

\paragraph{Sparse CNN-based encoders} These encoders apply 3D convolution on sparse voxels from the finest level to the coarsest level. Multiple convolution layers and resblocks are commonly used.
We interpolate the coarse voxel features at the centroids of the unmasked patches and use these interpolated features and the patch centroids to form our block feature pairs.

As suggested by~\cite{Pang2022MaskedAF}, the early leaking of masked point information to the network could jeopardize feature learning. We adopt this suggestion: feed the unmasked points to the encoder only, and leave the masked points to the decoder.

\subsubsection{Decoder Design} \label{subsec:decoder}

\paragraph{Decoder structure}
We design an attention-based decoder to restore the target features at masked regions. 
The decoder takes the block feature pairs $\mathcal{B}:=\{B_i\}_{i=1}^b$ from the encoder and a query point set $\mathcal{Q}$, \ie the masked point set $\mathcal{P}_M$.  It is composed of a stack of $l$ transformer blocks, where $l = 4$ in our case (See Fig.~\ref{fig:maskarch}). Each block contains a self-attention layer and a cross-attention layer. The self-attention layer takes the query points and their positional embeddings as input and outputs the per-query point features, denoted by $\mathcal{S}^{in}$.  Then $\mathcal{S}^{in}$ and the encoder block features $\mathcal{B}$ are passed into the cross-attention layer, where $\mathcal{S}^{in}$ serves as attention \textbf{query}, the block features serve as attention \textbf{key} and \textbf{value}, and the block centroids are the positional embedding of the block features. The output per-point features from the last block go through an MLP head to predict the target features at the query points.  

\paragraph{Efficacy of self-attention layers} At first glance, it is sufficient to use cross-attention layers only for predicting per-point features. The recent masked discrimination work~\cite{Liu2022maskdis} obeys this intuition for its decoder design, no information exchanged between different query points. Instead, we introduce the self-attention layer to propagate information over query points and use multiple attention blocks to strengthen the mutual relationship progressively. We found that our design significantly improves feature learning, as verified by our ablation study (See Section Ablation Study). 

\paragraph{Supporting of various encoders}
In the above design, the decoder needs block feature pairs only from the encoder, thus having great potential to leverage various encoder structures, not limited to ViT-based transformer structures. This advantage is verified by our experiments (See \ref{sec:results}).

\paragraph{Feature reconstruction versus position reconstruction} Note that our decoder and loss design do not explicitly model point positions, which are zero-order surface properties complementary to surface normals and surface variations. Instead, the decoder predicts feature values at query points. Therefore, the zero-order positional information is already encoded implicitly. This explains why our approach is superior to baseline approaches that reconstruct point positions for feature learning (See \ref{subsec:featurecomp}).

\paragraph{Query point selection} Due to the quadratic complexity of self-attention, the computational cost for a full query set could be much higher. In practice, we can randomly choose a point subset from $\mathcal{P}_M$ as the query set during training. By default, we use all masked points as queries.

%% file: chapters/MaskFeat3D/Figures/fig-arch.tex
\begin{figure*}[h]
    \centering
    \begin{overpic}[width=1\linewidth]{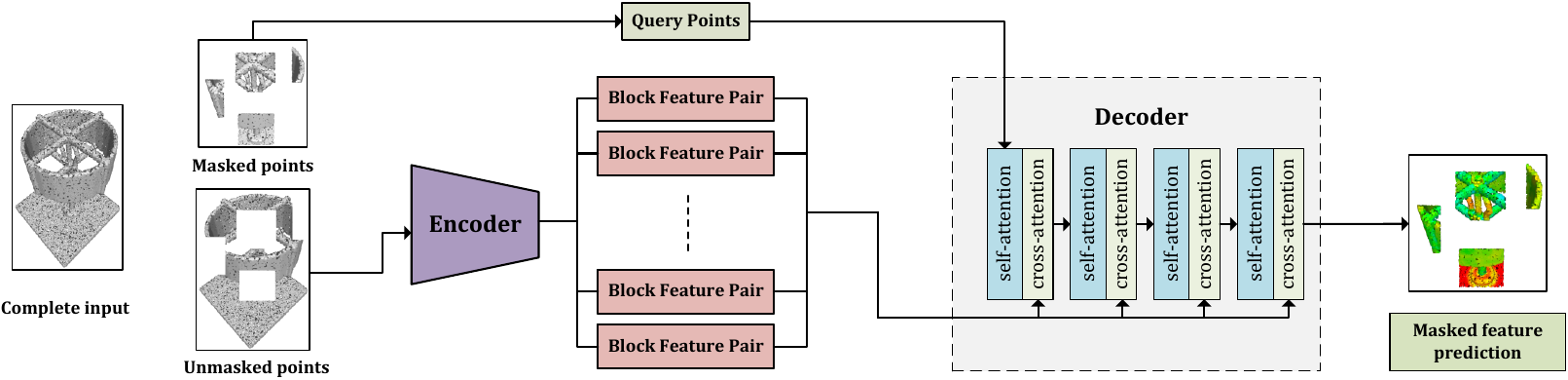}
    \end{overpic}
    \caption{The pretraining pipeline of our masked 3D feature prediction approach. Given a complete input point cloud, we first separate it into masked points and unmasked points (We use cube mask here for better visualization). We take unmasked points as the encoder input and output the block feature pairs. Then the decoder takes the block feature pairs and query points(i.e., masked points) as the input, and predicts the per-query-point features.}
    \label{fig:maskarch} 
\end{figure*}

%% file: chapters/MaskFeat3D/Figures/fig-vis-feature.tex
\begin{figure}[h]
    \centering
    \begin{overpic}[width=\linewidth]{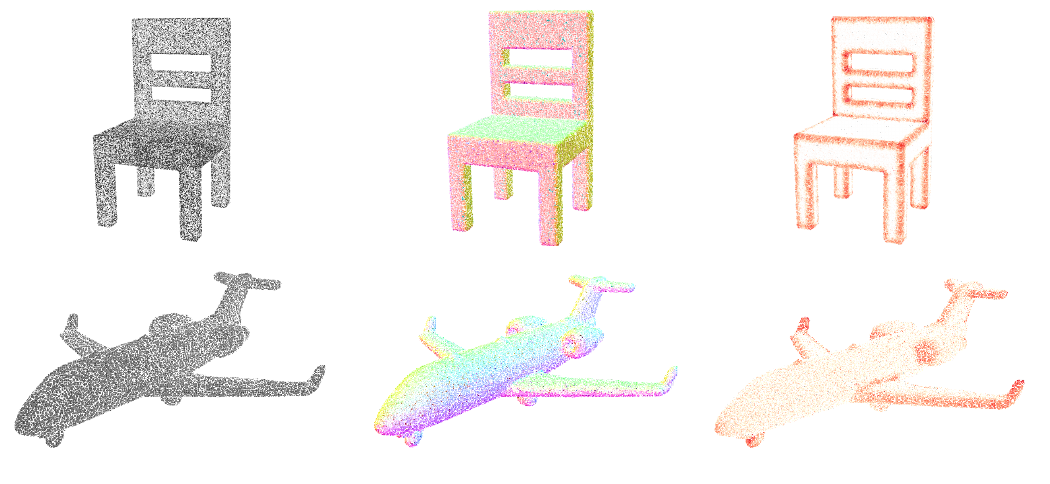}
    \put(4,-3.5){point cloud}
    \put(35,-3.5){point normal}
    \put(65,-3.5){surface variation}
    \end{overpic}
    \vspace{2pt}
    \caption{Visualization of point features. The point normal is color-coded by the normal vector. The surface variation is color-coded where white indicates low value and red indicates high value.}
    \label{fig:variation} 
\end{figure}

%% file: chapters/MaskFeat3D/Sections/04-Evaluation.tex
\vspace{-5pt}
\subsection{Experiment Analysis} \label{sec:results}
We conducted a series of experiments and ablation studies to validate the efficacy and superiority of our masked 3D feature prediction approach, in short \textbf{MaskFeat3D},  for point cloud pretraining. 
\vspace{-5pt}
\subsubsection{Experiment Setup}

\paragraph{Pretraining dataset} We choose ShapeNet~\cite{chang2015shapenet} dataset for our pretraining, following the practice of PointBERT~\cite{Yu_2022_CVPR} and previous 3D MAE-based approaches~\cite{Pang2022MaskedAF,zhang2022masked,Liu2022maskdis}. 
ShapeNet~\cite{chang2015shapenet} contains \num{57748} synthetic 3D shapes from 55 categories.  
We sample \num{50000} points uniformly on each shape and select $128$ nearest points from them for each point in the point cloud for constructing the local region to approximate surface variation. During pretraining, $N = 2048$ points are randomly sampled to create the point cloud. 


\paragraph{Network training}
We integrated different encoders with our masked 3D feature prediction approach, including the ViT-based transformer used by \cite{Pang2022MaskedAF}, sparse-CNN-based encoder~\cite{ChoyGS19}, 
and  PointNeXt encoder~\cite{PointNeXt} which is an advanced version of PointNet++. We implemented all pretraining models in PyTorch and used AdamW optimizer with $10^{-4}$ weight decay. We use PointBERT’s masking strategy for ShapeNet pretraining. We set $K = 128$ in FPS, $k=32$-nearest points to form the point patch, and the best masking ratio is $60\%$ empirically. The number of transformer blocks in the decoder is 4. The learning rates of the encoder and the decoder are set to $10^{-3}$ and $10^{-4}$, respectively. Standard data augmentation such as rotation, scaling, and translation are employed.
All models were trained with 300 epochs on eight \SI{16}{GB} Nvidia V100 GPUs. The total batch size is $64$.

\paragraph{Downstream Tasks} We choose shape classification and shape part segmentation tasks to validate the efficacy and generalizability of our pretrained networks.  
\begin{enumerate}[leftmargin=*]\setlength\itemsep{0mm}
    \item[-] \textbf{Shape classification}: The experiments were carried out on two different datasets: ModelNet40~\cite{WuSKYZTX15} and ScanObjectNN~\cite{UyPHNY19}.
ModelNet40 is a widely used synthetic dataset that comprises 40 classes and contains \num{9832} training objects and \num{2468} test objects. In contrast, ScanObjectNN is a real-world scanned dataset that includes approximately \num{15000} actual scanned objects from 15 classes. 
As the domain gap between ShapeNet and ScanObjectNN is larger than that between ShapeNet and ModelNet40, the evaluation on ScanObjectNN is a good measure of the generalizability of pretrained networks. 

\item[-] \textbf{Shape part segmentation}  ShapeNetPart Dataset~\cite{10.1145/2980179.2980238} contains \num{16880} models from 16 shape categories, and each model has 2$\sim$6 parts. Following the standard evaluation protocol~\cite{qi2017pointnet++}, \num{2048} points are sampled on each shape. For evaluation, we report per-class mean IoU (cls. mIoU) and mean IoU averaged over all test instances (ins. mIoU).



\end{enumerate}

The training-and-test split of the above tasks follows existing works. For these downstream tasks, we employ the task-specific decoders proposed by PointMAE~\cite{Pang2022MaskedAF} and reload the pretrained weights for the encoder. Training details are provided in the supplemental material.

\subsubsection{Efficacy of 3D Feature Prediction}
\label{subsec:featurecomp}


\input{chapters/MaskFeat3D/Tables/comp-mae-3}

\input{chapters/MaskFeat3D/Tables/comp-all-2}


The advantages in learning discriminative features by our masked feature prediction approach are verified by its superior performance in downstream tasks.

\paragraph{Comparison with MAE-based approaches} 
We compare our approach with other MAE-based approaches that use the same encoder structure. \ref{tab:comp-mae} reports that: (1) the performance of all MAE-based methods surpasses their supervised baseline -- PointViT; (2) our strategy of reconstructing point features instead of point positions yields significant improvements in ScannObjectNN classification,  improving overall accuracy on the most challenging split, PB-T50-RS, from $85.7\%$ (MaskSurfel) to
$87.7\%$, and showing consistent improvements on other splits and ShapeNetPart segmentation.

We also compare the performance of our approach with PointBERT~\cite{Yu_2022_CVPR} ,PointMAE~\cite{Pang2022MaskedAF}, and MaskDiscr~\cite{Liu2022maskdis} on ShapeNetPart segmentation with less labeled data. In this experiment, we randomly select $1\%$ labeled data from each category, and finetune the network with all selected data. The performance is reported in \ref{tab:comp-mae}, which shows that using our pretrained network leads to much better performance than the baseline methods. 



\paragraph{Comparison with supervised approaches} 
Compared with state-of-the-art supervised methods, our approach again achieves superior performance than most existing works as seen from \ref{tab:comp-all}, including PointNet++~\cite{qi2017pointnet++} ,PointCNN~\cite{pointcnn},  DGCNN~\cite{dgcnn:2019:tog}, MinkowskiNet~\cite{ChoyGS19}, PointTransformer~\cite{pointr} and PointMLP~\cite{PointMLP}. It is only inferior to the approaches that use advanced encoder structures such as stratified transformer~\cite{stratified} and PointNeXt~\cite{PointNeXt}. 

\paragraph{Encoder replacement}
To make a more fair comparison, we replaced the PointViT encoder with the PointNeXt's encoder, and retrained our pretraining network, denoted as MaskFeat3D (PointNeXt). From \ref{tab:comp-all}, we can see that our pretraining approach with this enhanced encoder can yield SOTA performance on all the downstream tasks, surpassing PointNeXt trained from scratch. We also used MinkowskiNet~\cite{ChoyGS19} as our pretraining encoder, the performance gain over MinkowskiNet trained from scratch is +0.7\% overall accuracy improvement on ScanObjectNN classification, and +0.3\% on ShapeNetPart segmentation. Please refer to the supplementary material for details.

\input{chapters/MaskFeat3D/Tables/fewshot_cls}

\paragraph{Few-shot Classification} To perform few-shot classification on ModelNet40, we adopt the ``K-way N-shot" settings as described in prior work~\cite{wang2021unsupervised,Yu_2022_CVPR, Pang2022MaskedAF}. Specifically, we randomly choose K out of the 40 available classes and sample N+20 3D shapes per class, with N shapes used for training and 20 for testing. We evaluate the performance of MaskFeat3D under four few-shot settings: 5-way 10-shot, 5-way 20-shot, 10-way 10-shot, and 10-way 20-shot. To mitigate the effects of random sampling, we conduct 10 independent runs for each few-shot setting and report the mean accuracy and standard deviation. Additionally, more ModelNet40 results can be found in the supplementary material.

\input{chapters/MaskFeat3D/Tables/supp_modelnet}
\input{chapters/MaskFeat3D/Tables/supp_modelnet_svm}
\paragraph{Shape classification on ModelNet40}
\label{sec:supp-mn40}
In this section, we evaluate our method by supervised fine-tuning on ModelNet40 dataset. As shown in Table~\ref{tab:modelnet40}, under the same Transformer backbone, PointViT, our method shows the best performance.

\paragraph{Linear SVM on ModelNet40}
We also evaluate the performance of different pretrained networks by fixing the pretrained weights and using linear SVMs on ModelNet40 classification (see Table~\ref{tab:svm}). Compared with other MAE-based approaches that use the same ViT-based encoder structure, our approach achieves the best classification accuracy.

Overall, the improvements of our approach are consistent across different backbone encoders and datasets. 

\input{chapters/MaskFeat3D/Figures/fig-feavis}
\subsubsection{Feature Visualization}

We also assess the discriminativity of learned features by visualizing point features as follows. For a given point on a point cloud, we linearly interpolate block features at this point, using the inverse of its distance to the block centroids as the interpolation weights. We then project all the interpolated point features into the 3D space, via T-SNE embedding~\citep{tsne}. The 3D space is treated as the RGB color space for assigning colors to points. 
Here, the block features are from the pretrained encoder, not finetuned on downstream tasks. The color difference between different points characterizes their feature differences, and distinguishable features are preferred by many downstream tasks. 
The first column of Fig.~\ref{fig:feavis} shows the point features of two chair shapes learned by our approach using the ViT-based encoder. 

We also visualize the point features learned by PointBERT~\citep{Yu_2022_CVPR} and other MAE-based pretraining approaches: PointMAE~\citep{Pang2022MaskedAF}, MaskDiscr~\citep{Liu2022maskdis}, MaskSurfel~\citep{zhang2022masked}, where all these methods use a similar ViT-based encoder. 
Fig.~\ref{fig:feavis} shows that our learned features are more discriminative than other methods. For instance,  the legs of the chair in the first row from our method --- MaskFeat3D, are more distinguishable than the results of PointBert and MaskSurfel, because of their clear color difference; the two arms and different legs in the second row from MaskFeat3D are also more discriminative than those from other methods. 

\input{chapters/MaskFeat3D/Figures/supp-fig-feat-vis-v2}

\subsubsection{Zero-Shot Correspondence Learning}

We also study how our model's learned feature aids zero-shot correspondence learning. As shown in Fig.~\ref{fig:feat-vis-v2}, we randomly selected a query point (red) on the object to the left and showed the nearest 300 points (blue) in the feature space on that object and another one from the same category. We observed that the nearest points shared similar semantic information across both objects, thereby demonstrating the efficacy of our model's feature learning.

\input{chapters/MaskFeat3D/Tables/ablation_feat}

\subsubsection{Ablation Study}
\label{sec:ablation:study}
We proceed to present an ablation study to justify various design choices. For simplicity, we choose the shape classification task on ScanObjectNN, where the gaps under different configurations are salient and provide meaningful insights on the pros and cons of various design choices. Due to space constraints, additional ablation studies are available in the supplementary material.

\paragraph{Decoder design} The primary question that arises is whether it is essential to disregard point position recovery. PointMAE's decoder follows a standard ViT-like architecture, utilizing a fully connected (FC) layer to directly predict the masked point coordinates. We implemented this decoder to predict our target features. However, since their decoder design does not encode masked point position, it cannot solely predict target features without predicting point position. To address this, we follow the approach proposed in ~\cite{zhang2022masked} and employ position-index matching for feature loss computation. As shown in ~\ref{tab:feature_ablation}, even though incorporating point features as the predicting target can enhance performance, the overall performance still significantly lags behind our design. This experiment highlights the significance of both point feature prediction and disregarding point position recovery.

\vspace{-5pt}
\paragraph{Target feature choice} In \ref{tab:feature_ablation}, the experiment shows that: (1) All combinations of point normal and surface variation can yield significant improvements over existing MAE approaches that recover point positions (\cf \ref{tab:comp-mae}); (2) using both point normals  and surface variations yields the best performance. As discussed in \ref{subsec:feature}, this is due to the fact that they correspond to first- and second-order differential properties. They are relevant but complementary to each other. Therefore, reconstructing them together forces the encoder to learn more informative features than merely reconstructing one of them. 

\vspace{-5pt}
\paragraph{Decoder depth} \ref{tab:ablation}-a varies the number of transformer blocks (decoder depth). A sufficient deep decoder is necessary for feature learning. Increasing the number of blocks from 2 to 4 provides +1.5\% improvement on ScanObjectNN classification task. The performance drops when increasing the depth further, due to the overfitting issue. Interestingly, we note that a 1-block decoder can strongly achieve 85.8\% accuracy, which is still higher than the runner-up method (PointMAE).

\vspace{-5pt}
\paragraph{Data augmentation} \ref{tab:ablation}-b studies three traditional data augmentation methods: rotation, scaling, and translation. Since the standard scaling could change the surface normal and variation, we scale the shape by using the same factor on 3 different axis. The experiments show that rotation and scaling play a more important role.
\input{chapters/MaskFeat3D/Tables/ablation_study2}

\vspace{-5pt}
\paragraph{Masking ratio.} \ref{tab:ablation}-c varies the masking ratio of input point cloud, which is another important factor on our approach. When the masking ratio is too large, \eg 90\%, the remaining part contains too limited information, which makes the task too hard to complete. When masking ratio is too small, \eg 40\%, the task becomes too simple and impedes the feature learning. In our experiments, masking ratio=60\% shows the best performance.


\paragraph{Decoder block design} We tested whether the self-attention layer in our decoder is essential. By simply removing self-attention layers and using cross-attention layers only, we find that the performance has a large drop (-2.0), see \ref{tab:ablation}-d.

\vspace{-5pt}
\paragraph{Number of query points} Finally, we varied the number of query points used by our decoder to see how it affects the network performance. \ref{tab:ablation}-e shows that more query points lead to better performance. Here, ``query/mask'' is the ratio of selected query points with respect to the total number of masked points. 

\input{chapters/MaskFeat3D/Tables/scene_task}

\subsubsection{Scene-level Pretraining Extension} 

In principle, masked point cloud autoencoders could be scaled to noisy, large-scale point clouds. Additionally, we conducted an extension experiment on real-world scene-level data to evaluate our approach. Specifically, we pretrained our model on the ScanNet~\cite{DBLP:conf/cvpr/DaiCSHFN17} dataset and evaluated its performance on 3D object detection task using the ScanNet and SUN RGB-D~\cite{song2015sun} dataset. The training details can be found in the supplementary material.
In this experiment, we observed that surface normal has a minor influence on the pretraining, while surface variation remains a robust feature. Moreover, we discovered that color signal could be an effective target feature. Hence, we pretrained our model with surface variation and color as the target features, and then fine-tuned the pretrained encoder on the downstream tasks. As shown in Tab.~\ref{tab:maskfeat_detection}, given that previous studies lack a unified network backbone, we selected two of the most common works, VoteNet and Point-M2AE, along with the latest work, CAGroup3D, as the network backbones respectively. And our model exhibits consistent improvements in all the settings, which further proves the generalizability of our approach on noisy, large-scale point clouds. Although the concrete scene-level experiments are not the main focus of this paper, the results indicate that this is a promising direction.

%% file: chapters/MaskFeat3D/Tables/comp-mae-3.tex
\begin{table*}[t]
    \centering
    \resizebox{1.\linewidth}{!}{  
    \begin{tabular}{lccccccc}
        \toprule
        \multirow{2}{*}{\textbf{Method}} & \multicolumn{3}{c}{\textbf{ScanObjectNN}} & \multicolumn{2}{c}{\textbf{ShapeNetPart}} & \multicolumn{2}{c}{\textbf{ShapeNetPart}(1\% labels)} \\ 
      \cmidrule(lr){2-4}   \cmidrule(lr){5-6}    \cmidrule(lr){7-8}         &  OBJ-BG    & OBJ-ONLY & PB-T50-RS              & ins. mIoU                                 & cls. mIoU                                             & ins. mIoU     & cls. mIoU     \\
        \midrule
        PointViT$^{\dagger}$ ~\cite{Yu_2022_CVPR}     & 79.9 & 80.6 & 77.2              & 85.1                                      & 83.4                                                  & 77.6          & 72.2          \\
        PointBERT~\cite{Yu_2022_CVPR}    & 87.4 & 88.1 & 83.1                 & 85.6                                      & 84.1                                                  & 79.2          & 73.9          \\
        MaskDiscr~\cite{Liu2022maskdis}  & 89.7 & 89.3 & 84.3                   & 86.0                                      & 84.4                                                  & 78.8           & 72.3         \\
        MaskSurfel~\cite{zhang2022masked} & 91.2 & 89.2 & 85.7                  & 86.1                                      & 84.4                                                  & -             & -             \\
        PointMAE~\cite{Pang2022MaskedAF} & 90.0 & 88.3 & 85.2                   & 86.1                                      & -                                                     & 79.1          & 74.4          \\
        MaskFeat3D (PointViT)    & \textbf{91.7}(91.6) & \textbf{90.0}(89.6)  & \textbf{87.7}(87.5)   & \textbf{86.3}(86.3)                & \textbf{84.9}(84.8)                         & \textbf{80.0}(79.9) & \textbf{75.1}(75.0) \\
        \bottomrule
    \end{tabular}
    }
    \caption{Performance comparison of MAE-based approaches on downstream tasks. All the methods in the first section use the same transformer backbone architecture, PointViT. $^{\dagger}$ represents the \textit{from scratch} results and all other methods represent the \textit{fine-tuning} results using pretrained weights. The average result of 3 runs is given in brackets.}
    \label{tab:comp-mae}
\end{table*}

%% file: chapters/MaskFeat3D/Tables/comp-all-2.tex
\begin{table*}[t]
    \centering
    \setlength\tabcolsep{8pt}
    \resizebox{1.\linewidth}{!}{
    \begin{tabular}{lccccc}
        \toprule
                     \multirow{2}{*}{\textbf{Method}}                         & \multicolumn{3}{c}{\textbf{ScanObjectNN}} & \multicolumn{2}{c}{\textbf{ShapeNetPart}}                             \\
       \cmidrule(lr){2-4}   \cmidrule(lr){5-6}                    &  OBJ-BG    & OBJ-ONLY & PB-T50-RS  & ins. mIoU                                 & cls. mIoU                 \\
        \midrule
        PointNet~\cite{qi2017pointnet}                  & 73.3 &  79.2       & 68.0                        & -                                      & -                    \\        
        PointNet++~\cite{qi2017pointnet++}                  & 82.3 &  84.3       & 77.9                        & 85.1                                      & 81.9                      \\
        PointCNN~\cite{pointcnn}                & 86.1 &  85.5           & 78.5       &    86.1	                                       &    84.6                       \\
        DGCNN~\cite{dgcnn:2019:tog}             & 82.8 &   86.2                & 78.1             & 85.2                                      & 82.3                      \\
        MinkowskiNet~\cite{ChoyGS19}& 84.1 & 86.1 & 80.1    & 85.3               & 83.2 \\  
        PointTransformer~\cite{pointr} & - & - & -                              & 86.6                                      & 83.7                      \\
        PointMLP~\cite{PointMLP}             & 88.7 &  88.2              & 85.4              & 86.1                                      & 84.6                      \\
        StratifiedTransformer~\cite{stratified}          & - &        -         & -       & 86.6                                      & 85.1                      \\
        PointNeXt~\cite{PointNeXt}            & 91.9 & 91.0               & 88.1   & 87.1                         & 84.7        \\

        MaskFeat3D (PointViT)     & 91.7(91.6) &  90.0(89.6)  & 87.7(87.5)  & 86.3(86.3)                & 84.9(84.8) \\
        MaskFeat3D (MinkowskiNet) & 85.1(85.0) & 87.0(86.7) & 80.8(80.6)   & 85.6(85.5)              & 83.5(83.5) \\  
        MaskFeat3D (PointNeXt) & \textbf{92.7}(92.6) & \textbf{92.0}(91.9) & \textbf{88.6}(88.5) & \textbf{87.4}(87.4)               & \textbf{85.5}(85.5) \\
        \bottomrule
    \end{tabular}
    }
    \caption{Comparison with supervised methods. The average result of 3 runs is given in brackets.}
    \label{tab:comp-all}
\end{table*}

%% file: chapters/MaskFeat3D/Tables/fewshot_cls.tex
\begin{table*}[h]

\centering
\setlength\tabcolsep{4pt}

\begin{tabular}{lcccc}
\toprule
\multirow{2}{*}{\textbf{Method}} & \multicolumn{2}{c}{\textbf{5-way}} & \multicolumn{2}{c}{\textbf{10-way}} \\
\cmidrule(lr){2-3}   \cmidrule(lr){4-5}
 & 10-shot & 20-shot & 10-shot & 20-shot \\
\hline
DGCNN$^{\dagger}$ & $31.6 \pm 2.8$ & $40.8 \pm 4.6$ & $19.9 \pm 2.1$ & $16.9 \pm 1.5$ \\
OcCo & $90.6 \pm 2.8$ & $92.5 \pm 1.9$ & $82.9 \pm 1.3$ & $86.5 \pm 2.2$ \\
CrossPoint & $92.5 \pm 3.0$ & $94.9 \pm 2.1$ & $83.6 \pm 5.3$ & $87.9 \pm 4.2$ \\
\hline
Transformer$^{\dagger}$ & $87.8 \pm 5.2$ & $93.3 \pm 4.3$ & $84.6 \pm 5.5$ & $89.4 \pm 6.3$ \\
OcCo & $94.0 \pm 3.6$ & $95.9 \pm 2.3$ & $89.4 \pm 5.1$ & $92.4 \pm 4.6$ \\
PointBERT & $94.6 \pm 3.1$ & $96.3 \pm 2.7$ & $91.0 \pm 5.4$ & $92.7 \pm 5.1$ \\
MaskDiscr & $95.0 \pm 3.7$ & $97.2 \pm 1.7$ & $91.4 \pm 4.0$ & $93.4 \pm 3.5$ \\
PointMAE & $96.3 \pm 2.5$ & $97.8 \pm 1.8$ & $92.6 \pm 4.1$ & $95.0 \pm 3.0$ \\
MaskFeat3D & $\pmb{97.1} \pm \pmb{2.1}$ & $\pmb{98.4} \pm \pmb{1.6}$ &  $\pmb{93.4} \pm \pmb{3.8}$ & $\pmb{95.7} \pm \pmb{3.4}$ \\
\bottomrule
\end{tabular}
\caption{Few-shot classification on ModelNet40. We report the average accuracy (\%) and standard deviation (\%) of 10 independent experiments.}
\label{tab:mn_fewshot}

\end{table*}

%% file: chapters/MaskFeat3D/Tables/supp_modelnet.tex
\begin{table*}[t]
    \centering
    \begin{tabular}{lc}
        \toprule
        \textbf{Method} & \textbf{ModelNet40} \\
        \midrule
        PointViT$^{\dagger}$ ~\cite{Yu_2022_CVPR}     & 91.4 \\
        PointBERT~\cite{Yu_2022_CVPR}    & 93.2 \\
        MaskDiscr~\cite{Liu2022maskdis}  & 93.8 \\
        MaskSurfel~\cite{zhang2022masked} & 93.6 \\
        PointMAE~\cite{Pang2022MaskedAF} & 93.8 \\
        MaskFeat3D (PointViT)    & \textbf{93.9} \\
        \bottomrule
    \end{tabular}
    \captionof{table}{Shape classification fine-tuned on ModelNet40. All the methods use the same transformer backbone architecture. $^{\dagger}$ represents the \textit{from scratch} results and all other methods represent the \textit{fine-tuning} results using pretrained weights.}
    \label{tab:modelnet40}
\end{table*}

%% file: chapters/MaskFeat3D/Tables/supp_modelnet_svm.tex
\begin{table*}[h]
 \centering
    \begin{tabular}{lc}
        \toprule
        \textbf{Method} & \textbf{Acc. (\%)} \\
        \midrule
        3D-GAN~\cite{wu2016learning}  & 83.3 \\
        Latent-GAN~\cite{achlioptas2018learning} & 85.7 \\
        SO-Net~\cite{li2018so} & 87.3 \\
        MAP-VAE~\cite{han2019multi} & 88.4 \\
        Jigsaw~\cite{sauder2019self} & 84.1 \\
        FoldingNet~\cite{yang2018foldingnet} & 88.4 \\
        DGCNN + OcCo~\cite{wang2021unsupervised} & 89.7 \\
        DGCNN + STRL~\cite{huang2021spatio} & 90.9 \\
        \midrule 
        PointViT + OcCo$^*$~\cite{wang2021unsupervised} & 89.6 \\
        PointBERT~\cite{Yu_2022_CVPR}$^*$ & 87.4 \\
        PointMAE~\cite{Pang2022MaskedAF}$^*$ & 88.5 \\
        MaskFeat3D (PointViT) $^*$ & \textbf{91.1}\\
        
        \bottomrule
    \end{tabular}
    \captionof{table}{Linear evaluation for shape classification on ModelNet40. This task is sensitive to the encoder backbone. Different * methods use the same Transformer encoder backbone.}
    \label{tab:svm}
\end{table*}

%% file: chapters/MaskFeat3D/Figures/fig-feavis.tex
\begin{figure*}[h]
    \centering
    \begin{overpic}[width=0.8\linewidth]{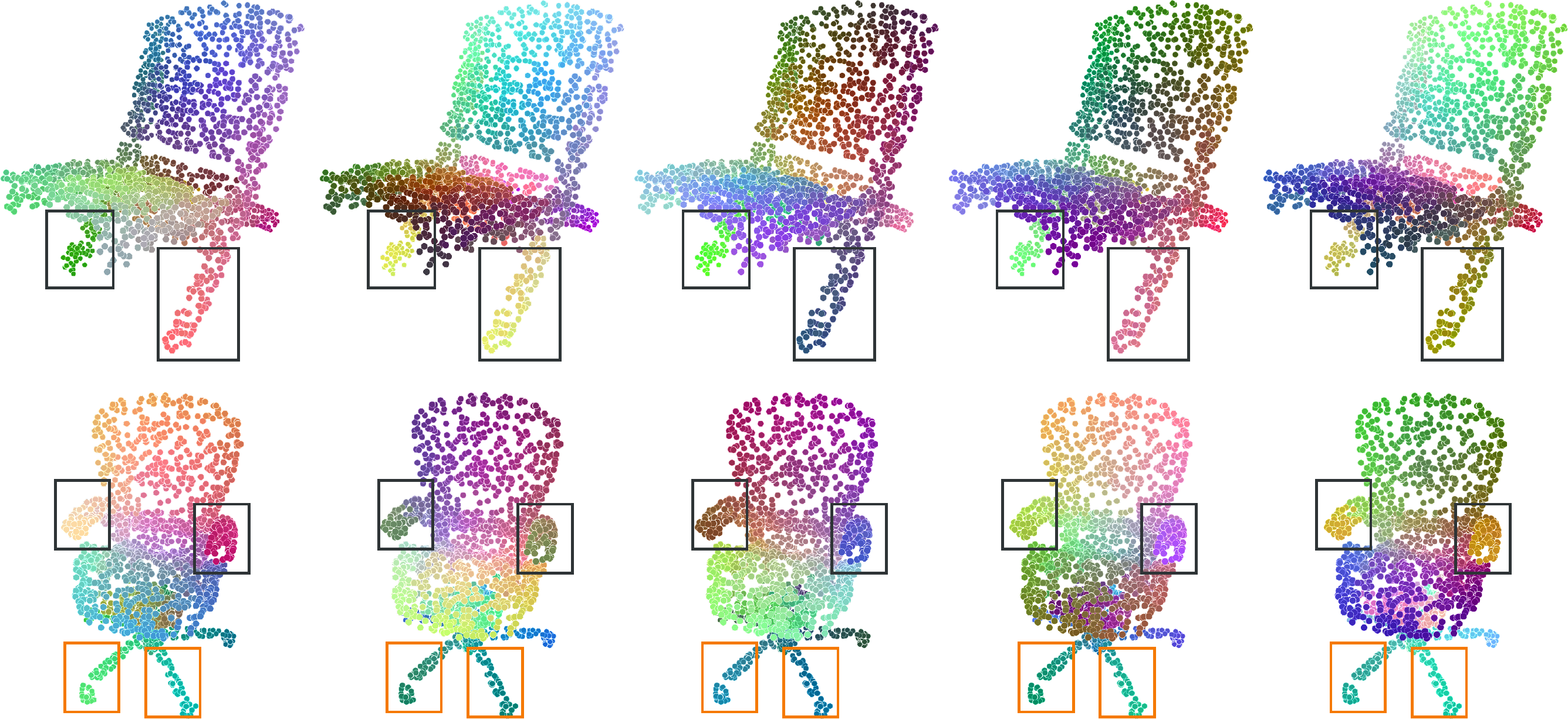}
    \put(4,-3.5){\scriptsize  \textbf{MaskFeat3D}}
    \put(25,-3.5){\scriptsize \textbf{PointBERT}}
    \put(45,-3.5){\scriptsize \textbf{PointMAE}}
    \put(65,-3.5){\scriptsize \textbf{MaskDiscr}}
    \put(85,-3.5){\scriptsize \textbf{MaskSurfel}}
    \end{overpic}
    \vspace{2mm}
    \caption{Feature visualization of different pretrained approaches including our MaskFeat3D,  PointBERT~\cite{Yu_2022_CVPR}, PointMAE~\cite{Pang2022MaskedAF}, MaskDiscr~\cite{Liu2022maskdis}, MaskSurfel~\cite{zhang2022masked}. Here, note that the absolute colors from different methods are not comparable as their feature spaces are not the same, instead the point color difference from the same method is a good visual measurement for assessing feature discriminativity.
        First row: By comparing the colors of the two front legs, we can see our pretrained network produces more discriminative features than PointBERT and MaskSurfel. Second row: It is clear the features on the different chair legs from our methods, as well as the features of two chair arms, are more discriminative.   }
    \label{fig:feavis} 
\end{figure*}

%% file: chapters/MaskFeat3D/Figures/supp-fig-feat-vis-v2.tex
\begin{figure}[h]
    \centering
    \includegraphics[width=0.7\linewidth]{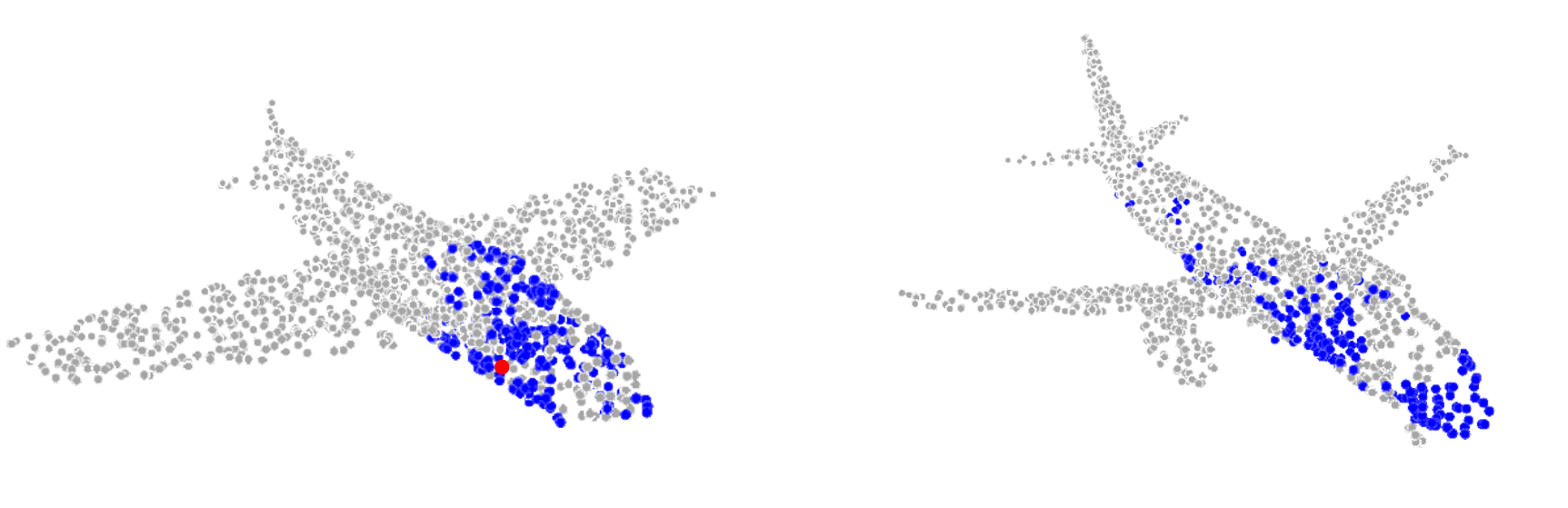}
    \caption{Illustration of zero-shot correspondence learning.}
    \label{fig:feat-vis-v2}
\end{figure}

%% file: chapters/MaskFeat3D/Tables/ablation_feat.tex
\begin{table*}[h]

\centering
\begin{tabular}{llc}
        \toprule
      \textbf{Method}  & \textbf{Target Feature} & \textbf{ScanNN}\\
      \midrule
      \multirow{4}{*}{PointMAE}  & position only & 85.2 \\
        & position + normal$^*$ & 85.7 \\
        & position + surface variation$^*$ & 85.9 \\
        & position + normal + variation$^*$ & 86.0 \\
        \midrule
       \multirow{3}{*}{MaskFeat3D} & normal & 86.5 \\
        & surface variation & 87.0\\
         & normal + variation & \textbf{87.7}\\
        \bottomrule
\end{tabular}

\captionof{table}{Ablation study on different features. $^*$ uses position-index matching~\cite{zhang2022masked} for feature loss computation.}
\label{tab:feature_ablation}

\end{table*}

%% file: chapters/MaskFeat3D/Tables/ablation_study2.tex
\begin{table*}[h]
\setlength\tabcolsep{5pt}

\begin{minipage}[b]{.5\linewidth}
\centering  
    \begin{tabular}{cc}
        \toprule
         \# blocks & ScanNN\\
        \midrule
        1 & 85.8 \\
        2 & 86.2\\
         4 & \textbf{87.7}\\
        8 & 87.5\\
        12 & 87.1\\
        \bottomrule
    \end{tabular}
    \caption{Decoder depth}
     \label{tab:ab_depth}
\end{minipage}
\begin{minipage}[b]{.5\linewidth}
\centering  
\begin{tabular}{cccc}
\toprule
        rot & scale & trans & ScanNN \\ 
        \midrule
        $\surd$ & - & - & 87.0\\
        - & $\surd$ & - & 85.9\\
        $\surd$ & $\surd$ & - & \textbf{87.7}\\
        - & $\surd$ & $\surd$ & 85.1\\
        $\surd$ & $\surd$ & $\surd$ & 86.7 \\
        \bottomrule
    \end{tabular}
    \caption{Data augmentation}
    \label{tab:ab_aug}
\end{minipage}

\begin{minipage}[b]{.5\linewidth}
\centering  
    \begin{tabular}{cc}
    \toprule
     ratio & ScanNN \\
        \midrule
        40\% & 86.8 \\
         60\% & \textbf{87.7} \\
        90\% & 86.5 \\
        \bottomrule
    \end{tabular}
    \caption{Mask ratio}
    \label{tab:ab_ratio}
\end{minipage}
\begin{minipage}[b]{.5\linewidth}
\centering  
\begin{tabular}{cc}
\toprule
     attention type & ScanNN \\
        \midrule
        cross only & 85.7\\
        cross+self & \textbf{87.7} \\
       \bottomrule
    \end{tabular}
    \caption{Decoder attention} 
    \label{tab:ab_att1}
\end{minipage}
 \hspace*{0.02\textwidth}
\begin{minipage}[b]{.5\linewidth}
\centering   
\begin{tabular}{cc}
\toprule
     query/mask & ScanNN \\
        \midrule
        25\% & 85.7 \\
        50\%&  86.2 \\
        75\% & 86.6 \\
        100\% & \textbf{87.7} \\
    \bottomrule
    \end{tabular}
    \caption{Query point ratio}
    \label{tab:ab_att}
\end{minipage}

\caption{Ablation studies of our design choices. Please refer to \ref{sec:ablation:study} for a detailed analysis.}
\label{tab:ablation}
\end{table*}

%% file: chapters/MaskFeat3D/Tables/scene_task.tex
\begin{table}[h]

\centering
   \setlength\tabcolsep{5pt}
    \begin{tabular}{llcccc}
        \toprule
     \multirow{2}{*}{Method} & \multirow{2}{*}{Backbone} & \multicolumn{2}{c}{ScanNet} & \multicolumn{2}{c}{SUN RGB-D} \\
        \cmidrule(lr){3-4}   \cmidrule(lr){5-6} & & $\text{AP}_{25}$ & $\text{AP}_{50}$ & $\text{AP}_{25}$ & $\text{AP}_{50}$ \\
        \midrule
        STRL & VoteNet & 59.5 & 38.4 & 58.2 & 35.0 \\
        RandomRooms & VoteNet & 61.3 & 36.2 & 59.2 & 35.4 \\
        PointContrast & VoteNet & 59.2 & 38.0 & 57.5 & 34.8 \\
        DepthContrast & VoteNet & 62.1 & 39.1 & 60.4 & 35.4 \\
        Point-M2AE & Point-M2AE & 66.3 & 48.3 & - & - \\
          MaskFeat3D & VoteNet & 63.3 & 41.0 & 61.0 & 36.5 \\
         MaskFeat3D & Point-M2AE & 67.5 & 50.0 & - & - \\
         MaskFeat3D & CAGroup3D & \textbf{75.6} & \textbf{62.3} & \textbf{67.2} & \textbf{51.0}\\
        \bottomrule
    \end{tabular}
    \caption{3D object detection results. For fair comparison, we evaluate our method on two different backbones, Point-M2AE and CAGroup3D.}
    \label{tab:maskfeat_detection}
\end{table}

%% file: chapters/MaskFeat3D/Sections/06-Conclusion.tex
\subsection{Conclusion}

Our study reveals that restoration of masked point location is not essential for 3D MAE training. By predicting geometric features such as surface normals and surface variations at the masked points via our cross-attention-based decoder, the performance of 3D MAEs can be improved significantly, as evaluated through extensive experiments and downstream tasks. Moreover, the performance gains remain consistent when using different encoder backbones. We hope that our study can inspire future research in the development of robust MAE-based 3D backbones.

%% file: chapters/pc_2dto3d.tex
\chapter{Transfer-Learning-Based Representation Learning}\label{ch:pc-2dto3d}

\renewcommand{\thefootnote}{\color{red}\fnsymbol{footnote}}
\footnote[6]{The content of this chapter includes my publication ``Multi-View Representation is What You Need for Point-Cloud Pre-Training''~\cite{yan2023multi} in International Conference on Learning Representations (ICLR) 2024. I am the first author of the publication.}

In the previous chapter, we demonstrated the effectiveness of self-supervised learning methods for point cloud representation learning. While these methods are promising, achieving more robust feature learning often involves scaling up deep models, which in turn requires large-scale datasets. However, the current availability of extensive 3D datasets is limited due to the high costs associated with data acquisition and the labor-intensive nature of data annotation.

To address this challenge, one potential solution is to leverage transfer learning to bridge the gap between 3D pre-training and 2D models. Transfer learning allows us to utilize the rich knowledge embedded in 2D pre-trained models, which have been trained on vast amounts of 2D image data, to improve the performance of 3D point cloud representation learning. This approach can significantly reduce the dependence on large-scale 3D datasets by transferring learned features and representations from the more abundant 2D domain.

In this chapter, we design a new method to transfer the rich knowledge of 2D pre-trained model to help the 3D point cloud representation learning. This method not only bridges the gap between 2D and 3D data but also enhances the robustness and generalization capabilities of 3D point cloud models. Through comprehensive experiments and evaluations, we demonstrate that our approach leads to significant improvements in various 3D tasks, including object recognition, segmentation, and classification. I am the main author of the presented works.

\input{chapters/MVNet/MVNet}

\pagebreak

%% file: chapters/MVNet/MVNet.tex
\section{Multi-View Representation is What You Need for Point-Cloud Pre-Training}\label{sec:mvnet}

\renewcommand{\thefootnote}{\color{red}\fnsymbol{footnote}}
\footnote[6]{The content of this section is based on my publication ``Multi-View Representation is What You Need for Point-Cloud Pre-Training''~\cite{yan2023multi} in International Conference on Learning Representations (ICLR) 2024. I am the first author of the publication.}

A promising direction for pre-training 3D point clouds is to leverage the massive amount of data in 2D, whereas the domain gap between 2D and 3D creates a fundamental challenge. This paper proposes a novel approach to point-cloud pre-training that learns 3D representations by leveraging pre-trained 2D networks. Different from the popular practice of predicting 2D features first and then obtaining 3D features through dimensionality lifting, our approach directly uses a 3D network for feature extraction. We train the 3D feature extraction network with the help of the novel 2D knowledge transfer loss, which enforces the 2D projections of the 3D feature to be consistent with the output of pre-trained 2D networks. To prevent the feature from discarding 3D signals, we introduce the multi-view consistency loss that additionally encourages the projected 2D feature representations to capture pixel-wise correspondences across different views. Such correspondences induce 3D geometry and effectively retain 3D features in the projected 2D features. Experimental results demonstrate that our pre-trained model can be successfully transferred to various downstream tasks, including 3D shape classification, part segmentation, 3D object detection, and semantic segmentation, achieving state-of-the-art performance.

\input{chapters/MVNet/sections/01_introduction}
\input{chapters/MVNet/sections/02_related_work}
\input{chapters/MVNet/sections/03_method}
\input{chapters/MVNet/sections/04_results}
\input{chapters/MVNet/sections/05_ablation}
\input{chapters/MVNet/sections/06_conclusion}

%% file: chapters/MVNet/sections/01_introduction.tex
\subsection{Introduction}
\label{sec:intro}

The rapid development of commercial data acquisition devices~\cite{daneshmand20183d} and point-based deep learning networks~\cite{qi2017pointnet, qi2017pointnet++, choy20194d} has led to a growing research interest in models that can directly process 3D point clouds without voxelization. Remarkable success has been achieved in various applications, including but not limited to object detection~\cite{misra2021end, liu2021group, wang2022cagroup3d}, segmentation~\cite{qian2022pointnext, tang2022contrastive, zhao2021point, yan2021hpnet}, and tracking~\cite{qi2020p2b, zheng2021box, shan2021ptt}.

Despite the significant advances in 3D point cloud processing, acquiring task-specific 3D annotations is a highly expensive and severely limited process due to the geometric complexity. 
The shortage of data annotations highlights the need for adapting pre-training paradigms. Instead of training the deep network from randomly initialized weights, prior work suggests that pre-training the network on a relevant but different pre-task and later fine-tuning the weights using task-specific labels often leads to superior performance. In natural language processing~\cite{devlin2018bert} and 2D vision~\cite{he2022masked,radford2015unsupervised, doersch2015unsupervised, he2020momentum, zhuang2021unsupervised, zhuang2019self}, pre-trained models are the backbones of many exciting applications, such as real-time chatbots~\cite{touvron2023llama, openai2023gpt4} and graphic designers~\cite{meng2021sdedit, wang2022pretraining}. 
However, pre-training on point clouds has yet to demonstrate a universal performance improvement. From-scratch training remains a common practice in 3D vision.

Initial attempts towards 3D point-cloud pre-training primarily leverage contrastive learning~\cite{chopra2005learning}, especially when the point clouds are collected from indoor scenes~\cite{xie2020pointcontrast, rao2021randomrooms, liu2021learning, zhang2021self, chen20224dcontrast}. However, the broad application of contrastive learning-based pre-training techniques is impeded by the requirement of large batch sizes and the necessity to carefully define positive and negative pairs. In contrast to natural language processing and 2D vision, pre-training on 3D point clouds presents two unique challenges. 
\input{chapters/MVNet/figures/radar_chart}
First, the data is extremely scarce, even without annotations. Public 3D datasets are orders of magnitude smaller than 2D image datasets. 
Second, the lack of data annotations necessitates 3D pre-training methods to adhere to the self-supervised learning paradigm. 
Without strong supervision, pre-task design becomes particularly crucial in effective knowledge acquisition.

Figure~\ref{Fig:Model} illustrates our novel approach to 3D pre-training, which is designed to address the aforementioned challenges. Our key idea is to leverage pre-trained 2D networks in the image domain to enhance the performance of 3D representation learning while bridging the 2D-3D domain gap. We begin by applying a 3D-based network to an input point cloud with $M$ points, which generates a 3D feature volume represented by an $M \times C$-dimensional embedding, where $C$ is the length of the per-point feature vectors. To compensate for data scarcity, we align the 3D feature volume with features predicted by a 2D encoder trained on hundreds of millions of images. To this end, we project the 3D feature volume into pixel embeddings and obtain $H \times W \times C$ image-like feature maps in 2D. We then use pre-trained image encoders to extract hierarchical features from corresponding RGB images and train a similarly structured encoder on projected 2D maps to enforce feature alignment. To ensure geometrical meaningfulness, we project the same 3D feature volume into multiple camera views and perform hierarchical feature alignment in each individual view. Our pipeline is generic enough to support all major 2D pre-trained models (e.g., CLIP~\cite{radford2021learning}, SAM~\cite{kirillov2023segment}, DINOv2~\cite{oquab2023dinov2}). In our experiments, DINOv2 exhibits the best performance. 

One issue with the approach described above is that the 3D feature learning may overfit to the pre-trained 2D networks. Consequently, it could potentially discard 3D features that are critical for 3D recognition but not well-captured during image-based 2D pre-training. To address this issue, we introduce an auxiliary pre-task to predict 2D multi-view pixel-wise correspondences from pairs of projected 2D feature maps. As such correspondences induce 3D depth information, the learned feature representations are forced to capture 3D signals. 

Our approach is fundamentally different from existing work based on multi-view representations~\cite{DBLP:conf/iccv/SuMKL15,DBLP:conf/cvpr/KalogerakisAMC17,DBLP:conf/cvpr/KanezakiMN18,DBLP:conf/cvpr/WeiYS20,DBLP:conf/eccv/KunduYFRBFP20,DBLP:conf/iccv/HamdiGG21,DBLP:conf/iclr/HamdiGG23}, which either aggregate image-based analysis results or lift 2D feature maps into 3D for understanding. Our approach employs a 3D network for feature extraction trained with the additional help of \textbf{pre-trained} 2D models. While multi-view based 3D understanding approaches~\cite{DBLP:conf/iccv/SuMKL15,DBLP:conf/cvpr/KalogerakisAMC17} tremendously benefit from the rich experience in 2D vision and have long dominated the design of shape classification networks when 3D data is limited, recent 3D-based understanding networks~\cite{DBLP:conf/cvpr/RieglerUG17,qi2017pointnet,qi2017pointnet++,DBLP:journals/tog/WangLGST17} are shown to outperform their multi-view counterparts. The superior performance is the result of enhanced architecture design and the clear advantage of directly learning 3D patterns for geometry understanding. In our setting, the 3D feature extraction network provides implicit regularization on the type of features extracted from pre-trained 2D models. We seek to combine the advantage of multi-view representations and the geometric capacity of 3D networks, while leveraging the rich knowledge embedded in pre-trained 2D networks. 



In summary, we make the following contributions:

\begin{itemize}[leftmargin=*]
    \item We formulate point-cloud pre-training as learning a multi-view consistent 3D feature volume.
    \item To compensate for data scarcity, we leverage pre-trained 2D image-based models to supervise 3D pre-training through perspective projection and hierarchical feature-based 2D knowledge transfer.
    \item To prevent overfitting to pre-trained 2D networks, we develop an auxiliary pre-task where the goal is to predict the multi-view pixel-wise correspondences from the 2D pixel embeddings. 
    \item We conduct extensive experiments to demonstrate the effectiveness of our approach across a wide range of downstream tasks, including shape classification, part segmentation, 3D detection, and semantic segmentation, achieving consistent improvement over baselines (Figure~\ref{fig:radar}). 
\end{itemize}

%% file: chapters/MVNet/figures/radar_chart.tex
\begin{figure}[h]
    \centering   \includegraphics[width=\textwidth]{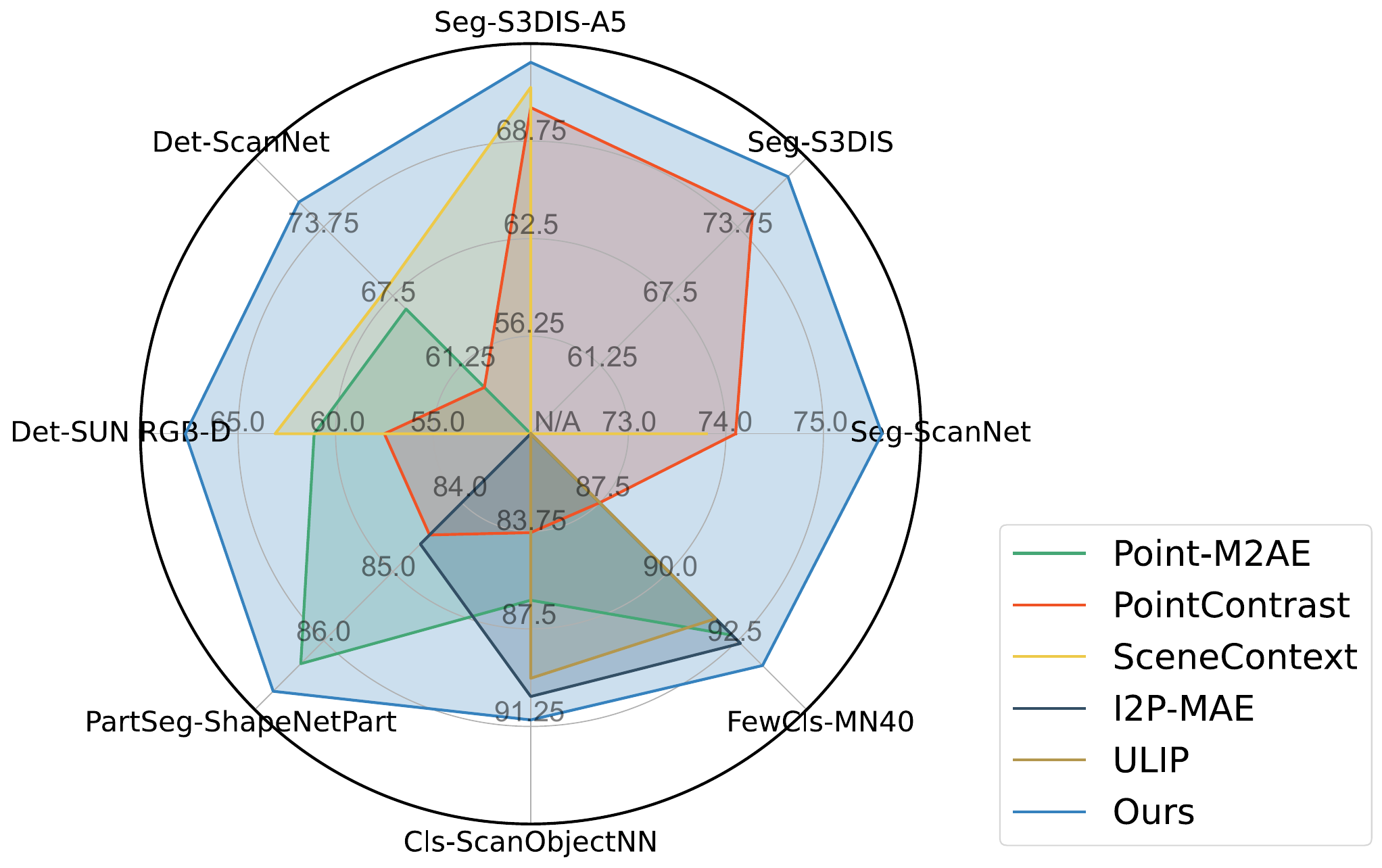}
    \caption{Our model (blue) achieves state-of-the-art performance across a broad range of tasks at both the scene and shape levels. The distance to the origin indicates the task result.}
    \label{fig:radar}
\end{figure}

%% file: chapters/MVNet/sections/02_related_work.tex
\subsection{Related Work}


\paragraph{Point-cloud pre-training}
The success of point-based deep neural networks demonstrates the potential for machine learning models to directly perceive point clouds. Point-based pre-training refers to the practice of \textit{pre-training} the point-based prediction network on one or more \textit{pre-tasks} before fine-tuning the weights on \textit{downstream tasks}. The expectation is that the knowledge about pre-tasks will be transferred to downstream tasks, and the network can achieve better performance than random parameter initialization. 
We refer readers to~\cite{xiao2023unsupervised} for a comprehensive survey of the field, in which we highlight one important observation, stating that point-based pre-training ``still lags far behind as compared with its
counterparts'', and training from-scratch ``is still the prevalent approach''. The challenges and opportunities call for innovations in point-based pre-training methods.

\textit{Shape-level pre-training.}
Self-reconstruction is a popular pre-task at the shape level, where the network encodes the given point cloud as representation vectors capable of being decoded back to the original input data~\cite{yang2018foldingnet, sauder2019self}. To increase the pre-task difficulty, it is common for self-reconstruction-based methods to randomly remove a certain percentage of the points from the input~\cite{wang2021unsupervised, pang2022masked, zhang2022point, yu2022point, yan2023implicit, yan20233d}. In addition to self-reconstruction, 
\cite{rao2020global} develop a multi-task approach that unifies contrastive learning, normal estimation, and self-reconstruction into the same pipeline. There are also attempts to connect 3D pre-training to 2D~\cite{afham2022crosspoint, xu2021image2point, dong2022autoencoders, zhang2022learning}, or to explore the inverse relationship \cite{hou2023mask3d}. For example, \cite{dong2022autoencoders} utilize pre-trained 2D transformers as cross-modal teachers. \cite{zhang2022learning} leverage self-supervised pre-training and masked autoencoding to obtain high-quality 3D features from 2D pre-trained models.
However, these methods focus on synthetic objects, leading to significant performance degradation in downstream tasks using real data from indoor scenes.

\textit{Scene-level pre-training.}
Existing scene-level pre-training methods focus on exploiting the contrastive learning paradigm. As the first pre-training method with demonstrated success at the scene level, PointContrast~\cite{xie2020pointcontrast} defines the contrastive loss using pairs of 3D points across multiple views. RandomRoom~\cite{rao2021randomrooms} exploits synthetic datasets and defines the contrastive loss using pairs of CAD objects instead. Experimentally, RandomRoom additionally confirms that synthetic shape-level pre-training is beneficial to real-world scene-level downstream tasks. DepthContrast~\cite{zhang2021self} simplifies PointContrast by proposing an effective pre-training method requiring only single-view depth scans. 4DContrast~\cite{chen20224dcontrast} further includes temporal dynamics in the contrastive formulation, where the pre-training data comprises sequences of synthetic scenes with objects in different locations. In contrast to these approaches, we focus on utilizing pre-trained 2D networks to boost the performance of 3D pre-training without leveraging contrastive learning.

%% file: chapters/MVNet/sections/03_method.tex
\subsection{Method}
\label{Sec:Method}

We present our pre-training approach pipeline, denoted as \textbf{MVNet} in the following sections. As illustrated in Figure~\ref{Fig:Model}, given a pair of RGB-D scans, we first project the 2D pixels into 3D point clouds using camera parameters. We then extract the point-cloud feature volumes using the feature encoding network (Section~\ref{sec:feat_encode}). Subsequently, we project the feature volume onto two different views to generate 2D feature maps (Section~\ref{sec:projection}). We design the 2D knowledge transfer module to learn from large-scale 2D pre-trained models (Section~\ref{sec:2d_transfer}). Finally, we utilize the multi-view consistency module to ensure the agreement of different view features (Section~\ref{sec:consistent}), promoting that the 3D feature encoding network extracts 3D-aware features from pre-trained 2D image models. Finally, the weights of the feature encoding network are transferred to downstream tasks for fine-tuning.

\input{chapters/MVNet/figures/main_figure}

\subsubsection{Feature Encoding Network}
\label{sec:feat_encode}

Let $(I_1, D_1)$ and $(I_2, D_2)$ be two RGB-D scans from the same scene, where $I_1$ and $I_2$ denote the RGB images, and $D_1$ and $D_2$ represent the depth maps. Using camera parameters, we project the RGB-D scans into a colored point cloud $P \in \mathbb{R}^{M \times 6}$ with $M$ points, with the first three channels representing coordinates and the remaining three channels representing RGB values. The feature encoding network of MVNet takes $P$ as input and outputs a dense feature field $F_P \in \mathbb{R}^{M \times C}$, where each point is associated with a feature vector of dimension $C$.


The network architecture adopts the design of Sparse Residual U-Net (SR-UNet)~\cite{choy20194d}. Our network includes 21 convolution layers for the encoder and 13 convolution layers for the decoder, where the encoder and the decoder are connected with extensive skip connections. Specifically, the input point cloud $P$ is first voxelized into $M'$ voxels, yielding a grid-based representation $V \in \mathbb{R}^{M'\times6}$. In our implementation, we use the standard Cartesian voxelization with the grid length set to 0.05 m. The output of the network is a set of $C$-dimensional per-voxel features, $F_V \in \mathbb{R}^{M' \times C}$, jointly generated by the encoder and the decoder. Due to space constraints, we refer interested readers to the supplementary material for implementation details.

Next, we interpolate the per-voxel features $F_V$ to obtain a $C$-dimensional feature vector for each of $M$ points in the input point cloud. This is achieved via the classical KNN algorithm based on point-to-point distances. As shown in the middle of Figure~\ref{Fig:Model}, the resulting $F_P$ is our point-cloud feature volume, awaiting further analysis and processing.

Importantly, unlike existing work that utilizes multi-view representations from 2D feature extraction networks, MVNet exploits a 3D feature extraction network. As a result, MVNet has the intrinsic ability to capture valuable 3D patterns for 3D understanding. 




\subsubsection{Point-Cloud Feature Volume Projection}
\label{sec:projection}

The key idea of MVNet is to learn 3D feature representations through 2D projections, enabling us to leverage large-scale pre-trained models in 2D. To this end, we proceed to project the feature volume $F_P$ back onto the two input views, generating 2D feature maps, $F_1$ and $F_2$, both with dimensions $H \times W \times C$. Here, $H \times W$ is the spatial resolution of the input.

The projection operation uses the one-to-one mapping between each pixel in each input view and the corresponding 3D point in the concatenated 3D point cloud $P$. We refer readers to the supplementary material for the implementation details and the conversion formula between 2D and 3D coordinates.





\subsubsection{2D Knowledge Transfer Module}
\label{sec:2d_transfer}
\input{chapters/MVNet/figures/2d_transfer}
The first feature learning module of MVNet, $\mathcal{F}_{2d}$, aims to transfer knowledge from large-scale 2D pre-trained models. 
To this end, we consider each input RGB image $I_i$ and the corresponding projected 2D feature map $F_i$. Our goal is to train $F_i$ using a pre-trained 2D model $f_p$ that takes $I_i$ as input. 
Our implementation uses ViT-B~\cite{dosovitskiy2020image} as the network architecture. More details are provided in Section~\ref{sec:2d_ablation}.
The technical challenge here lies in the semantic differences between $F_i$ and $I_i$. As illustrated in Figure~\ref{fig:2d_transfer}, we introduce an additional network $f_k$ that takes $F_i$ as input. $f_k$ shares the same network architecture as $f_p$ except the first layer, which is expanded channel-wise to accommodate the $C$-dimensional input. Let $N$ be the number of layers in $f_k$ and $f_p$. We define the training loss as 

\begin{equation}
\mathcal{L}_{2d} = \sum_{j=3}^{N} \lVert B_j^{\text{pre-trained}} - B_j^{\text{2D}} \rVert_2^2
\label{eq:transfer_loss}
\end{equation}

where $B_j^{\text{pre-trained}}$ denotes the output feature map of the $j^\text{th}$ block in the pre-trained 2D model $f_p$, and $B_j^{\text{2D}}$ denotes the output feature map of the $j^\text{th}$ block in $f_k$. The objective of the 2D knowledge transfer process is to minimize $\mathcal{L}_{\text{2d}}$, thereby enabling $f_k$ to learn the hierarchical features through knowledge transfer from the pre-trained 2D model. We add the loss starting from the third layer, which experimentally leads to the best performance. 

During training, the weights of the 2D pre-trained model $f_p$ are unchanged, whereas $f_k$ and the feature encoding network are jointly optimized. In our implementation, we select ViT-B~\cite{dosovitskiy2020image} as the network backbone for both 2D neural network $f_k$ and $f_p$. We take the pre-trained weights of DINOv2~\cite{oquab2023dinov2} for $f_p$. DINOv2 has been demonstrated to generate features with strong transferability. We refer readers to Section~\ref{sec:2d_ablation} for a detailed analysis of these design choices through extensive ablation studies.

\input{chapters/MVNet/figures/multi_view}

\subsubsection{Multi-View Consistency Module}
\label{sec:consistent}

The goal of the previous module is to leverage pre-trained 2D networks. However, pre-trained 2D networks only contain feature representations that are primarily suitable for 2D tasks. A complete reliance on the 2D knowledge transfer module encourages the 3D feature encoding network to discard geometric cues that are important for 3D recognition. To address this issue, we introduce a novel multi-view consistency module $\mathcal{F}_m$, where the goal is to use the projected features, $F_1$ and $F_2$, to predict dense correspondences between the two corresponding input images. Prior research has demonstrated that dense pixel-wise correspondences between calibrated images allows the faithful recovery of 3D geometry~\cite{Zhou_2016_CVPR}. Therefore, enforcing ground-truth correspondences to be recoverable from $F_1$ and $F_2$ prevents the learned feature volume from discarding 3D features. 

Our dense correspondence module adopts a cross attention-based transformer decoder $\mathcal{D}$. As depicted in Figure~\ref{fig:consistent}, we concatenate the two views' feature maps side by side, forming a feature map $F_c$. The context feature map $F_c$ is fed into a cross attention-based transformer decoder $\mathcal{D}$, along with the query point $x$ from the first view $I_1$. 
Finally, we process the output of the transformer decoder with a fully connected layer to obtain our estimate for the corresponding point, $x'$, in the second view $I_2$:
\begin{equation}
x' = \mathcal{F}_m(x| F_1, F_2) = \mathcal{D}(x, F_1 \oplus F_2)
\end{equation}
Following~\cite{jiang2021cotr}, we design the loss for correspondence error and cycle consistency:
\begin{equation}
\small
\mathcal{L}_{m} = ||x^{gt} - x'||^2_2 + ||x - \mathcal{F}_m(x'|F_1, F_2)||^2_2
\label{Eq:consistency:loss}
\end{equation}
where $x^{gt}$ denotes the ground truth corresponding point of $x$ in the second view $I_2$.

Combing (\ref{eq:transfer_loss}) and (\ref{Eq:consistency:loss}), the overall loss function is defined as:
\begin{equation}
\mathcal{L} = \mathcal{L}_{2d} + \lambda \mathcal{L}_m
\end{equation}
where $\lambda=0.5$ is the trade-off parameter.

%% file: chapters/MVNet/figures/main_figure.tex
\begin{figure}[h]
    \begin{center}
        \includegraphics[width=1\linewidth]{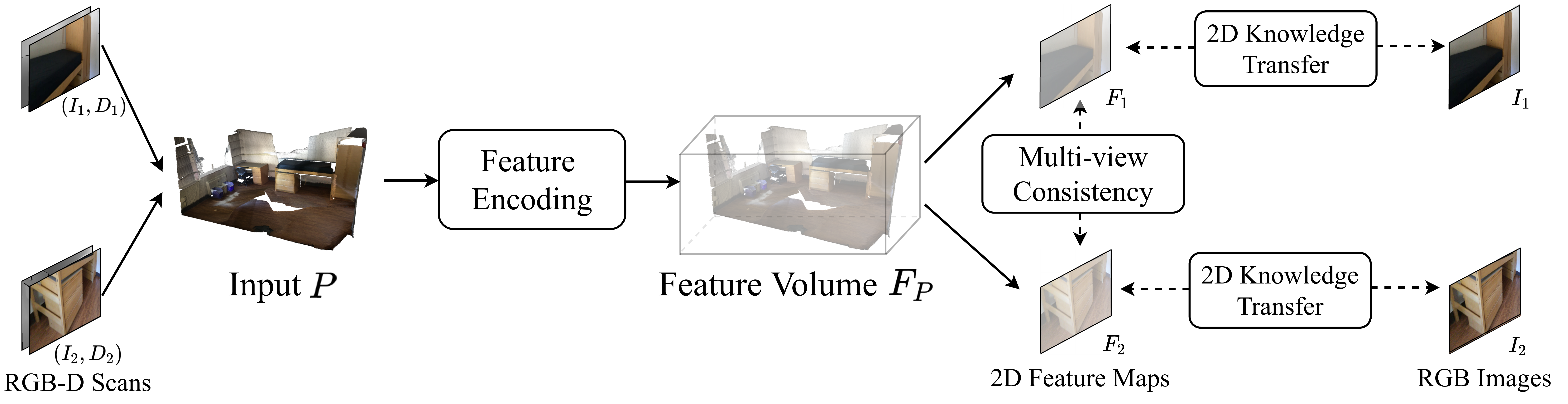}
    \end{center}
    \caption{Approach overview. We used complete and semi-transparent point clouds to represent the input $P$ and the feature volume $F_P$ for better visualization. We encourage readers to frequently reference to this figure while reading Section~\ref{Sec:Method}.}
    \label{Fig:Model}
\end{figure}

%% file: chapters/MVNet/figures/2d_transfer.tex
\begin{figure}[h]
\centering 
\includegraphics[width=0.8\textwidth]{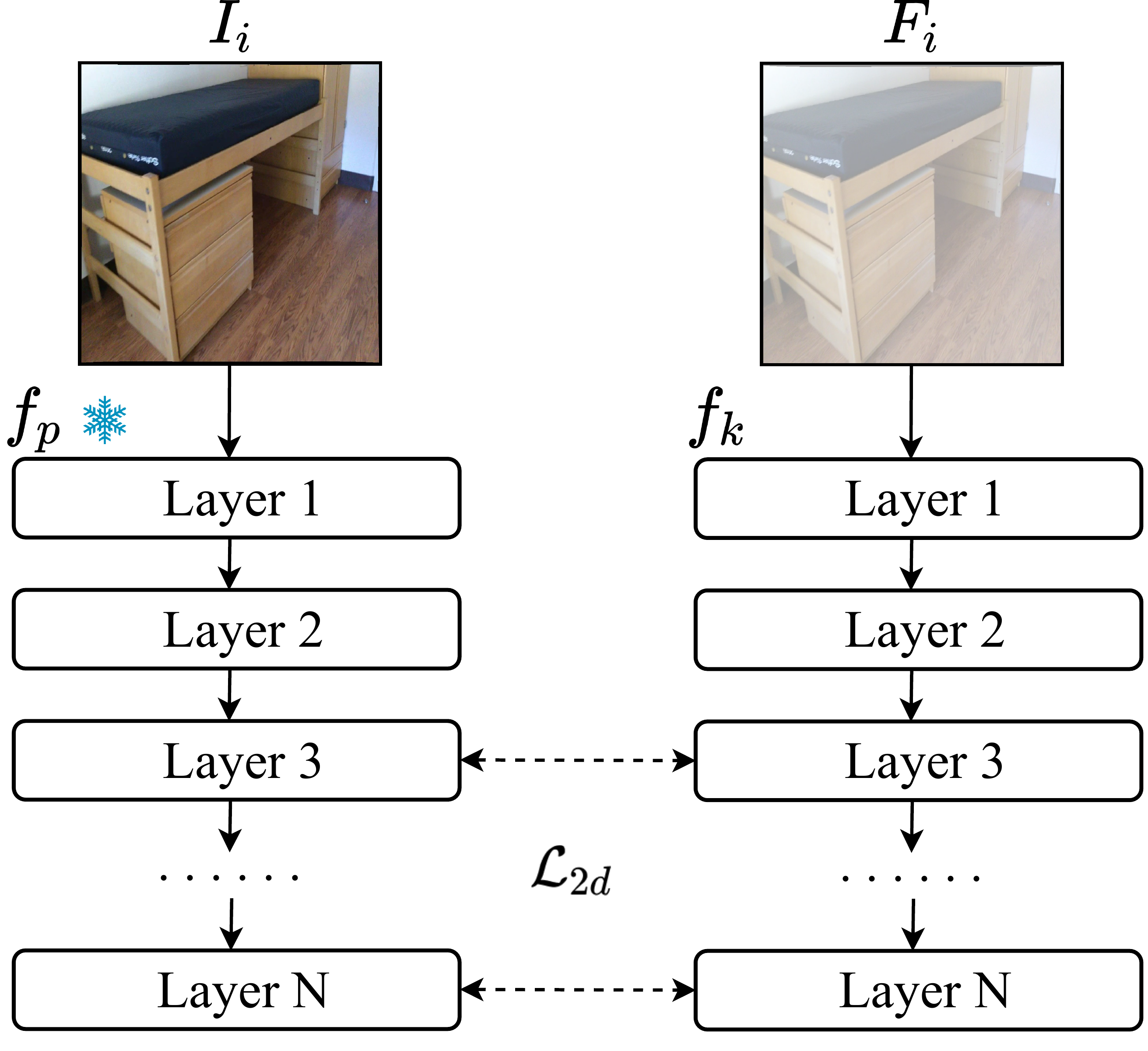}
  \caption{2D knowledge transfer module. $f_p$ is completely frozen during training. For simplicity, we represent the $C$-dimensional feature map $F_i$ using a semi-transparent image.}
\label{fig:2d_transfer}
\end{figure}

%% file: chapters/MVNet/figures/multi_view.tex
\begin{figure}[h]
\centering
\includegraphics[width=0.6\textwidth]{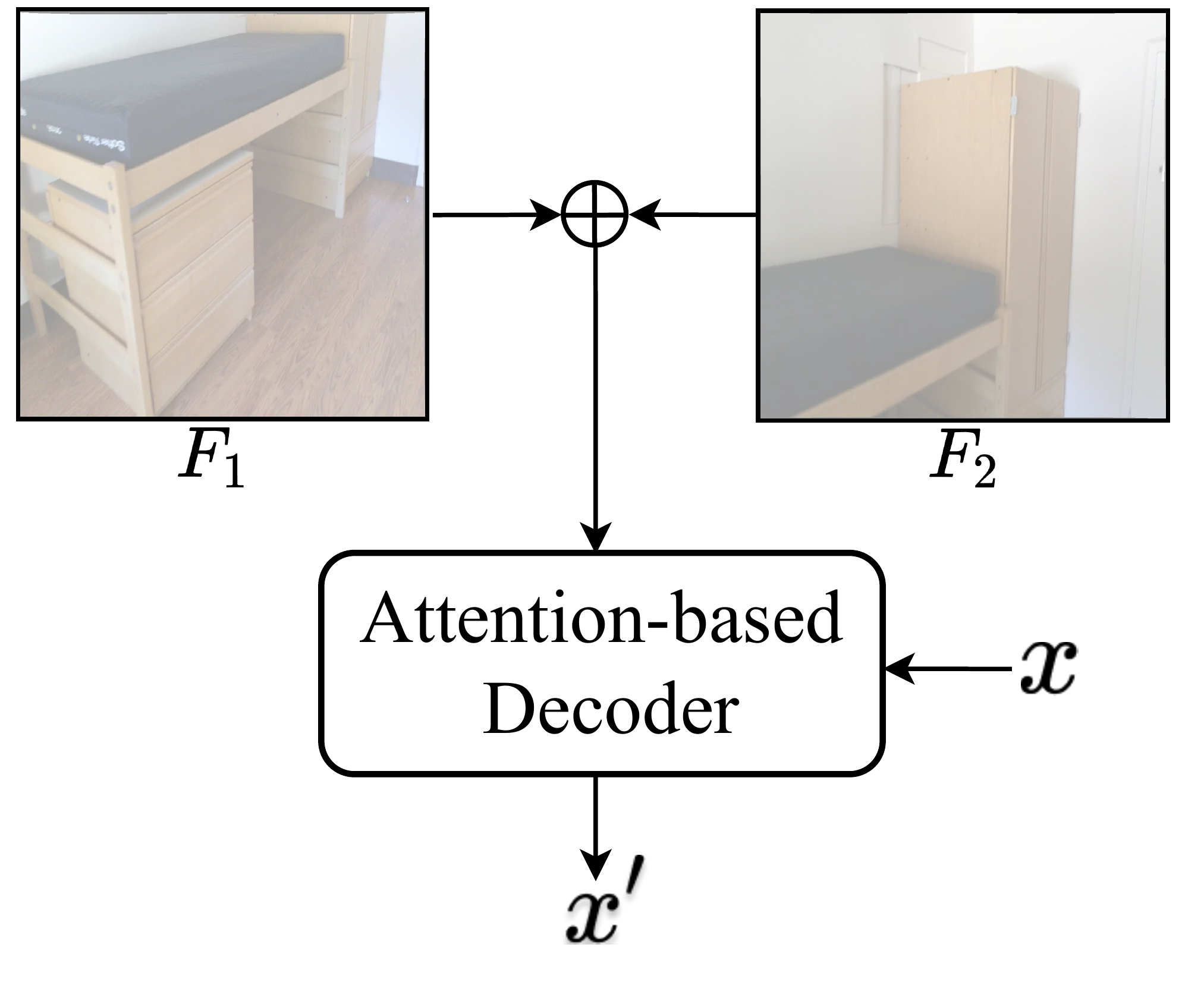}
\caption{Illustration of multi-view consistency module. $\oplus$ denotes side-by-side concatenation.}
\label{fig:consistent}
\end{figure}

%% file: chapters/MVNet/sections/04_results.tex
\subsection{Experimental Analysis}
\label{Sec:Experiment}

\subsubsection{Pre-Training Setting}
\label{sec:pretrain_setting}

\input{chapters/MVNet/tables/segmentation_4}
\paragraph{Data preparation}
We choose ScanNet~\cite{dai2017scannet} as the pre-training dataset, which contains approximately 2.5M RGB-D scans from 1,513 indoor scenes. Following \cite{qi2019deep}, we down-sample 190K RGB-D scans from 1,200 video sequences in the training set. For each scan, we build the scan pair by taking it as the \textit{first} scan $(I_1, D_1)$ and select five other frames with an overlap ratio between $[0.4, 0.8]$ as the \textit{second} scan $(I_2, D_2)$ candidates. We also evaluate our model at the shape level. We pre-trained our model on Objaverse~\cite{deitke2023objaverse}, which contains approximately 800K real-world 3D objects. We also pre-trained our model on ShapeNet~\cite{chang2015shapenet} for fair comparison with previous approaches. For each 3D object, we use Blender~\cite{kent20153d} to render 6 images with their camera angles spaced equally by 60 degrees. At each training step, we randomly choose two consecutive images for pre-training.

\vspace{-5pt}

\paragraph{Network training}
Our pre-training model is implemented using PyTorch, employing the AdamW optimizer~\cite{loshchilov2017decoupled} with a weight decay of $10^{-4}$. The learning rate is set to $10^{-3}$. The model is trained for 200 epochs on eight 32GB Nvidia V100 GPUs, with 64 as the batch size. 

\subsubsection{Downstream Task}

The goal of pre-training is to learn features that can be transferred effectively to various downstream
tasks. In the following experiments, we adopt the \textbf{supervised fine-tuning strategy}, a popular way to evaluate the transferability of pre-trained features. Specifically, we initialize the network with pre-trained weights as described in Section~\ref{Sec:Method} and fine-tune the weights for each downstream task.

\noindent{\textbf{Semantic segmentation }}
We use the original training and validation splits of ScanNet~\cite{dai2017scannet} and report the mean Intersection over Union (IoU) on the validation split. We also evaluate on S3DIS~\cite{2017arXiv170201105A} that has six large areas, where one area is chosen as the validation set, and the remaining areas are utilized for training. We report the Overall Accuracy (OA), mean Accuracy (mAcc), and mean IoU for Area-5 and the 6-fold cross-validation results.


We adopt SR-UNet~\cite{choy20194d} as the semantic segmentation architecture and add a head layer at the end of the feature encoding network for semantic class prediction. Although it is intuitive to transfer both the encoder and decoder pre-trained weights in segmentation fine-tuning, we observe that only transferring the encoder pre-trained weights results in a better performance.
We fine-tune our model using the SGD+momentum optimizer with a batch size of 48, 10,000 iterations, an initial learning rate of 0.01, and a polynomial-based learning rate scheduler with a power factor of 0.9. The same data augmentation techniques in~\cite{chen20224dcontrast} are used.

As demonstrated in Table~\ref{tab:semseg}, on both datasets, we achieve the best performance compared to other pre-training methods, with +1.5 val mIoU improvement on ScanNet compared to the second-best method. On the S3DIS dataset, we report +1.6 mIoU improvement on the Area 5 validation set.

\input{chapters/MVNet/tables/votenet_detection}
\noindent{\textbf{3D object detection}
ScanNet~\cite{dai2017scannet} contains instance labels for 18 object categories, with 1,201 scans for training and 312 for validation. SUN RGB-D~\cite{song2015sun} comprises of 5K RGB-D images annotated with amodal, 3D-oriented bounding boxes for objects from 37 categories.

We employ the encoder of the SR-UNet~\cite{choy20194d} as the detection encoder and transfer the pre-trained weights for network initialization. For the decoder design of 3D detection, we consider VoteNet~\cite{qi2019deep} and CAGroup3D~\cite{wang2022cagroup3d}. VoteNet is a classical 3D detection approach commonly used by existing pre-training methods~\cite{xie2020pointcontrast, zhang2021self, rao2021randomrooms, huang2021spatio}. However, VoteNet's performance lags behind the state-of-the-art. Therefore, we also evaluate our approach on CAGroup3D, which offers a stronger decoder architecture and delivers state-of-the-art 3D detection performance.
During training, for the VoteNet setting, we fine-tune our model with the Adam optimizer (the batch size is 8, and the initial learning rate is 0.001). The learning rate is decreased by a factor of 10 after 80 and 120 epochs. For the CAGroup3D setting, we decrease the learning rate at 80 and 110 epochs.


As shown in Table~\ref{tab:detection_vote}, on ScanNet, our approach (MVNet) improves VoteNet by +4.5/+3.1 points on mAP@0.25 and mAP@0.5, outperforming other pre-training approaches. When replacing VoteNet with CAGroup3D, MVNet further elevates the mAP@0.25/mAP@0.5 to 76.0/62.4. On the SUN RGB-D dataset, MVNet also exhibits consistent improvements. Specifically, MVNet achieves +3.8/+4.6 points on mAP@0.25 and mAP@0.5 under the VoteNet backbone and 67.6/51.7 points on mAP@0.25 and mAP@0.5 under the CAGroup3D backbone.

\noindent{\textbf{3D shape part segmentation}
In addition to scene-level downstream tasks, we also evaluate MVNet at the shape level. Our experiments on 3D shape part segmentation utilizes the ShapeNetPart~\cite{10.1145/2980179.2980238}, which comprises 16,880 models across 16 shape categories. We adopt the standard evaluation protocol as described by~\cite{qi2017pointnet++}, sampling 2,048 points from each shape. We report the evaluation metrics in terms of per-class mean Intersection over Union (cls. mIoU) and instance-level mean Intersection over Union (ins. mIoU).

To make a fair comparison, in accordance with~\cite{yu2022point}, we employ a standard transformer network with a depth of 12 layers, a feature dimension of 384, and 6 attention heads, denoted as MVNet-B. The experimental results are summarized in Table~\ref{tab:shape-result}. With the same network backbone, our approach demonstrates consistent improvements. Additionally, more results involving pre-training approaches with different network backbones can be found in the supplementary material.

\input{chapters/MVNet/tables/shape_task}

\noindent{\textbf{3D shape classification}
We also conduct experiments on the real-world ScanObjectNN dataset~\cite{UyPHNY19}, which contains approximately 15,000 scanned objects distributed across 15 classes. The dataset is partitioned into three evaluation splits: OBJ-BG, OBJ-ONLY, and PB-T50-RS, where PB-T50-RS is considered the most challenging one. We employ the same transformer-based classification network (MVNet-B) as described by~\cite{yu2022point}. The experimental outcomes, presented in Table~\ref{tab:shape-result}, reveal consistent performance gains across all three test sets.

\noindent{\textbf{Model scalability}
The scalability of a model is a crucial factor when evaluating the effectiveness of a pre-training approach. Unfortunately, many previous methods overlook this aspect. The commonly used transformer architecture has limited capacity due to the small number of parameters. To further substantiate the efficacy of our proposed approach, we improve the network's feature dimension to 768 and double the number of attention heads to 12. As evidenced by the results in Table~\ref{tab:shape-result}, our scaled model, MVNet-L, attains significant improvements across all evaluated settings.

%% file: chapters/MVNet/tables/segmentation_4.tex
\begin{table*}
\centering
\tiny
\setlength\tabcolsep{3pt}
\begin{tabular}{lllllllll}
    \toprule
    \multirow{2}{*}{Method} & \multirow{2}{*}{Backbone} & \multicolumn{3}{c}{S3DIS Area 5} & \multicolumn{3}{c}{S3DIS 6-Fold} & ScanNet \\
     \cmidrule(lr){3-5} \cmidrule(lr){6-8} \cmidrule(lr){9-9} & & OA & mAcc & mIoU & OA & mAcc & mIoU & mIoU \\
    \midrule
    Jigsaw~\citep{sauder2019self} & DGCNN & 82.8 & - & 52.1 & 84.4 & - & 56.6& - \\
    OcCo~\citep{wang2020unsupervised} & DGCNN & 85.9 & - & 55.4 & 85.1 & - & 58.5 & - \\
    STRL~\citep{huang2021spatio} & DGCNN & 82.6  & - & 51.8 & 84.2 & - & 57.1 & -\\ 
    MM-3DScene~\citep{xu2023mm} & PointTrans & - & 78.0 & 71.9 & - & - & - & 72.8 \\
    PointContrast~\citep{xie2020pointcontrast} & SR-UNet & - & 77.0 & 70.9 & - & - & - & 74.1 \\ 
    DepthContrast~\citep{zhang2021self} & SR-UNet & -  & - & 70.6 & - & - & - & 71.2\\ 
    SceneContext~\citep{hou2021exploring} & SR-UNet & - & - & 72.2 & - & - & - & 73.8 \\ 
     MVNet & SR-UNet & \textbf{91.7}(91.6) & \textbf{79.5}(79.3) & \textbf{73.8}(73.3) & \textbf{91.8}(91.7) & \textbf{86.3}(86.2) & \textbf{78.3}(78.1) & \textbf{75.6}(75.2)\\
    
    \bottomrule
\end{tabular}
\caption{Comparison of semantic segmentation results with pre-training methods. We show a comprehensive evaluation across all the benchmarks. The average result of 3 runs is given in parentheses.}

\label{tab:semseg}
\end{table*}

%% file: chapters/MVNet/tables/votenet_detection.tex
\begin{table*}[t]
\footnotesize
\centering
   \setlength\tabcolsep{4pt}
    \begin{tabular}{lllllll}
        \toprule
     \multirow{2}{*}{Method} & \multirow{2}{*}{\parbox{2cm}{Detection\\Architecture}} & \multicolumn{2}{c}{ScanNet} & \multicolumn{2}{c}{SUN RGB-D} \\
        \cmidrule(lr){3-4}   \cmidrule(lr){5-6} 
       &  & mAP@0.25 & mAP@0.5 & mAP@0.25 & mAP@0.5 \\
        \midrule
        STRL~\citep{huang2021spatio} & VoteNet & 59.5 & 38.4 & 58.2 & 35.0 \\
        Point-BERT~\citep{yu2022point} & 3DETR &  61.0 & 38.3 & - & - \\
        RandomRooms~\citep{rao2021randomrooms} & VoteNet & 61.3 & 36.2 & 59.2 & 35.4 \\
        MaskPoint~\citep{Liu2022maskdis} & 3DETR & 63.4 & 40.6 & - & - \\
        PointContrast~\citep{xie2020pointcontrast} & VoteNet & 59.2 & 38.0 & 57.5 & 34.8 \\
        DepthContrast~\citep{zhang2021self} & VoteNet & 62.1 & 39.1 & 60.4 & 35.4 \\
        MM-3DScene~\citep{xu2023mm} & VoteNet & 63.1 & \textbf{41.5} & 60.6 & 37.3 \\
        SceneContext~\citep{hou2021exploring} & VoteNet & - & 39.3 & - & 36.4 \\
        4DContrast~\citep{chen20224dcontrast} & VoteNet & - & 40.0 & - & 38.2 \\
         MVNet & VoteNet & \textbf{64.0}(63.4) & \textbf{41.5}(41.0) & \textbf{62.0}(61.7) & \textbf{39.6}(39.2) \\
         MVNet & CAGroup3D & \textbf{76.0}(75.7) & \textbf{62.4}(61.8) & \textbf{67.6}(67.2) & \textbf{51.7}(51.2) \\
        \bottomrule
    \end{tabular}
\caption{Comparison of 3D object detection results with pre-training methods. 
We show mean Average Precision~(mAP) across all semantic classes with 3D IoU thresholds of 0.25 and 0.5. The average result of 3 runs is given in parentheses.}
\label{tab:detection_vote}
\vspace{-4mm}
\end{table*}

%% file: chapters/MVNet/tables/shape_task.tex
\begin{table*}[t]
    \centering
    \resizebox{1.\linewidth}{!}{  
    \begin{tabular}{lcccccc}
        \toprule
        \multirow{2}{*}{Method} & \multirow{2}{*}{Dataset} & \multicolumn{3}{c}{ScanObjectNN} & \multicolumn{2}{c}{ShapeNetPart} \\
        \cmidrule(lr){3-5} \cmidrule(lr){6-7} & & OBJ-BG & OBJ-ONLY & PB-T50-RS & ins. mIoU & cls. mIoU \\
        \midrule
        Transformer$^{\dagger}$ ~\citep{yu2022point} & - & 79.9 & 80.6 & 77.2 & 85.1 & 83.4 \\
        PointBERT~\citep{yu2022point} & ShapeNet & 87.4 & 88.1 & 83.1 & 85.6 & 84.1 \\
        MaskDiscr~\citep{Liu2022maskdis} & ShapeNet & 89.7 & 89.3 & 84.3 & 86.0 & 84.4 \\
        MaskSurfel~\citep{zhang2022masked} & ShapeNet & 91.2 & 89.2 & 85.7 & 86.1 & 84.4 \\
        ULIP~\citep{xue2023ulip} & ShapeNet & 91.3 &  89.4 & 86.4 & - & - \\
        PointMAE~\citep{pang2022masked} & ShapeNet & 90.0 & 88.3 & 85.2 & 86.1 & - \\
        MVNet-B & ShapeNet & \textbf{91.4} & \textbf{89.7}  & \textbf{86.7} & 86.1 & \textbf{84.8} \\
        MVNet-B & Objaverse & \textbf{91.5} & \textbf{90.1}  & \textbf{87.8} & \textbf{86.2} & \textbf{84.8} \\       
        MVNet-L & Objaverse & \textbf{95.2} & \textbf{94.2}  & \textbf{91.0} & \textbf{86.8} & \textbf{85.2} \\
        \bottomrule
    \end{tabular}
    }
    \caption{Performance comparison with other pre-training approaches on shape-level downstream tasks. All the methods and MVNet-B use the same transformer backbone architecture. $^{\dagger}$ represents the \textit{from scratch} results and all other methods represent the \textit{fine-tuning} results using pretrained weights.}
    \label{tab:shape-result}
\end{table*}

%% file: chapters/MVNet/sections/05_ablation.tex
\subsubsection{Ablation Study}
\label{Sec:Ablation}

We proceed to present an ablation study to justify our design choices. Due to the space constraint, we focus on the semantic segmentation results using the ScanNet dataset.

\input{chapters/MVNet/figures/ablation_2d_model}
\noindent{\textbf{Choice of 2D pre-trained models }}
\label{sec:2d_ablation}
As one of the most popular backbone architectures, ViT~\cite{dosovitskiy2020image} has two variants: ViT-S and ViT-B. ViT-S (Small) contains 12 transformer blocks, each of which has 6 heads and 384 hidden dimensions. ViT-B (Base) contains 12 transformer blocks, each of which has 12 heads and 768 hidden dimensions. We also compare four different pre-training methods. As shown in Figure~\ref{fig:ab_2d_model}, `ViT-X',`CLIP-X', `SAM-X', and `DINOv2-X' represent pre-training the model by ImageNet-1K classification~\cite{deng2009imagenet}, the model with the CLIP approach~\cite{radford2021learning}, the model with the SAM approach~\cite{kirillov2023segment}, and the model with DINOv2~\cite{oquab2023dinov2}, respectively. We observe that all the models show improvements compared with training from scratch (72.2 mIoU). Larger models (X-B) show consistent improvements compared with smaller models (X-S). DINOv2-B yields the best performance in terms of semantic segmentation accuracy. This demonstrates the advantage of leveraging DINOv2's strong transferability properties in our method.

\noindent{\textbf{Loss design }}
\label{sec:loss_design}
Table~\ref{tab:ab_loss} demonstrates the performance of our method with different loss configurations. We investigate the effectiveness of each loss component: $\mathcal{L}_{2d}$ (2D knowledge transfer loss) and $\mathcal{L}_m$ (multi-view consistency loss). The table shows that using either loss independently leads to an improvement in performance. However, combining both losses yields the best results, validating the importance of incorporating both 2D knowledge transfer and multi-view consistency in our approach.

\noindent{\textbf{Masking ratio }}
\label{sec:masking_ratio}
During pre-training, our approach involved masking the input point cloud guided by the corresponding RGB image. Following the conventional practice in masked autoencoder-based pre-training~\cite{he2022masked}, we divided the image into regular non-overlapping patches. Then we sample a subset of patches and mask the remaining ones. The remaining part is projected onto the point cloud, forming the input for our method. We employ this masking operation as a data augmentation technique, which assists in the development of more robust features during pre-training. As evidenced in Table~\ref{tab:ab_mask}, an appropriate masking operation (i.e., 30\%) during pre-training could enhance performance in downstream tasks.

\input{chapters/MVNet/tables/ablation_1}

\noindent{\textbf{View overlap ratio}}
In Table~\ref{tab:ab_overlap}, we examine the impact of the view overlap ratio on the performance of our method. The view overlap ratio refers to the percentage of a shared field of view between the two views used for multi-view consistency. It is important to note that in this experiment, we only use \textbf{one} candidate second view. Compared to the results without the multi-view consistency loss (74.7 mIoU), overlaps between 40\% and 80\% all result in improvements. However, when the overlap is too small, specifically, at 20\%, we fail to observe any improvement. We believe this is because when the overlap is insufficient, multi-view correspondences become too sparse, obtaining very limited 3D signals. As a result, our default setting selects views with overlaps in the range of $[40\%, 80\%]$ as candidate second views, which ensures a balance between shared and unshared information for robust model performance.

\noindent{\textbf{Number of views}}
\label{sec:view_choice}
In Table~\ref{tab:ab_view}, we investigate the impact of using different numbers of candidate second views. We experiment with 1 view, 2 views, 4 views, and 5 views. $\mathcal{L}_{2d}$ denotes only adding 2D knowledge transfer loss, while $\mathcal{L}_{m}$ represents only adding multi-view consistency loss. We find that incorporating more views leads to improved performance for both modules. However, it shows minor performance variations for 2D knowledge transfer module. Our interpretation is that since the loss is applied on each view independently, the different number of candidate views exert minimal influence. Note that we did not experiment with more views as we observed that when the number of views exceeds 5, some candidates have a very small overlap with the first scan, which, as proven in the previous paragraph, can impede the learning of the multi-view consistency module.

\input{chapters/MVNet/figures/ablation_layer_loss}

\paragraph{2D knowledge transfer loss}
\label{sec:2d_knowledge_transfer_layer_loss}

We proceed to investigate the optimal locations to enforce the 2D knowledge transfer loss. We explore the impact of incorporating the loss starting from different layers of $f_k$. As shown in Figure~\ref{fig:ab_2d_layer}, adding the feature-based loss starting from the third layer yields the best performance in terms of semantic segmentation val mIoU on ScanNet. When adding the loss on the first and second layers, the model's performance is negatively impacted. We believe this is because the inputs of the 2D pre-trained model and the 2D neural network differ. Forcefully adding the loss on the early layers may not be beneficial. Furthermore, the performance drops when the loss is added starting from the fourth layer and beyond. This demonstrates that only regularizing the last few layers weakens the knowledge transferred from the pre-trained networks.  

%% file: chapters/MVNet/figures/ablation_2d_model.tex
\begin{figure}[h]
    \centering
\includegraphics[width=.7\linewidth]{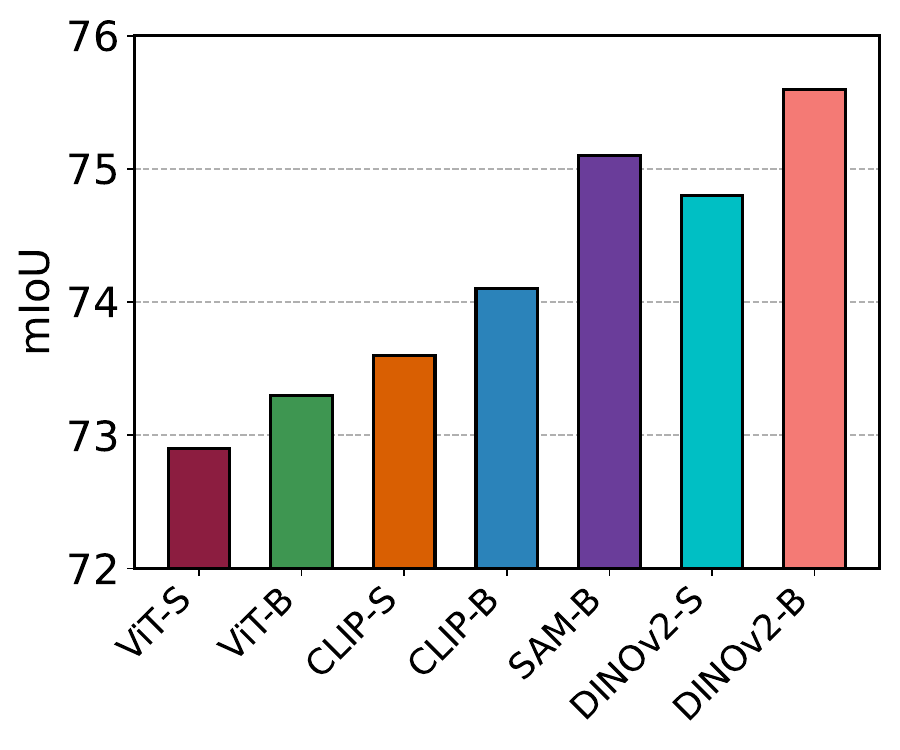}
    \caption{Comparison of different 2D pre-trained models. 
    }
    \label{fig:ab_2d_model}
\end{figure}

%% file: chapters/MVNet/tables/ablation_1.tex
\begin{table*}[t]

\setlength\tabcolsep{5pt}

\begin{minipage}[b]{.5\linewidth}
\centering  
\begin{tabular}{ccc}
        \toprule
        $\mathcal{L}_{2d}$ &$\mathcal{L}_m$ & ScanNet \\
        \midrule
        - & - & 72.2\\
         $\surd$ & - & 74.7\\
        -& $\surd$ &  73.6\\
         $\surd$ & $\surd$ & \textbf{75.6}\\
        \bottomrule
\end{tabular}
\caption{Loss design}
\label{tab:ab_loss}
\end{minipage}
\begin{minipage}[b]{.5\linewidth}
\centering  
\begin{tabular}{cc}
        \toprule
        Ratio (\%) & ScanNet \\ 
        \midrule
        10 & 74.8 \\
        30 & \textbf{75.6} \\
        60 & 74.2 \\
        80 & 73.5 \\
        \bottomrule
\end{tabular}
\caption{Masking ratio}
\label{tab:ab_mask}
\end{minipage}
\begin{minipage}[b]{.5\linewidth}
\centering 
    \begin{tabular}{cc}
        \toprule
     Overlap (\%) & ScanNet \\
        \midrule
        20 & 74.7 \\
        40 & 75.1 \\
       60 & 75.3 \\ 
       80 & \textbf{75.4} \\
       \bottomrule
    \end{tabular}
    \caption{View overlap} 
\label{tab:ab_overlap}
\end{minipage}
\begin{minipage}[b]{.5\linewidth}
\centering 
    \begin{tabular}{ccc}
        \toprule
     Num & $\mathcal{L}_{2d}$ & $\mathcal{L}_m$ \\
        \midrule
        1 & 74.5 & 72.9 \\
        2 & 74.6 & 73.1 \\
        4 & 74.6 & 73.3 \\
         5 & \textbf{74.7} & \textbf{73.6} \\
        \bottomrule
    \end{tabular}
    \caption{View number}
\label{tab:ab_view}
\end{minipage}

\caption{Ablation studies of our design choices. We report the best val mIoU result on ScanNet. Please refer to Section~\ref{Sec:Ablation} for a detailed analysis.}
\label{tab:ablation}
\end{table*}

%% file: chapters/MVNet/figures/ablation_layer_loss.tex
\begin{figure}[h]
    \centering
    \includegraphics[width=.7\linewidth]{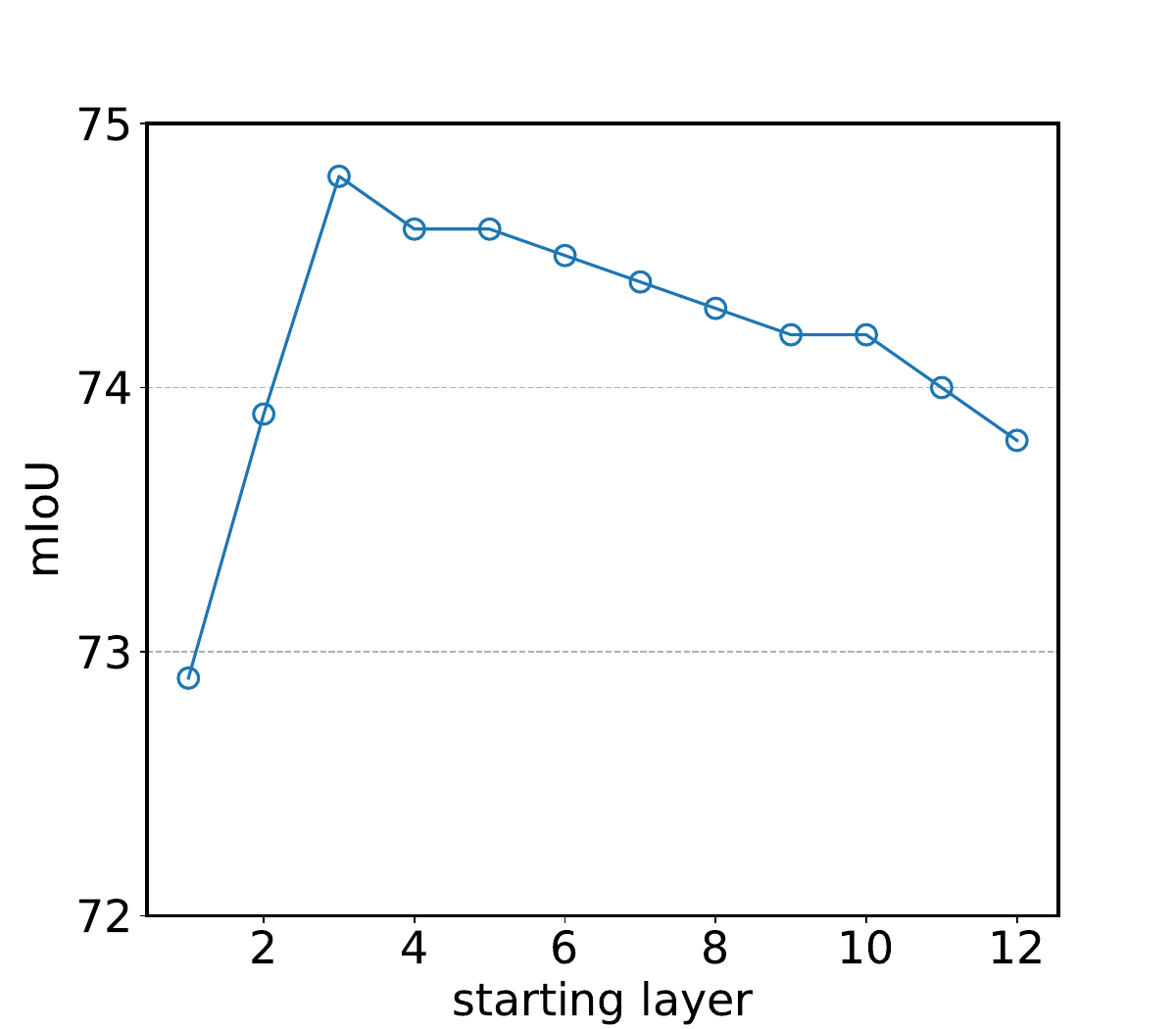}
\caption{
\small{Comparison of 2D knowledge transfer loss on different starting layers.}
}
    \label{fig:ab_2d_layer}
\end{figure}

%% file: chapters/MVNet/sections/06_conclusion.tex
\subsection{Conclusion}
\label{Sec:conclusion}

In this section, we present a novel approach to 3D point-cloud pre-training. We leverage pre-trained 2D networks to supervise 3D pre-training through perspective projection and hierarchical feature-based 2D knowledge transfer. We also introduce a novel pre-task to enhance the geometric structure through correspondence prediction. Extensive experiments demonstrate the effectiveness of our approach.

Potential avenues for future work include exploring additional pre-training data, such as NYUv2~\cite{couprie2013indoor}, and investigating the effect of increasing the size of the feature encoding network.  Furthermore, we plan to extend our approach to outdoor settings. MVNet opens up new avenues for 3D point-cloud pre-training and provides a promising direction for future research.

%% file: chapters/conclusion.tex
\chapter{Conclusion}
\label{sec:conclusion}

This dissertation presents significant advancements in point cloud representation learning through innovative approaches in supervised, self-supervised, and transfer learning. We introduced a novel method, HPNet, for point cloud primitive segmentation, integrating geometric heuristics with deep learning techniques to achieve superior performance. Our asymmetric point-cloud autoencoder, the Implicit AutoEncoder, demonstrates enhanced capability in capturing generalizable features from true 3D geometry, addressing the limitations of traditional autoencoders. Additionally, we developed a masked autoencoding method that focuses on predicting intrinsic point features, further improving self-supervised learning efficiency. Lastly, by leveraging pre-trained 2D image-based models to enhance 3D representation learning, we bridged the gap between 2D and 3D domains, showcasing significant improvements across various downstream tasks. These contributions collectively advance the state-of-the-art in point cloud understanding, providing robust frameworks for future research and practical applications in diverse fields.

In the future, we hope to investigate a more general representation for point cloud as a foundation model that allows multiple tasks to be performed effectively and efficiently.